\documentclass[manuscript,screen]{acmart} %,review
\usepackage{xcolor}

\newcommand{\ST}[1]{\textcolor{black}{#1}} % blue
\newcommand{\cmt}[1]{\textcolor{black}{#1}} % red
\renewcommand{\footnote}{*}
\AtBeginDocument{%
	\providecommand\BibTeX{{%
			\normalfont B\kern-0.5em{\scshape i\kern-0.25em b}\kern-0.8em\TeX}}}

\setcopyright{acmcopyright}
\copyrightyear{2022}
\acmYear{20xx}
\acmDOI{xxx}

\acmConference[Woodstock '18]{Woodstock '18: ACM Symposium on Neural
	Gaze Detection}{June 03--05, 2018}{Woodstock, NY}
\acmBooktitle{Woodstock '18: ACM Symposium on Neural Gaze Detection,
	June 03--05, 2018, Woodstock, NY}
\acmPrice{15.00}
\acmISBN{978-1-4503-XXXX-X/18/06}

\usepackage{lineno}
\modulolinenumbers[5]
\usepackage{booktabs}
\usepackage{adjustbox}
\usepackage{array}
\usepackage{amsmath,amsfonts} %,amssymb

\usepackage{algorithmic}
\usepackage{graphicx}
\usepackage{caption,subcaption}
\usepackage{multirow}
\usepackage{multicol}
\usepackage{textcomp}
\usepackage{wrapfig}
\usepackage{color}
\usepackage{tikz}
 
\usepackage{longtable,booktabs}
\usepackage{wrapfig}
\usepackage{graphicx}
\usepackage{longtable}
\usepackage{rotating}
\newcommand\n{\\\hline}

\usepackage{float} 
\usepackage[symbol]{footmisc}
\begin{document}
	
	%%
	%% The "title" command has an optional parameter,
	%% allowing the author to define a "short title" to be used in page headers.
	\title{Deep Learning for Plant Identification and Disease  Classification from Leaf Images: Multi-prediction Approaches}
	
	%%
	%% The "author" command and its associated commands are used to define
	%% the authors and their affiliations.
	%% Of note is the shared affiliation of the first two authors, and the
	%% "authornote" and "authornotemark" commands
	%% used to denote shared contribution to the research.
	\author{Jianping Yao}
	%\authornote{Joint first authors, both authors contributed equally to this research.}
	\email{jianping.yao@utas.edu.au}
	\authornote{Jianping and Son are joint first authors (equal contribution).}
	%\orcid{1234-5678-9012}
	
	\affiliation{%
		\institution{School of  Information and Communication Technology, University of Tasmania}
		\streetaddress{TAS 7248}
		%\city{Launceston}
		\state{Tasmania}
		\country{Australia}
		\postcode{7248}
	}
	
	\author{Son N. Tran}
	%\authornotemark[1]
	\authornote{Corresponding author (son.tran@deakin.edu.au).}
	%\thanks{aaa}
	\email{son.tran@deakin.edu.au}
	
	\affiliation{%
		\institution{School of  Information Technology, Deakin University}
		\streetaddress{VIC 3125}
		%\city{Launceston}
		\state{Victoria}
		\country{Australia}
		\postcode{3125}
	}
	\author{Saurabh Garg}
	\affiliation{%
		\institution{School of  Information and Communications Technology, University of Tasmania}
		\streetaddress{TAS 7248}
		%\city{Launceston}
		\state{Tasmania}
		\country{Australia}
		\postcode{7248}
	}
	\author{Samantha Sawyer}
	\affiliation{%
		\institution{Tasmania Institute of Agriculture, University of Tasmania}
		\state{Tasmania}
		\country{Australia}
	}

	%%
	%% By default, the full list of authors will be used in the page
	%% headers. Often, this list is too long, and will overlap
	%% other information printed in the page headers. This command allows
	%% the author to define a more concise list
	%% of authors' names for this purpose.
	\renewcommand{\shortauthors}{Yao and Tran, et al.}
	
	%%
	%% The abstract is a short summary of the work to be presented in the
	%% article.
	\begin{abstract}
\ST{Deep learning plays} an important role in modern agriculture, especially in plant pathology using leaf images where convolutional neural networks (CNN) are attracting a lot of attention. \ST{While numerous reviews have explored the applications of deep learning within this research domain, there remains a notable absence of an empirical study to offer insightful comparisons due to the employment of varied datasets in the evaluation. Furthermore, a majority of these approaches tend to address the problem as a singular prediction task, overlooking the multifaceted nature of predicting various aspects of plant species and disease types. Lastly, there is an evident need for a more profound consideration of the semantic relationships that underlie plant species and disease types}. In this paper, we start our study by surveying current deep learning approaches for plant identification and disease classification. We categorise the approaches into multi-model, multi-label, multi-output, and multi-task, in which different backbone CNNs can be employed. Furthermore, based on the survey of existing approaches in plant pathology and the study of available approaches in machine learning, we propose a new model named Generalised Stacking Multi-output CNN (GSMo-CNN). To investigate the effectiveness of different backbone CNNs and learning approaches, we conduct an intensive experiment on three benchmark datasets Plant Village, Plant Leaves, and PlantDoc. The experimental results demonstrate that InceptionV3 can be a good choice for a backbone CNN as its performance is better than AlexNet, VGG16, ResNet101, EfficientNet, MobileNet, and a custom CNN developed by us. Interestingly, there is empirical evidence to support the hypothesis that using a single model for both tasks can be comparable or better than using two models, one for each task. Finally, we show that the proposed GSMo-CNN achieves state-of-the-art performance on three benchmark datasets.
	\end{abstract}
	
	%%
	%% The code below is generated by the tool at http://dl.acm.org/ccs.cfm.
	%% Please copy and paste the code instead of the example below.
	%%
	\begin{CCSXML}
		<ccs2012>
		<concept>
		<concept_id>10010147.10010178.10010224.10010225</concept_id>
		<concept_desc>Computing methodologies~Computer vision tasks</concept_desc>
		<concept_significance>500</concept_significance>
		</concept>
		<concept>
		<concept_id>10010405.10010476.10010480</concept_id>
		<concept_desc>Applied computing~Agriculture</concept_desc>
		<concept_significance>500</concept_significance>
		</concept>
		</ccs2012>
	\end{CCSXML}
	
	\ccsdesc[500]{Computing methodologies~Computer vision tasks}
	\ccsdesc[500]{Applied computing~Agriculture}
	
	%%
	%% Keywords. The author(s) should pick words that accurately describe
	%% the work being presented. Separate the keywords with commas.
	\keywords{deep learning, convolutional neural networks, multi-prediction, plant identification, leaf disease classification, plant pathology}
	
	%%
	%% This command processes the author and affiliation and title
	%% information and builds the first part of the formatted document.
	\maketitle
	
	\section{Introduction}
	\label{sec:intro}
	Deep learning (DL) has been a major disruptor in a wide range of real-life applications from autonomous vehicles to clinical decision support. \ST{In agriculture, deep learning approaches have been emerging as a revolutionising tool for sustainable production. In particular, they play an important role in precision agriculture (PA)/smart agriculture (SA) \cite{9418245, edssjs.4D1044AE20210101, edssjs.8410D5A320210101, 9238318, edssjs.15D0A2D220200101}, including,  but not limited to, pest, weeds or irrigation control, automatic harvesting, yield estimation, and plant disease/fruit detection, etc. Deep learning within the field of plant pathology has garnered significant attention from both the research and industrial sectors. In these domains, the identification of plants and the classification of diseases based on leaf images have witnessed extensive study and practical application \cite{computers8040077, 9804121, Shelke2022, 9631212, 4458016, 8974752, 9137986, S187705092030690620200101, 9418013, S004579061930002320190601, 9342729, 9412643}}.
	The reason why leaf images are commonly used is leaves are an important part of plants where they participate in the important photosynthesis process and are the most visible part of most plants throughout their growth \cite{edssjs.15D0A2D220200101}. Leaf colour, texture, and shape can characterise plant species and, therefore, are useful in plant identification for large-scale plant and crop management \cite{4458016}. Besides, many plant diseases can be visible from leaves. 
	%Leaf diseases will cause plants to wither and even lower crop yields. Thus, an effective leaf disease classification approach could help ensure the state of plants, increase crop yields and reduce losses from plant diseases \cmt{paraphrase it, similar to the review paper}. 
	Leaf disease is one of the important factors that disrupt the health of plants as a whole and are one of the main causes of reduced crop yield. Therefore, it is critical for farmers to detect the occurrence of leaf disease as early as possible and minimise its negative impact or, at least, keep it under control. In general, there is a rising demand from the industry for effective methods to accurately detect and/or classify leaf diseases.
	
	Deep learning is revolutionising traditional methods for SA/PA, especially in plant and disease classification. Although being popular in the past, traditional approaches have several obvious limitations, mostly caused by the manual costs, such as labours training, requirement of human involvement in many stages of the prediction process, and experts' knowledge, etc. Manual methods are difficult to detect timely, and the diagnosis may be based on subjective judgment. Nowadays, with the assistance of computer vision, Internet of Things (IoT) and machine learning, we could detect leaf diseases in real-time through various devices, e.g., mobile phone applications\cite{9397001}, websites\cite{9342653}, \ST{IoT application \cite{8871173}} and smart glasses\cite{9182146}. \ST{For example, \cite{8871173} introduced a combined approach of IoT and AI models to detect rice blast disease. Compared to our work, this approach uses a custom convolutional neural network (CNN), similar to the first model we implement in 3.3.1, but it work on different input data. In particular, in \cite{8871173} an IoT platform for soil cultivation was utilised to extract non-image data while our study focuses on image data.} 
	
	With these advanced technologies, the difficulty of the production operation can be greatly reduced, and we can achieve improvements in accuracy and efficiency. Among them, machine learning has been emerging as a key player, leading the innovation pathway for more effective prediction solutions in plant identification and disease classification.
	During the earlier stage, researchers employed traditional machine learning approaches, combining feature extraction and classification,  with limited successes \cite{9362812, 9130019, 9077134, 9182128, 9210294, a_15100615120210401, 9422499, 9076371}.
	%In addition, it focuses on specific species and diseases (i.e., rice blast disease), which may slightly limited applicability for various leaf diseases from various species. 
	The key issue here is that such approaches rely heavily on an independent step to craft features from the images to classify plant species and disease types. The feature engineering process is normally defined by domain experts or generated by general image processing techniques such as Scale-invariant Feature Transform (SIFT) \cite{a_15100615120210401}, grey-level co-occurrence matrix (GLCM) \cite{9418680,9076371,9422499,8944556,9277379}, etc.
	
	As being independent of the later step of learning a classifier, these handcrafted features may not be optimal for prediction. Deep learning has been emerging recently as an effective solution. For example, CNNs can provide an end-to-end classification pipeline to identify plants and classify diseases directly from leaf images. An advantage of CNNs is their ability to learn distinctive features tailored to specific tasks. Furthermore, their adaptability allows for fine-tuning models to maximise effectiveness for unique datasets. Finally, their distributed computation capability makes them an ideal choice for large-scale solutions, enabling efficient processing and analysis of substantial datasets. %  line 58, all the papers that classify both leaf and disease
	
	Although the application of DL for plant identification and disease classification is not new, we found that most current studies develop CNN models as single prediction classifier.  It would be more convenient if there exists a multi-prediction approach for plant species and disease types as they have different indicative features from leaf images. More importantly, it is apparent that besides common types of disease different plant species will be prone to different diseases of their own. Therefore, we hypothesise that by incorporating the two tasks together in one single model, we can improve the prediction performance for each task. To investigate the proposed hypothesis, we employ multi-label/multi-output/multi-task approaches with deep learning as the core. \ST{The main constraint of the task is a requirement for multiple labels of each image to enable the training of deep models}. Let us formally define the problem as follows.
	
	\textbf{Problem Statement:} {\it Given a data set $\mathcal{X}=\{(x^{(n)},p^{(n)},d^{(n)})|n=1,...,N\}$, where $x^{(n)}\in \mathbb{R}^{W\times H \times C}$ is an image with a width $W$, height $H$, and $C$ channels; $p^{(n)} \in \mathcal{P}$ is a plant species; and $d^{(n)} \in \mathcal{D}$ is a plant disease, how to train a deep learning model $\mathcal{N}$ to accurately identify the plant species and the disease type from an unseen image $x^*$?}
	
	A plethora of deep learning models, mostly CNNs, have been employed for plant and leaf classification separately, including  AlexNet \cite{S004579061930002320190601, 9137986, 8974752, t_14770449620201201}, GoogLeNet \cite{9418245, 8374024}, VGG \cite{S187705092030690620200101, t_14770449620201201, 9231174,8974752,9291694, 9261801}, Inception \cite{S187705092030690620200101, 15100606920210401, electronics10121388, 14844519620210101}, ResNet \cite{t_14770449620201201, 9408806, 9155585, 9418245}, MobileNet \cite{mwebaze2019icassava, 9392051, S187705092030690620200101, 9291694}, etc.
	However, there are several questions which are not been studied properly, including (i) which back-bone CNNs can be most useful for plant identification and disease classification; (ii) what other deep learning approaches can be employed for this task; (iii)  whether separate models for plant identification and disease classification perform better than a single model for both tasks; and (iv) whether their performance comparison is consistent across different datasets.

	In this paper, we aim to answer the above questions, and finally verify our hypothesis, by surveying, developing, and comparing a wide range of CNN architectures. To this end, we solve the problem of plant identification and disease classification by employing and evaluating a variety of deep learning approaches. We conduct an empirical study to analyse the usefulness of current deep learning models for plant species identification and leaf disease detection from leaves. We categorise the deep learning approaches into:
	\begin{itemize}
		\item Multi-model: This is an ensemble of two CNNs models, one is tasked to predict plant species from leaf images while the other is tasked to predict diseases. For completeness, we will use and compare different backbone CNNs for these models, including our custom CNNs, AlexNet, VGG16, ResNet101, EfficientNet, InceptionV3, and MobileNetV2.
		\item Multi-label: This is a single CNN model with an output of multi labels. Different from the standard multi-label learning \cite{MADJAROV20123084}, in this case, we combine the labels to make a power-set label for the prediction. In other words, we use a combined label to present multiple labels. We also employ different backbone CNNs for this approach.
		\item Multi-output: This is a single  CNN model with multiple output layers, each for a prediction target. Similarly, different backbone CNNs are used for this model as well.
		\item Multi-task.  We can also adapt multi-task learning to this problem. In this case, the target tasks are different but the input data for the tasks will be from the same distribution. Multi-task learning has been employed effectively in many computer vision problems \cite{ 6220250, edssjs.752074520130101} and deep learning can bolster this class of approaches because one task’s features may benefit other tasks’ learning \cite{9392366, 10.1007/978-3-319-10599-4_7}.
	\end{itemize}

	Furthermore, based on the theoretical analysis of the above approaches, we propose a new method to improve the performance of plant identification and disease classification. After that, we review and select suitable benchmark datasets for an extensive experiment. For the empirical analysis, we use 3 different datasets including Plant Village \cite{hughes2016open}, Plant Leaves \cite{chouhan_kaul_singh_jain_2019}, and PlantDoc \cite{Singh_2020}.  The experimental results show that {InceptionV3} is the best CNN backbone for both plant identification and disease classification. Interestingly, there is empirical evidence to support the hypothesis stated earlier, as single models for both tasks can achieve better performance than an ensemble of independent models. Finally, our proposed model can achieve state-of-the-art results on the benchmark datasets studied in this paper.
	% \clearpage
	The contribution of this paper is threefold:
	% \begin{itemize}
		%     \item Applications of different deep learning approaches, including those that have not been employed before, to plant identification and disease classification.
		%     \item A comparative analysis of deep learning models' performance in plant and disease classification on multiple datasets. 
		%     \item A new approach to improve the performance of plant identification and disease classification. We achieve state-of-the-art results in all datasets studied in this paper.
		% \end{itemize}
	\ST{\begin{itemize}
			\item We conduct a detailed literature review through appropriate selection criteria, including recent common machine learning methods and  available public datasets in the field of plant identification and disease classification. We generalise the methods to two classes of multi-prediction paradigm (multi-model CNNs, multi-label CNNs) and extend the study by re-introduce several methods available in the machine learning literature (multi-ouput CNNs, multi-task learning) but have not been studied deeply in the research field of plant pathology. As far as we know, this is the first study to survey and compare different deep multi-prediction techniques. 
			\item  We proposed a novel model, namely Generalised Stacking Multi-output CNNs (GSMo-CNN), to improve the performance of plant identification and disease classification. Our model creates multiple output layers and stacks them one on top another to form a chain of classification in a hierarchical manner. By doing this, we aim to represent the relationship between plant species and disease types during the inference process. The experimental results show that GSMo-CNN achieves state-of-the-art performance  in all datasets studied in this paper. 
			\item Our study reveals several important empirical findings. First, we show that selection of backbone CNNs is critical for achieving good performance and InceptionV3  has the best performance in our study. Second, we demonstrate the advantage of single models for multi-prediction over multiple models. This is an interesting finding as single models are more compact and easier to train.  Third, we showcase the effectiveness of combining labels in a hierarchical manner, together with transfer learning we can significantly improve the prediction performance for both plant species and leaf diseases. These findings offer valuable insights for researchers and practitioners in agriculture. It would save their time to search for suitable approaches for more accurate and efficient plant species identification and disease classification, ultimately contributing to improved crop management and disease control. For the sake of reproducibility, we share the source code and datasets in this study at: \url{https://github.com/funzi-son/plant_pathology_dl}.
	\end{itemize}}
	
	%The paper will provide an analytic view of which backbone CNNs are most useful for predicting plant species and diseases. We also show the effectiveness of different deep learning approaches on benchmark datasets, providing comparative evidence of state-of-the-art methods and an empirical benchmark for future studies. As far as we know, this paper is the first study to present a comprehensive comparison of deep learning models for plant identification and disease classification from leaf images. Furthermore, we provide a method to improve the classification performance of both plant species and diseases. 

	%The paper is organised as follows. In section \ref{sec:related_work}, we survey and discuss the related work in deep learning for plant identification and leaf disease classification, which will show the current trends in this research topic. { Section \ref{sec:methodology} will detail the setup for the empirical study. We survey and select suitable datasets and evaluation metrics. In this section, we also survey deeper the deep learning approaches, most of which haven't been but can be employed to solve the plant identification and disease classification problem and verify our hypothesis. In section 	\ref{sec:experiment} the experimental results will be discussed and analysed}. Here, we also summarise our findings and provide the ablation study. Finally, we conclude the paper and outline several directions for future work in section \ref{sec:conclusion}.
	
	% \clearpage
	
	\ST{\section{Machine Learning for Plant Identification and/or Disease Classification}}
	\label{sec:related_work}
	\ST{\subsection{Article Selection Criteria}}
	\begin{table}[ht]%{0.7\textwidth}
		%\vskip -0.5cm
		% \begin{table}[H]
			% \centering
			\begin{center}
				% \begin{minipage}{\textwidth}
					\caption{\ST{The Publication Years of Referenced Academic Articles (* Denotes Review Papers).}}\label{Selection}%
					\resizebox{\textwidth}{!}{%
						\begin{tabular}{|c|m{0.95\textwidth}|c|c|c|}
							\hline
							\multirow{1}{*}{Year} & \multicolumn{1}{c|}{Paper} & Tech  & Review  & Total \\
							%\cline{2-7}
							\hline
							\hline
							% 2022 & ,   &  8 & 4 & 12\\
							% \hline					
							2021 - 2023 & \cite{9587775}, \cite{SUJATHA2021103615}, \cite{9418245}, \cite{9422499}, \cite{9418680},  \cite{9396023}\footnote, \cite{9397001}, \cite{9408806}, \cite{15100606920210401}, \cite{15107730820210601}, \cite{9418013}, \cite{15100606020210401}, \cite{edssjs.30A9493B20210101}, \cite{9399342}\footnote, \cite{edssjs.8410D5A320210101}, \cite{edssjs.4D1044AE20210101}, \cite{S004579062100047120210301}, \cite{a_15100615120210401}, \cite{electronics10121388}, \cite{9568324}\footnote, \cite{14844519620210101}, \cite{Kathiresan_2021}, \cite{9725870}\footnote, \cite{154722790}, \cite{10.1007/978-3-030-84522-3_24}, \cite{9336293}, \cite{hassanin2021learning}, \cite{9412643}, 
							\cite{ABDALGANI2023100643}, \cite{10150884}, \cite{s11042-023-16347-0} & 27 & 4 & 31 \\
							\hline
							2020 &\cite{10.3389/fpls.2020.00751}, \cite{Singh_2020}, \cite{9362812}, \cite{9076371}, \cite{9130019}, \cite{9291694}, \cite{9261801}, \cite{9077134}, \cite{9350413}, \cite{9137986}, \cite{9155585}, \cite{9250911}, \cite{9250885}, \cite{9277379}, \cite{9392051}, \cite{9342653},   \cite{9142988}, \cite{9342729}, \cite{9212816}, \cite{9231174}, \cite{9182128}, \cite{9182146}, \cite{9238318}\footnote, \cite{9210294}, \cite{9057889}, \cite{S187705092030690620200101}, \cite{t_14770449620201201}, \cite{edssjs.15D0A2D220200101}\footnote, \cite{FU2020122}, \cite{8871173} & 28 & 2 & 30 \\
							\hline
							2019 - 2015 &\cite{S004579061930002320190601}, \cite{8974752}, \cite{8944556}, \cite{mwebaze2019icassava}, \cite{8374024}, \cite{8566635}, \cite{7746160}, \cite{DBLP:journals/corr/HughesS15}, \cite{hughes2016open}, \cite{DOSSANTOSFERREIRA2017314}, \cite{9016966}, \cite{Misra_2016_CVPR}, \cite{shinohara16b_interspeech}, \cite{Wang_2016_CVPR} & 14 & 0 & 14 \\
							%						\bottomrule
							\hline
						\end{tabular} %
					}
					% \end{minipage}
				%\footnote{Review papers}
			\end{center}
			\vskip -0.5cm
		\end{table}
		\ST{The academic papers selected for this study primarily focus on three key aspects. Firstly, the publication timeline, as illustrated in Table \ref{Selection}, reveals that over 80\% of the research articles in this study were published between 2020 and 2023. This timeframe was chosen to ensure the effectiveness and timeliness of this research. Secondly, the degree of citation and the impact factor of the journals (such as Q1 for journals and CORE A/A* for conferences) were considered. Lastly, the relevance to this research was determined through keyword searches, including "leaf disease," "plant disease," "machine learning," "deep learning," "classification," and "detection." These keywords were used to search in reputable databases such as EBSCO host and Scopus, and Google Scholar.}

		% \begin{figure*}[h]
			% 	\centering
			% 	\includegraphics[width=0.7\textwidth]{Figures/Chapter_3/amount_year_2.png}
			% 	\centering
			% 	\caption{The Amount \& Years of Referenced Articles}
			% 	\label{fig:amount_year_2}
			% \end{figure*}
		
		\subsection{Traditional Machine Learning versus Deep Learning}
		
		In the earlier years, traditional (shallow) machine learning was used for plant identification and leaf disease classification \cite{154722790, 10.3389/fpls.2020.00751, S187705092030690620200101, 9261801, 8566635, 9350413, 9250911}. This machine learning paradigm in plant pathology consists of two different steps: feature extraction \cite{9396023, 9399342, 9725870, 9587775} and classifier training \cite{9362812, 9130019,8944556, 9076371}.  In some cases, researchers also consider including data segmentation after collecting and pre-processing data before applying feature extraction \cite{9725870, 9587775}. Among many feature extraction techniques,  K-means clustering \cite{9362812,7746160,9277379,9212816} and grey-level co-occurrence matrix (GLCM) \cite{9076371,8944556,9418680,9277379,9422499} are the most common feature extraction methods. In terms of shallow learning classifiers, Support Vector Machines (SVMs) \cite{9362812, 9130019, 9077134, 9182128, 9210294, 9422499, 7746160, 9076371, 9418680, 9277379, 9212816, edssjs.30A9493B20210101} was the most popular, followed by K-Nearest Neighbor (KNN) \cite{8944556, 9076371}, Random Forest (RF) \cite{9422499}, Multilayer Perceptron (MLP) \cite{a_15100615120210401}, and Decision Tree \cite{9142988}. They are all popular in machine learning applications and achieve good performance in leaf disease detection and/or classification. In the case where data segmentation is used, we will need to detect the region of interest as a part of the data before feeding it to feature extractors and then classifiers. As we can see, the whole process is complex, involving several consecutive steps such as data acquisition, data processing, feature extraction, and prediction \cite{9396023, 9399342, 9725870, 9587775}.
		
		% \begin{wraptable}{i}{0.55\textwidth}
			%\vskip -0.5cm
			\begin{table}[h!]
				\begin{center}
					% \begin{minipage}{\textwidth}
						\caption{\ST{Comparison Between Deep Learning (DL) and Non-deep Learning (Non-DL) Approaches for Plant Pathology Using Leaf Images.}}\label{ML_DL}%
						\resizebox{0.55\textwidth}{!}{%
							%				\begin{tabular}{@{}lcp{0.15\textwidth}p{0.4\textwidth}@{}}
								\begin{tabular}{@{}ccp{0.2\textwidth}p{0.08\textwidth}p{0.08\textwidth}}
									%					\begin{tabular}{@{}ccp{0.4\textwidth}p{0.4\textwidth}@{}}					
										\toprule
										Paper & Year & Dataset &  Non-DL & DL\n
										%						\midrule
										\cite{9057889} & 2020 & Plant Village (Part)  &  66.4\% & \textbf{98\%} \n		
										\cite{9418013} & 2021 & Plant Village (Part) & 90\% & \textbf{96\%}\n
										\cite{8974752} & 2019 & Plant Village (Grape)& 97.5\% &  \textbf{99\%} \n		
										\cite{S004579061930002320190601} & 2019 & Plant Village (Whole)  &  87.87\% & \textbf{97.87\%} \n	
										\cite{9155585} & 2020 & Plant Village (Modified)  & 88.06\% & \textbf{96.51\%}  \n	
										\cite{9418245} & 2021 & Plant Village (Apple) &   68.73\% & \textbf{97.62\%} \n
										\cite{9137986} & 2020 & Tomato Leaves  & 92.94\% &\textbf{98.12\%} \n
										\cite{SUJATHA2021103615} & 2021 & Citrus Leaves\cite{RAUF2019104340} & 87\% &  \textbf{89.5\%}\n
										\cite{ABDALGANI2023100643} & 2023 &  Citrus Leaves \cite{RAUF2019104340} & 86\% & \textbf{99.98\%} \\	
										\bottomrule
									\end{tabular} % 
								}
								\footnotetext[1]{Acc \& F1}
								
								% \end{minipage}
						\end{center}
						%\vskip -0.3cm
					\end{table}
					% \end{wraptable}			
				\ST{Substantial changes in technology adoption can be seen under the rise of deep learning. From recent studies, \ST{as Table \ref{ML_DL} shows}, researchers have affirmed that deep learning can achieve better performance than traditional (non-deep) learning approaches \cite{SUJATHA2021103615, 9057889, 9418013, 8974752, S004579061930002320190601, 9155585, 9418245}}. A class of deep learning models, known as convolutional neural networks (CNNs), have been widely applied for plant identification and/or leaf disease classification. In \cite{DOSSANTOSFERREIRA2017314} the authors employed a popular CNN model named AlexNet for plant classification. In this work, AlexNet was shown to successfully classify grass and broadleaf with average accuracy up to 99\%. For leaf disease classification, AlexNet achieved 91.19\% accuracy on Apple leaf images \cite{9418245}, 86.5\% on Grape leaves \cite{8974752} and 95.75\% on tomato leaf images \cite{9137986}. Note that, they are separate models, each for a task. A deeper architecture, known as very deep Convolutional Neural networks (or VGG, VGGNet) has shown better performance than AlexNet in image classification tasks. When applied to plant identification, VGG achieved 97\% accuracy on Leaf1 Dataset (8 species), 96.57\% on Flavia Dataset (32 species) and 85.37\% on D-Leaf Dataset (43 species) \cite{9016966}. VGG for leaf disease classification also received promising results. For example, in grape leaf disease VGG-16 (16 hidden layers) has been applied to many datasets \cite{8974752,9291694, 9261801}. Notably, in \cite{9261801} VGG-16 with Average Pooling (GAP) layer achieved 98.4\% accuracy. Another common architecture of VGG is VGG with 19 hidden layers, known as VGG-19, which had achieved 96.86\% accuracy in tomato leaf disease classification \cite{9231174}.
				
				Another CNN architecture, known as Residual Networks or ResNets, has shown remarkable results in image classification tasks in recent years (it won the 2015 ImageNet competition). In \cite{10.1007/978-3-030-84522-3_24}, a set of ResNet variants has been employed for plant identification. An evaluation on a real leaf dataset that the authors collected with 15207 images (of 201 species) is reported as follows, 91.83\% (ResNet-50); 92.71\% (Res2Net-50); 92.95\% (Res2Net-101).
				In the case of leaf disease classification, ResNet-50 achieved 98.40\% accuracy for tomato leaves \cite{t_14770449620201201},  a customised ResNet model had 82.78\% accuracy on a modified Plant Village dataset \cite{9408806}, ResNet-34 achieved 99.40\% accuracy and  0.9651 F1-score in betel vine leaf disease \cite{9155585}. In \cite{9418245} ResNet-20 was tested on apple leaves images and achieved 92.76\% accuracy. Last but not least, InceptionV3 is recently emerging as a good model for classification tasks with leaf images. InceptionV3 is the third version of Google's Inception CNN. In \cite{S187705092030690620200101}, it achieved 63.4\% accuracy on tomato leaf diseases and in \cite{15100606920210401}, it achieved 95.41\%  in rice leaf disease classification. InceptionV3 has been applied to the famous Plant Village dataset and achieved very promising results, as shown in \cite{electronics10121388} (98.42\% accuracy) and in \cite{14844519620210101} (99.74\% accuracy).

				Besides the very deep and complex models discussed above, several studies also showed the advantages of light-weight CNNs in plant identification and leaf disease classification. One of the advantages of light-weight CNNs is they can run on low-resource devices, enabling a wider range of applications in smart agriculture and precision agriculture. For example, MobileNet has been deployed for smartphones and IoT devices. In \cite{S187705092030690620200101}, it achieved 63.75\% accuracy in tomato leaf disease classification and in \cite{9291694} it achieved 86\% accuracy in grape leaf disease classification. Another light-weight CNN model is EfficentNet whose different variants (B0, B4, B7) have been used to classify tomato leaf diseases (Plant Village) \cite{15107730820210601}. To enable high performance for EfficientNet the authors have relabeled the dataset in three subtasks: task 1: healthy \& unhealthy; task 2: 5 classes of leaf state, including bacterial \& fungal; task 3: 1 healthy \& 9 diseases. The results show that B7 got the best in task 1 (99.95\%) and task 2 (99.12\%) and B4 got the best in task3 (99.89\%).
				
				The above related work applied CNNs separately for plant identification and disease classification from leaf images. Each paper evaluates the CNNs on a different dataset, and sometimes on a modified dataset, making it difficult to benchmark their performance. Different from them, this paper sets up a comprehensive evaluation to provide a comparative view of the CNNs, using multiple datasets.
				
				Several attempts in recent years have shown promising approaches of using a single CNN for multiple tasks \cite{Misra_2016_CVPR, FU2020122, 9336293, shinohara16b_interspeech, hassanin2021learning,  Wang_2016_CVPR}. In the case of plant pathology, a CNN can learn to predict both plant species and diseases at the same time with leaf images as input. For example, in \cite{9412643} the authors showed the effectiveness of conditional multi-task learning for the simultaneous identification of plant species and classification of diseases. Unlike other studies on large-scale multi-task learning, the approach adapts the multi-task learning idea for interrelated labels, where input data from different tasks are from the same distribution. In Plant Village dataset and PlantDoc dataset, the labels are the combination of both species and diseases, e.g., Apple Black Rot in Plant Village dataset and Tomato Leaf Late Blight in PlantDoc dataset. This combination deals with the multi-prediction problem by creating a multi-label known as power-set. This can transform a multi-task or multi-label problem to a large-scale multi-class task. A trained model can predict the species and diseases simultaneously, and the predicted results are joint species-disease labels. Several examples employed this idea are shown in Table \ref{DL_Comparison}. In \cite{9057889, 9418013, 9342729} the authors applied power-set CNNs on a subset of Plant Village dataset, and in \cite{S004579061930002320190601, 9155585, 9408806, 9250885} the author worked on the whole or modified Plant Village dataset.  All these studied models achieved more than 90\% accuracy. For example, according to a study on PlantDoc dataset, with the support of image segmentation and power-set labelling VGG-16 achieved 60.41\% accuracy, InceptionV3 achieved 62.06\% accuracy, and InceptionResNet V2 achieved 70.53\% accuracy \cite{Singh_2020}.
				
				% \clearpage
				% \newgeometry{margin=1cm}
				% \thispagestyle{empty}
				\begin{table}[p!]
					\begin{center}
						\begin{minipage}{\textwidth}
							\caption{Existing Deep Learning Approaches for Plant Identification and/or Leaf Disease Classification}\label{DL_Comparison}%
							\resizebox{\textwidth}{!}{%
								\begin{tabular}{@{}lcp{0.2\textwidth}lp{0.2\textwidth}p{0.6\textwidth}@{}}
									%					\begin{tabular}{@{}ccp{0.4\textwidth}p{0.4\textwidth}@{}}					
										\toprule
										Paper & Year & Dataset & Categories& Size/Ratio (training/test)  & Methods \& Accuracy \\
										\hline
										\hline \multicolumn{6}{c}{\textbf{Plant Identification}}\\
										\\
										
										\cite{9804121} & 2022 &  Coriander \& Parsley (Private)  & 2 & 100 (70\%/ 30\%)  & CNN (90\%) \n
										% 79                   2591 (80\%/20\%)      DenseNet-161   97.3\%
										\cite{Shelke2022} & 2022 & Private Dataset  & 79 & 2591 (80\%/20\%)  & DenseNet-161 (97.3\%) \n
										
										& & Middle European Woody Plants \cite{NOVOTNY2013444}  & 119 & (80\%/20\%) & LR(98.72\%) \\	
										& & Flavia Dataset \cite{4458016}  & 32 & 1703 (80\%/20\%)& LR(99.58\%)\\					
										& & MalayaKew (MK) Leaf Dataset \cite{7350839}  & 44 & 2816 (80\%/20\%) & LR(89.35)\%\\
										\cite{9631212} & 2021 & MK Leaf \cite{7350839} + Synthetic Dataset   & 44  & (80\%/20\%) & LR(93.33)\%\\	
										& & Folio Dataset \cite{MUNISAMI2015740}  & 32 & (80\%/20\%) & LR(98.75\%)\\						
										& & Amazon Forest \cite{VIZCARRA2021101268}  & 9 & 59,441 (80\%/20\%) & LR(98.87\%)\\											
										& & LeafSnap Dataset \cite{10.1007/978-3-642-33709-3_36}  & 185 & 23,147 (80\%/20\%) & LR(89.27\%)\\
										& & Swedish Dataset \cite{Soderkvist303038} & 15 & 1125 (80\%/20\%) & LR(100\%)\n
										& & Swedish Dataset  & 15 & 1125 (70\%/15\%/15\%\footnotemark[1]) & \textbf{ANN (98.99\%)}, KNN (96.68\%) \& RF (97.12\%) \\
										\cite{154722790}& 2021 & Flavia Dataset  & 32 & 1907 (70\%/15\%/15\%\footnotemark[1]) & \textbf{ANN (96.29\%)}, KNN (93.79\%) \& RF (95.24\%) \\
										& & D-Leaf Dataset\cite{TAN2018}  & 43 & 1290 (70\%/15\%/15\%\footnotemark[1]) & \textbf{ANN (95.31\%)}, KNN (86.3\%) \& RF (91.5\%) \n
										
										& & Leaf1 Dataset & 8 & 75 (80\%/20\%) & \textbf{GoogLeNet (98\%)} \& VGG-16 (97\%)  \\
										\cite{9016966}& 2019 & Flavia Dataset  & 32 & 1879 (80\%/20\%) & GoogLeNet (94\%) \& \textbf{VGG-16 (96.57\%)} \\
										& & D-Leaf Dataset\cite{TAN2018}  & 43 & 1290 (80\%/20\%) & \textbf{GoogLeNet (88.74\%)} \& VGG-16 (85.37\%) \\
										
										\hline \hline 
										%%%%%%%%%%%%%%%%%%%%%%%%%%%%%%%%%%%%%%%%%%%%%%%%%%%%%%%%%%%%%%%%%%%%%%%%%%%%%%%%%%%%%%%%%%%%%%%%%%					    
										\multicolumn{6}{c}{\textbf{Disease Classification}}\\
										\\
										
										\cite{8974752} & 2019 & Plant Village (Grape)  & 4 & 3800/200  & %Decision Tree (80.5\%), Naive Bayes (69.5\%), SVM (85\%), LDA (80.5\%), KNN (96\%), LR (94\%), RF (97.5\%), ANN (87.5\%) \\				 
										%&  &    &   &   & &  N/A &  
										\textbf{CNN (99\%)}, Alexnet (86.5\%), VGG16 (97.5\%)   \n				
										\cite{9418245} & 2021 & Plant Village (Apple)  & 4 & 10888/2801  %&  SVM (68.73\%) \& BP (54.63\%) \n &  &    &   &   & 
										&  \textbf{CNN (97.62\%)}, AlexNet (91.19\%), GoogLeNet (95.69\%), ResNet-20 (92.76\%) \& VGG-16 (96.32\%)  \n 
										\cite{9137986} & 2020 & Tomato Leaves (Self) & 4 & N/A  &  \textbf{CNN (98.12\%)}, AlexNet (95.75\%) \& ANN (92.94\%)  \n
										%						\cite{SUJATHA2021103615} & 2021 & Citrus Leaves \cite{RAUF2019104340}  & 5 & 10-fold  %&  RF (76.8\%), SGD (86.5\%), SVM (87\%)   \n 
										%&  &  & &  &  &  & VGG-19 (87.4\%), InceptionV3 (89\%) \&\textbf{VGG-16 (89.5\%)}     \\	  				
										%\cite{8374024} & 2018 & Maize Leaves (Self \& Plant Village)& Multi-class & 9 & 80\%/20\% & N/A &  \textbf{Fine-tuned GoogLeNet (98.9\%)} \& Cifar10 (98.8\%)  \n
										
										%\cite{9291694} & 2020 & Grape Leaves (Self)  & 5 & 80\%/20\%  & Vanilla CNN (98\%), Improved VGG-16 (99\%), Improved MobileNet (97\%), Improved AlexNet (97\%) \&  \textbf{Ensemble (100\%)} \\
										%\cite{10.3389/fpls.2020.00751}& 2020 & Grape Leaves (Self) & 4 & 4,449   &  Faster R-CNN (81.1\%)  \n
										\cite{S187705092030690620200101} & 2020 & Plant Village (Tomato)  & 10 & 10,000/7,000/500\footnotemark[1]   & \textbf{CNN (91.2\%)}, Mobilenet (63.75\%), VGG-16 (77.2\%)\& InceptionV3 (63.4\%) \n
										%\cite{9261801} & 2020 & Grape Leaves (Self) & Multi-class & 6 & 80\%/20\% &  None & VGG-16 with Global Average Pooling (GAP) (98.4\%) \n	
										\cite{8566635} & 2018 & Plant Village (Tomato)  & 5 & 500(80\%/20\%)  & CNN (86\%) \n	
										\cite{15107730820210601} & 2021 & Plant Village (Tomato) & 2, 6, 10 & 5-fold  & EfficientNet B0, B4, B7 (97\% - 99\%) \n	
										%\cite{9350413} & 2020 & Tea Leaves (Self)  & 3 & 1000/270/30\footnotemark[1]   & CNN (95.93\%) \n
										%\cite{9250911} & 2020 & Betelvine Leaves (Self) & 3 & 1,014  & F1-score: \textbf{Proposed Mask-RCNN (ResNet50 \& Feature Pyramid Network) (84.07\%)}, Faster RCNN (74.32\%) \& Mask RCNN (83.11\%) \n
										%\cite{9392051} & 2020 & Cassava Leaves\cite{mwebaze2019icassava}  & 5 & 5,656/1,889/1,885\footnotemark[1]   & MobileNet (85.38\%) \n							
										%\cite{15100606020210401} & 2021 & Pepper Leaves (Self) & 2    & \textbf{DBN (91.956\%\&77.546\%)}, FFNN (91.156\%\&63.936\%), BPNN (91.306\%\&66.916\%), DNN (91.386\%\&67.246\%), RNN (91.436\%\&67.486\%) \& CNN (91.616\%\&72.046\%) \footnotemark[2] \n	
										%\cite{S004579062100047120210301} & 2021 & Tea Leaves (Self) & 3 & 318/80, 1400/200   & Faster Region-based CNN (91.22\%) \n	
										
										\cite{t_14770449620201201} & 2020 & Plant Village (Tomato) &  10 & 80\%/20\%  & \textbf{Xception V4 (99.45\%)}, AlexNet (90.1\%), Lenet (88.3\%),  Resnet (98.40\%) \& VGG-16 (90.1\%) \n	
										\cite{ABDALGANI2023100643} & 2023 &  Citrus Leaves \cite{RAUF2019104340} & 5 & 609 & C-GAN(99.6\% \& 97\%),CNN(99.97\%\&99.98\%), SGD(85\%\&86\%) \& \textbf{ACO-CNN (99.98\%\&99.99\%)}\footnotemark[2] \n
										\cite{10150884} & 2023 & Cotton Leaf Disease Dataset & 4 & 1661 & SVM(98.7\%\&98.7\%), CNN(98.8\%\&98.8\%) \& \textbf{Hybrid(98.9\%\&98.9\%)}\footnotemark[2] \n	
										\cite{s11042-023-16347-0} & 2023 & Plant Village (Tomato) &  10 & 80\%/20\%  & InceptionV3(94.58\%), MobileNetV1(82.7\%), MobileNetV2(92.1\%) \& \textbf{MX-MLF2(99.61\%)}\\		
										
										%						  &  &  & &  &  &   &   \\	  	
										\hline \hline
										%%%%%%%%%%%%%%%%%%%%%%%%%%%%%%%%%%%%%%%%%%%%%%%%%%%%%%%%%%%%%%%%%%%%%%%%%%%%%%%%%%%%%%%%%						
										\multicolumn{6}{c}{\textbf{Single model for Plant Identification \& Disease Classification}}\\
										\\
										\cite{9057889} & 2020 & Plant Village (Part) & 19 & N/A %& GB \& K-means & LR (66.4\%), KNN (54.5\%) \& SVM (53.4\%)  \\ & & & & & & N/A 
										& \textbf{CNN (98\%)}  \n
										\cite{9418013} & 2021 & Plant Village (Part) & 15 & 80\%/20\% %& \begin{tabular}{@{}l@{}} Co-occurrence \\ Matrix \end{tabular} &  ANN (90\%), KNN (88.6\%), SVM (85\%), Naïve Bayes (79.6\%) \& K Means (72.3\%)  \\ & & & & & & None 
										& \textbf{CNN (96\%)}  \n
										
										\cite{S004579061930002320190601} & 2019 & Plant Village (Whole)  & 38 & 55,636/1950 %& N/A & SVM (50.69\%), Decision Tree (72.24\%), LR (81.00\%) \& K-NN (87.87\%)  \\ &  &  & &  & & None 
										&\textbf{CNN (97.87\%)}, AlexNet (87.34\%), ResNet (92.56\%), VGG16 (92.87\%) \& InceptionV3 (94.32\%) \n
										\cite{9155585} & 2020 & Plant Village (Modified)  & 38 & 15,200 (80\%/20\%) %& GLCM & SVM (50.69\% \& 50.57\%), DT (72.23\% \& 72.02\%), LR (80.99\% \& 80.88\%) \& K-NN (87.86\% \& 88.06\%) \footnotemark[2]  \\	 	&  &  & &  & & None 
										& \textbf{ResNet34 (99.40\% \& 96.51\%)}\footnotemark[2]  \n 
										\cite{Singh_2020} & 2020 & PlantDoc (Cropped)  & 28 & 80\%/20\%  & VGG-16 (60.41\%), InceptionV3 (62.06\%) \& \textbf{InceptionResNet V2 (70.53\%)}    \n
										\cite{9408806} & 2021 & Plant Village (Modified) & 61 & 31718/4540 & \textbf{Stacking Model (87\%)}, ResNet (82.78\%), InceptionNet (82.22\%), DenseNet (83.44\%)\& InceptionResNet (84.07\%) \n	
										\cite{9250885} & 2020 & Plant Village (Whole)  & 38 & N/A  & \textbf{Hybrid (AlexNet + Linear SVM) Model (99.98\%)}, Basic AlexNet (96.34\%) \& AlexNet with GAP Layer (97.29\%) \n
										\cite{9342729} & 2020 & Plant Village (Peach, Pepper \& Strawberry) & 6 & 70\%/30\%  & Multi Convolutional Layered-based CNN (87.47\% - 99.25\%) with different epochs (50, 75,100 \& 125) \n

										\cite{9412643} & 2021 & Plant Village + Digipathos \cite{8444395} + Web images  & 1146\footnotemark[3]  & 10324 (80\%/20\%)  & InceptionV3 \& Conditional Multi-task Learning (CMTL), Total Top-1 Accuracy: 69.43\%, Total Avg Accuracy: 64.79, Disease Top\_1: 82.96\%, Species Top-1: 78.64\% \\
										
										% \hline
										% \hline
										\bottomrule
									\end{tabular} % 
									\footnotetext[1]{Training/Validation/Test Amount}
									\footnotetext[2]{Accuracy \& F1-score}
									\footnotetext[3]{311 species, 289 diseases}
									
								}
								
							\end{minipage}
						\end{center}
					\end{table}
					% \clearpage
					% \restoregeometry
					
					The above studies are all based on multi-class classification tasks, either directly or indirectly through the use of power-set labelling. Recently, researchers have also begun to adopt multi-task learning methods (different from power-set) to directly address the classification of both species and diseases. For example, in \cite{9412643} the authors proposed conditional multi-task learning approach (CMTL) with InceptionV3 as a backbone CNN to predict both leaf species and diseases. In addition, the paper showed that it is possible to use the predicted results of plant species to help improve the prediction of disease. The dataset for the evaluation of CMTL consists of Plant Village, Digipathos \cite{8444395} and Web images which made up to 12,290 leaf images with 1146 joint species-disease labels (311 species  \& 289 diseases). The experiment showed that, the total Top-1 accuracy for joint prediction of species-disease labels is 69.43\%, the total average accuracy is 64.79\%, the disease’s Top-1 accuracy is 82.96\%, and the species’ Top-1 accuracy is 78.64\%. Although power-set multi-label CNNs and CMTL are promising, there are many other approaches for multi-prediction that have not been deeply explored, which can be beneficial for plant identification and disease classification.
					
					In this paper, we survey, implement, and evaluate a wide range of multi-prediction approaches with different CNN backbones that can be employed for predicting both plant species and diseases. We also proposed a new deep learning architecture with a learning strategy to improve prediction performance.

					%%%%%%%%%%%%%%%%
					%The above studies are all multi-class classification tasks, \cite{9412643} noticed the effectiveness of the multi-task learning mechanism and tried conditional multi-task learning for simultaneous detection and classification of plant species and diseases. Unlike other studies focused on large-scale multi-task learning, they tried to use a multi-task learning approach to predict both leaf species and diseases (interrelated labels), in addition, they used predicted species results to help disease prediction.
					
					%The above studies are all multi-class classification tasks.  Unlike other studies focused on large-scale multi-task learning, \cite{9412643} tried to use a conditional multi-task learning approach (CMTL) with InceptionV3 to predict both leaf species and diseases (joint species-disease labels), in addition, they used predicted species results to help disease prediction. The proposed model would calculate the Top-1 accuracy of disease, species and joint species-disease labels, respectively. The dataset they used consists of Plant Village, Digipathos \cite{8444395} and Web images which are 12,290 leaf images containing 1146 joint species-disease labels (311 species  \& 289 diseases). The total Top-1 accuracy of joint species-disease labels is 69.43\%, the total average accuracy is 64.79\%, the disease’s Top-1 is 82.96\% and the species’ Top-1 is 78.64\%.	
					
					\section{Methodology}
					\label{sec:methodology}

					\subsection{Datasets}
					\label{Datasets}
					% \begin{wraptable}{i}{0.7\textwidth}
						% \vskip -1.5cm
						\begin{table}[h!]
							\begin{center}
								% \begin{minipage}{\textwidth}
									\caption{Public Leaf Disease Datasets.}\label{Public_Dataset_Address}%
									\vskip -0.3cm
									\resizebox{0.8\textwidth}{!}{%
										\begin{tabular}{@{}llcccl@{}}
											\toprule
											ID & Dataset & Year & Species  & Disease & Link \\
											\midrule
											1 & Plant Village   & 2016 & 14 & 22 & \url{https://data.mendeley.com/datasets/tywbtsjrjv/1}\\
											2& Plant Leaves & 2019 & 12 & 22 & \url{https://data.mendeley.com/datasets/hb74ynkjcn/1} \\
											3 & Plantae\_k &  2019 & 8 & 9  & \url{https://data.mendeley.com/datasets/t6j2h22jpx/1}  \\
											4 & PlantDoc & 2020 & 13 & 17 & \url{https://github.com/pratikkayal/PlantDoc-Dataset} \\
											5 & Plant Pathology 2021 - FGVC8  & 2021 & 1 & 6 & \url{https://www.kaggle.com/c/plant-pathology-2021-fgvc8/overview}  \\
											6 & Maize Leaf (NLB) & 2018 & 1 & 2 & \url{https://osf.io/p67rz/}\\
											7 & Citrus Leaves & 2019 & 1 & 5 &  \url{https://data.mendeley.com/datasets/3f83gxmv57/2}\\
											8 & Rice Diseases Image Dataset & 2019 & 1 & 4 & \url{https://www.kaggle.com/minhhuy2810/rice-diseases-image-dataset} \\
											5 & JMuBEN (Arabica Coffee Leaf Images)  & 2021 & 1 & 3 & \url{https://data.mendeley.com/datasets/t2r6rszp5c/1} \\
											
											6 &JMuBEN2  & 2021 & 1 & 2 &  \url{https://data.mendeley.com/datasets/tgv3zb82nd/1} \\
											7 &Cassava Diseases & 2019 & 1 & 5 & \url{https://www.kaggle.com/c/cassava-disease/data}  \\
											8 & UCI Rice Leaf Diseases & 2017 & 1 & 3 &  \url{https://archive.ics.uci.edu/ml/datasets/Rice+Leaf+Diseases} \\
											\bottomrule
										\end{tabular} %
									}
									% \end{minipage}
							\end{center}
							\vskip -0.3cm
						\end{table}
						% \end{wraptable}
					Data has a central role in modern AI technologies, including machine learning, deep learning, and computer vision. In this study, data is also necessary for the comparison of different methods. This section aims to survey and then select suitable datasets for the benchmarking in the next step of experiment and testing. Different from previous studies where, in most cases, only one dataset is used, in this paper, we select three data sources to make four evaluation sets to provide a comprehensive comparison of different approaches. The role of image datasets for computer vision in plant pathology is clearly important. In \cite{edssjs.15D0A2D220200101}, the authors showed that the foremost problem most researchers in this field have been facing is the lack of available data sets. This would greatly affect and restrict the research of machine learning for plant identification and disease classification from leaf images. Fortunately, in recent years, several attempts have been made successfully and researchers have devoted themselves to the collection of plant disease data, filling the data availability gap in this area. Table \ref{Public_Dataset_Address} shows recent available public datasets about plant leaf diseases for computer vision research. In the table, “Year” denotes the published year; “Species” denotes the number of plant species in the data; “Disease” denotes the number of diseases available in a dataset, because different plants may have different sets of diseases. In the experiment, we select three datasets and  we resize the images in  those datasets to an appropriate input shape for a model, for example, the input size for CNN and AlexNet will be $256 \times 256$.
					
					In Table \ref{Public_Dataset_Address}, the datasets can be divided into two groups, multi-prediction datasets (Plant Village, Plant Leaves, PlantDoc \& Plant Pathology 2021) and single-prediction datasets (Maize Leaf, Rice Diseases Image, JMuBEN, Cassava Diseases \& UCI Rice Leaf). A multi-prediction dataset has different plant species (as shown in the Species column in Table \ref{Public_Dataset_Address}) and different types of diseases (as shown in the "Disease" column in Table \ref{Public_Dataset_Address}). These datasets can be useful for both species and disease classification which will be employed this study to explore the usefulness of multi-prediction approaches and to verify our hypothesis. A single-prediction dataset normally only has a single plant species and a set of disease types for that plant. Therefore, the selected benchmark datasets for our study are detailed as follows:
					
					\subsubsection{Plant Village Dataset}
					% \begin{wrapfigure}{r}{0.4\textwidth}
						% % \vskip -3.cm
						%   \begin{center}
							%     \includegraphics[width=0.4\textwidth]{class_distribution/plant_village_SD.png}
							%   \end{center}
						%   \caption{Relationships Between Species And Disease (Plant Village)}\label{fig:plant_village_SD}
						%   % \vskip -5.cm
						% \end{wrapfigure}
					
\begin{figure}[h!]
						\centering
						\begin{subfigure}{0.33\textwidth}
							\centering
							\includegraphics[width=\textwidth]{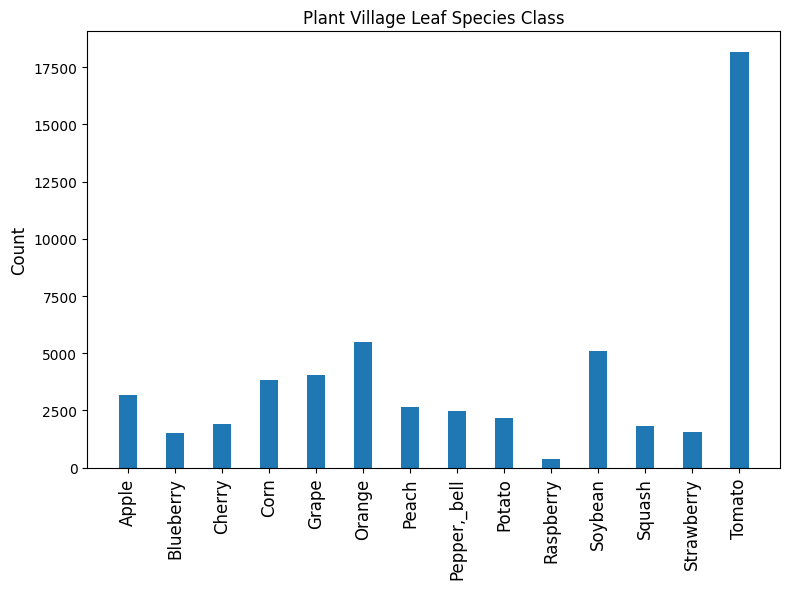}
							\caption{Plant Village}
							\label{fig:plant_village_LeafSpeciesClass}
						\end{subfigure}
						\begin{subfigure}{0.33\textwidth}
							\centering
							\includegraphics[width=\textwidth]{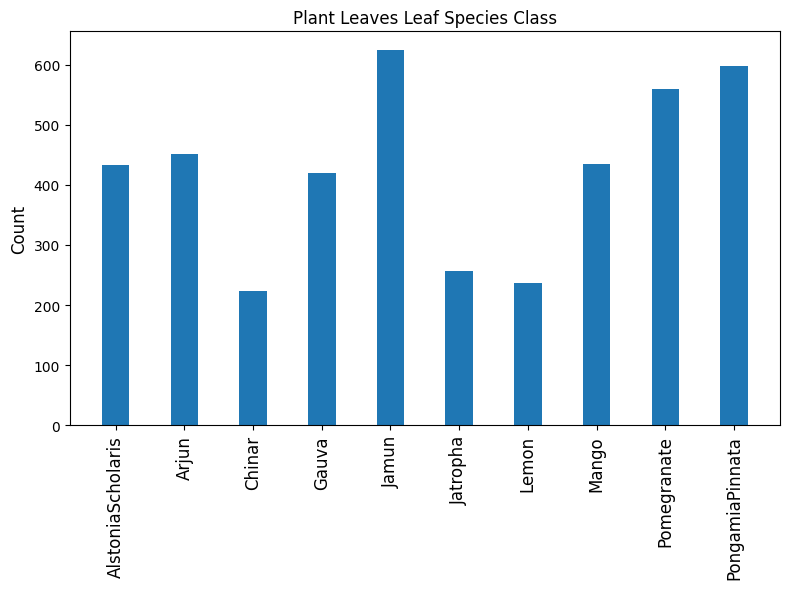}
							\centering
							\caption{Plant Leaves}
							\label{fig:plant_leaves_LeafSpeciesClass}
						\end{subfigure}
						\begin{subfigure}{0.33\textwidth}
							\centering
							\includegraphics[width=\textwidth]{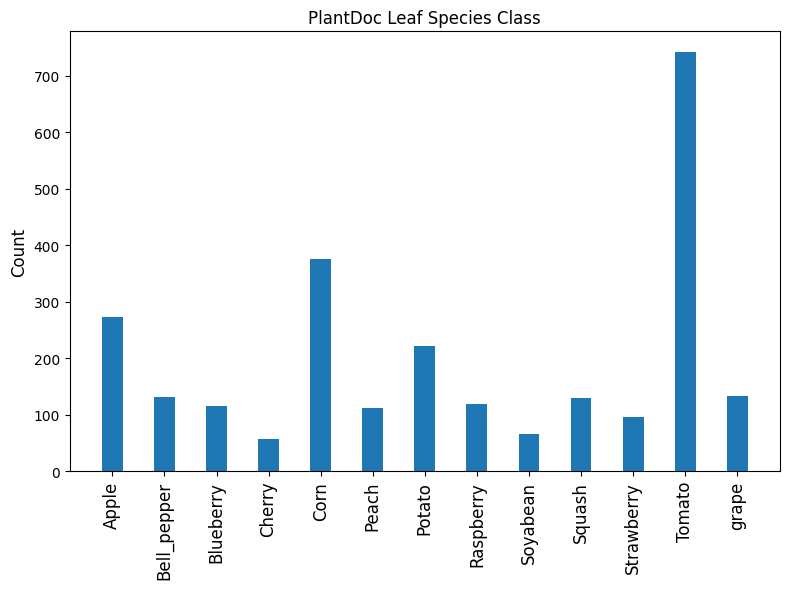}
							\caption{PlantDoc}
							\label{fig:PlantDoc_LeafSpeciesClass}
						\end{subfigure}
						\vskip -0.3cm   
						\caption{\ST{Class Distribution for Plant Species.}}
						\label{fig:ClassDistribution(Species)}
						\vskip -0.3cm   
					\end{figure}

\begin{figure}[h!]
						\centering
						\begin{subfigure}{0.33\textwidth}
							\centering
							\includegraphics[width=\textwidth]{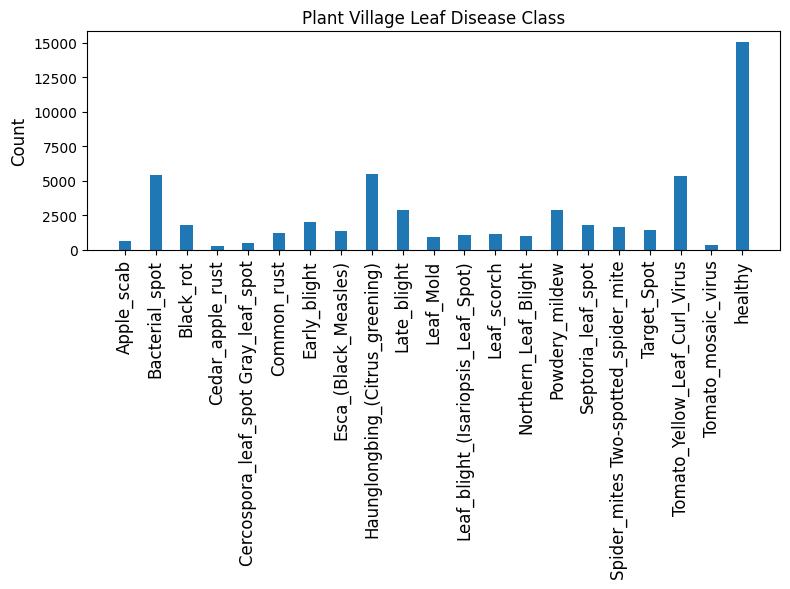}
							\caption{Plant Village}
							\label{fig:plant_village_LeafDiseaseClass}
						\end{subfigure}
						\begin{subfigure}{0.33\textwidth}
							\centering
							\includegraphics[width=\textwidth]{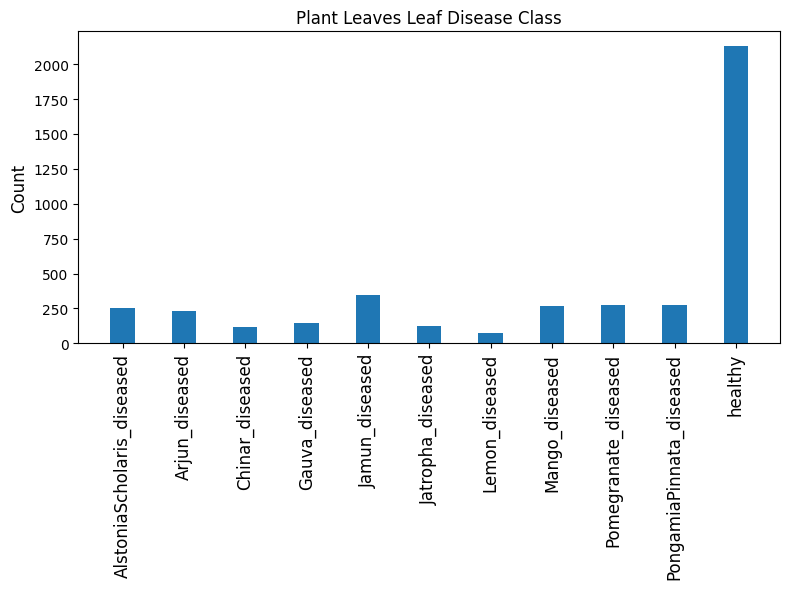}
							\centering
							\caption{Plant Leaves}
							\label{fig:plant_leaves_LeafDiseaseClass}
						\end{subfigure}
						\begin{subfigure}{0.33\textwidth}
							\centering
							\includegraphics[width=\textwidth]{{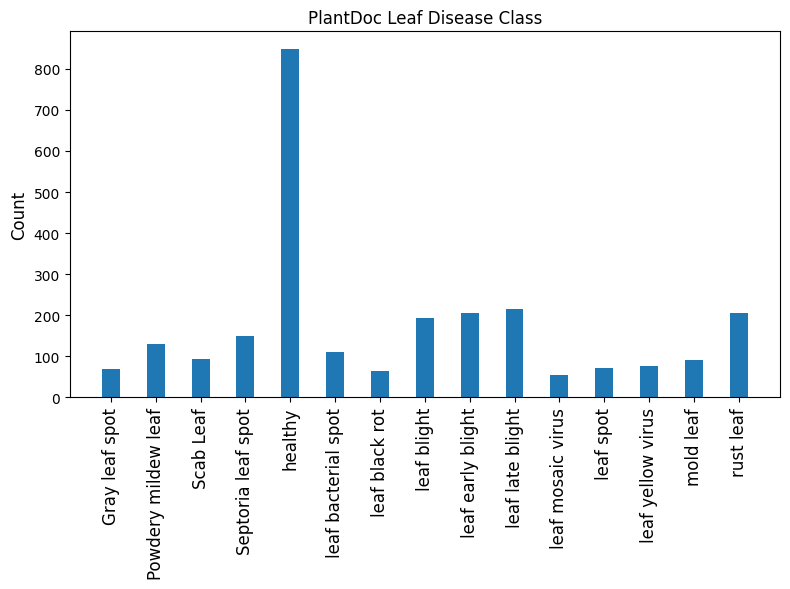}}
							\caption{PlantDoc}
							\label{fig:PlantDoc_LeafDiseaseClass}
						\end{subfigure}
						\vskip -0.3cm   
						\caption{\ST{Class Distribution for Disease Types.}}
						\label{ClassDistribution(Disease)}
						\vskip -0.3cm   
					\end{figure}

\begin{figure}[h!]
						\centering
						\begin{subfigure}{0.33\textwidth}
							\centering
							\includegraphics[width=\textwidth]{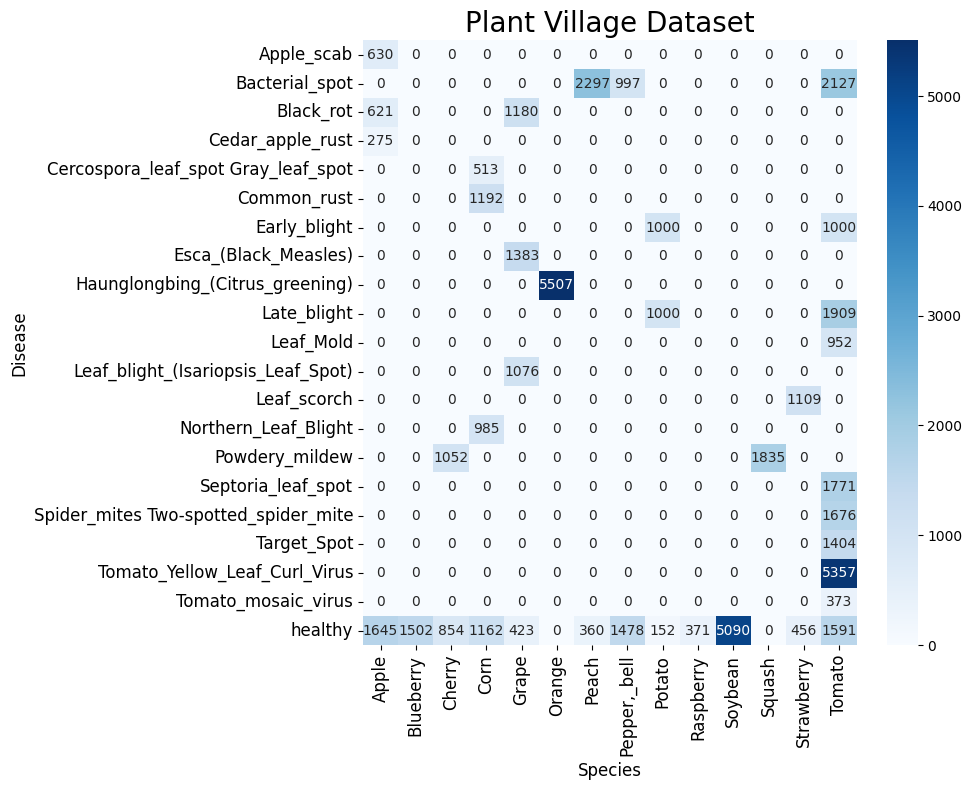}
							\caption{Plant Village}
							\label{fig:plant_village_SD}
						\end{subfigure}
						\begin{subfigure}{0.33\textwidth}
							\centering
							\includegraphics[width=\textwidth]{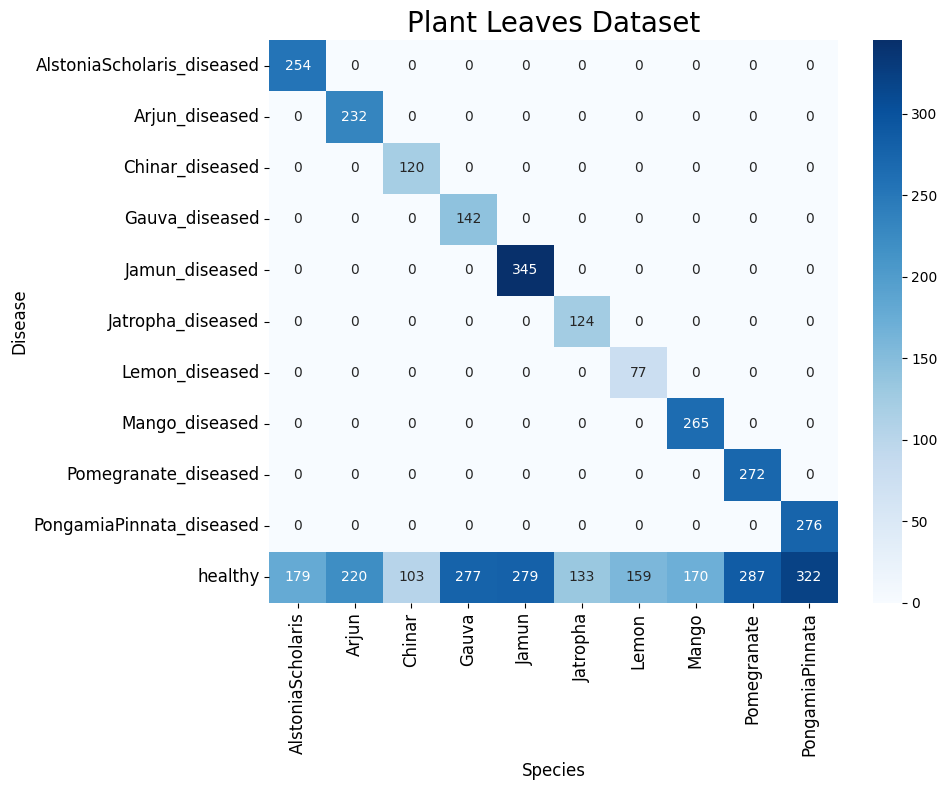}
							\centering
							\caption{Plant Leaves}
							\label{fig:plant_leaves_SD}
						\end{subfigure}
						\begin{subfigure}{0.33\textwidth}
							\centering
							\includegraphics[width=\textwidth]{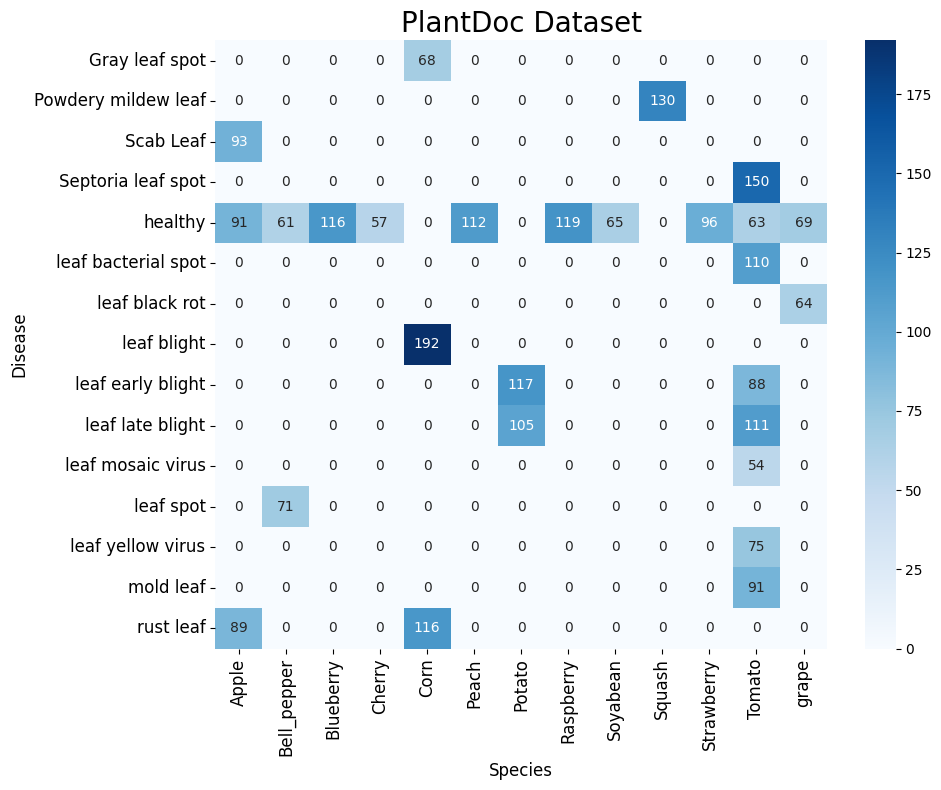}
							\caption{PlantDoc}
							\label{fig:PlantDoc_SD}
						\end{subfigure}
						\vskip -0.3cm   
						\caption{\ST{Relationships Between Species and Diseases.}}
						\label{fig:RelationshipsBetweenSpeciesAndDisease}
						\vskip -0.3cm   
					\end{figure}

					Plant Village dataset is currently one of the most widely used public datasets for research on leaf disease identification and classification. It has different versions, including an original version and a data augmentation version. The original dataset was published in 2016 \cite{hughes2016open}, it consists of $54,305$ images of diseased leaves and healthy leaves from 14 plant species (Apple, Blueberry, Cherry, Corn, Grape, Orange, Peach, Bell Pepper, Potato, Raspberry, Soybean, Squash, Strawberry \& Tomato). Each species has 1 to 10 classes of related diseases, resulting in 22 unique disease categories totally with some species sharing several diseases. In this dataset, there is a total of 38 unique combinations of species and diseases (e.g. Apple Black Rot), and one additional category about images without leaf ($1,143$ background images). The data augmentation version was released in 2019 \cite{DBLP:journals/corr/HughesS15}. In this version, the creators have applied six augmentation methods to enrich the data, including image flipping, Gamma correction, noise injection, principle component analysis (PCA) colour augmentation, rotation, and scaling, to improve the quality and quantity of the data. As the result, the number of samples in this dataset is $61,486$, which increased from $54,305$ in the original version. In our study, we carry out the experiment on the original version. We split the data into $70\%$-$10\%$-$20\%$ for training, validation, and test sets respectively.
					\ST{From Figures \ref{fig:plant_village_LeafSpeciesClass} and \ref{fig:plant_village_LeafDiseaseClass}, we can see the class distribution of Plant Village clearly, because Tomato has 9 groups of diseases and 1 group of healthy, it has the most number of pictures. Figure \ref{fig:plant_village_SD} shows the relationships between species and disease of Plant Village.}
					
					% \begin{figure}[h!]
						% 	\centering
						% 	\begin{subfigure}{0.4\textwidth}
							% %		\vskip 0pt
							% 		\centering
							% 		\includegraphics[width=\textwidth]{class_distribution/plant_village_LeafSpeciesClass.png}
							% 		\caption{}
							% 		\label{fig:plant_village_LeafSpeciesClass}
							% 	\end{subfigure}
						% 	\begin{subfigure}{0.4\textwidth}
							% %		\vskip 0pt
							% 		\centering
							% 		\includegraphics[width=\textwidth]{class_distribution/plant_village_LeafDiseaseClass.png}
							% %		\centering
							% 		\caption{}
							% 		\label{fig:plant_village_LeafDiseaseClass}
							% 	\end{subfigure}
						% 	\caption{Class Distribution of Plant Village}
						% 	\label{fig:ClassDistributionofPlant Village}
						% \end{figure}
					% \begin{figure}[h!]
						% 	\centering
						% 	\begin{subfigure}{0.4\textwidth}
							% 		%		\vskip 0pt
							% 		\centering
							% 		\includegraphics[width=\textwidth]{class_distribution/plant_leaves_LeafSpeciesClass.png}
							% 		\caption{}
							% 		\label{fig:plant_leaves_LeafSpeciesClass}
							% 	\end{subfigure}
						% 	\begin{subfigure}{0.4\textwidth}
							% 		%		\vskip 0pt
							% 		\centering
							% 		\includegraphics[width=\textwidth]{class_distribution/plant_leaves_LeafDiseaseClass.png}
							% 		%		\centering
							% 		\caption{}
							% 		\label{fig:plant_leaves_LeafDiseaseClass}
							% 	\end{subfigure}
						% 	\caption{Class Distribution of Plant Leaves}
						% 	\label{fig:ClassDistributionofPlantLeaves}
						% \end{figure}
					
					%%%%%%%%%%%%%%%%%%%%%%%%%%%%%%%%%%%%
					
					%%%%%%%%%%%%%%%%%%%%%%%%%%%%%%%%%%%%%%%%%%%%%%%%%%%%%%%%%%

					\subsubsection{Plant Leaves Dataset}
					% \begin{wrapfigure}{r}{0.4\textwidth}[h!]
						% 	\vskip -1.5cm
						% 	\centering
						% 	\includegraphics[width=0.4\textwidth]{class_distribution/plant_leaves_SD.png}
						% 	\caption{Relationships Between Species And Disease (Plant Leaves)}
						% 	\label{fig:plant_leaves_SD}
						% 	\vskip -0.5cm
						% \end{wrapfigure}
					Plant Leaves dataset consists of $4,502$ images of healthy and unhealthy leaves divided into 22 categories by species and their health condition. The images are in high-resolution JPG format. The dataset has 12 plant species:   AlstoniaScholaris, Arjun, Bael, Basil, Chinar, Gauva, Jamun, Jatropha, Lemon, Mango, Pomegranate, and PongamiaPinnata. We partition the data samples into three different sets with $70\%$ for training, $10\%$ for validation and $20\%$ for testing. \ST{Figures \ref{fig:plant_leaves_LeafSpeciesClass} and \ref{fig:plant_leaves_LeafDiseaseClass} show the class distribution of Plant Leaves, because each species has 1 group of disease and 1 group of healthy, the healthy category has the most number of pictures. The relationships between species and disease of Plant Leaves have been shown in Figure \ref{fig:plant_leaves_SD}.}
					
					% \begin{figure}[h!]
						% 	\centering
						% 	\includegraphics[width=1\textwidth]{class_distribution/plant_leaves_SD.png}
						% 	\caption{Relationships Between Species And Disease (Plant Leaves)}
						% 	\label{fig:plant_leaves_SD}
						% \end{figure}

					\subsubsection{PlantDoc Dataset}
					% \begin{wrapfigure}{r}{0.3\textwidth}
						% 	\centering
						%  \begin{minipage}{\textwidth}
							% 	\begin{subfigure}{0.15\textwidth}
								% 		\centering
								% 		\includegraphics[width=\textwidth]{plant_village_apple_scab.JPG}
								% 		\caption{Plant Village}
								% 		\label{fig:plant_village_apple_scab}
								% 	\end{subfigure}
							% 	\begin{subfigure}{0.15\textwidth}
								% 		\centering
								% 		\includegraphics[width=\textwidth]{PlantDoc_apple_scab.jpg}
								% 		% \centering
								% 		\caption{PlantDoc}
								% 		\label{fig:PlantDoc_apple_scab}
								% 	\end{subfigure}
							%  \end{minipage}
						% 	\caption{Apple Scab Leaf Samples}
						% 	\label{fig:Apple_Scab_Leaf}
						% \end{wrapfigure}

					Compared to Plant Village Dataset, PlantDoc dataset aims to establish a challenging benchmark with real-field images. The images in Plant Village %(as shown in Figure \ref{fig:plant_village_apple_scab}) 
					were taken in a laboratory setup and not in real conditions of cultivation fields that impact the trained model’s efficacy in practice \cite{Singh_2020}. The usefulness of Plant Village may be not fully potential for the development of applications to identify real-world leaf diseases. PlantDoc is a large-scale non-lab data set for leaf disease detection. The images of PlantDoc have cluttered and diverse backgrounds. %, as shown in Figure \ref{fig:PlantDoc_apple_scab}. 
					It has similar categories of plant species and disease types as Plant Village with $2,598$ leaf images, 13 plant species, and 17 unique diseases. There are 38 classes for a combination of species and diseases (e.g., Apple Scab Leaf). Originally, the data was partitioned into a training set of $2,360$ samples and a small test set of  $238$ samples. We refer to this set as PlantDoc-1.0. To facilitate deeper comparison (with other works and for future study) we re-partition the data to create another dataset from PlantDoc. We mixed and shuffled the whole dataset and split it into $70\%$-$10\%$-$20\%$ for training, validation, and testing respectively. This data is referred to as PlantDoc-0.2. We use these two versions of PlantDoc in this study. \ST{Figures \ref{fig:PlantDoc_LeafSpeciesClass} and \ref{fig:PlantDoc_LeafDiseaseClass} show PlantDoc’s class distribution, because Tomato has 7 groups of diseases and 1 group of healthy, it has the most number of leaf images. Figure \ref{fig:PlantDoc_SD} shows the relationships between species and disease of PlantDoc.}
					
					% \begin{figure}[h!]
						% 	\centering
						% 	\begin{subfigure}{0.4\textwidth}
							% 		%		\vskip 0pt
							% 		\centering
							% 		\includegraphics[width=\textwidth]{class_distribution/PlantDoc_LeafSpeciesClass.png}
							% 		\caption{}
							% 		\label{fig:PlantDoc_LeafSpeciesClass}
							% 	\end{subfigure}
						% 	\begin{subfigure}{0.4\textwidth}
							% 		%		\vskip 0pt
							% 		\centering
							% 		\includegraphics[width=\textwidth]{class_distribution/PlantDoc_LeafDiseaseClass.png}
							% 		%		\centering
							% 		\caption{}
							% 		\label{fig:PlantDoc_LeafDiseaseClass}
							% 	\end{subfigure}
						% 	\caption{Class Distribution of PlantDoc}
						% 	\label{fig:ClassDistributionofPlantDoc}
						%  \vskip -0.5cm
						% \end{figure}

					% \begin{wrapfigure}{i}{0.4\textwidth}
						% 	\vskip -1.5cm
						% 	\centering
						% 	\includegraphics[width=0.4\textwidth]{class_distribution/PlantDoc_SD.png}
						%  \vskip -0.3cm
						% 	\caption{Relationships Between Species And Disease (PlantDoc)}
						% 	\label{fig:PlantDoc_SD}
						% 	\vskip -0.8cm
						% \end{wrapfigure}
					
					\subsection{Models}
					\label{Models}
					In this section we will survey different deep learning approaches which have been or can be applied for plant identification and disease classification. The current deep learning models employed for plant identification or disease classification are CNN models we  will describe in Section \ref{sec:backbones}. However, most of the other approaches we present below have not been applied largely to plant pathology, although they are really relevant and already exist in machine learning literature.
					\subsubsection{Backbone CNNs}
					\label{sec:backbones}
					In recent years, many different architectures were designed based on Convolutional Neural Networks (CNN) to deal with spatial data such as images and videos, especially in computer vision tasks. With their flexible and computationally efficient architectures, CNNs can be adapted to different scenarios and tasks. In what follows, we re-introduce several CNN models, which are popular and have been proven with excellent performance on image classification tasks. They were tested on benchmark datasets such as CIFAR-100 and ImageNet, and also are the state-of-the-art approaches for plant identification and for disease classification in recent research, as shown in Table \ref{DL_Comparison}.

					\textbf{Convolutional Neural Networks (CNN).} 
					CNNs refer to a class of neural networks that employ convolutional operators for information propagation from layers to layers. The convolutional operation is useful for image analysis as it helps neural networks learn local features, which makes CNNs popular for image data \cite{726791}.  This is also the foundation for a series of deep neural network structures lately. A CNN has an input layer and an output layer, and between these two layers, there are several hidden layers where connections between a lower layer and an upper layer are formed by convolutional operators. The number of hidden layers is chosen depending on the complexity of a task. Recent advanced techniques in CNNs can improve the performance and allow CNNs to be scalable for learning from larger datasets. These include Rectified Linear units (ReLU) and other activation functions, pooling (average and max pooling), normalisation (Batch Norm and Layer Norm), etc. Normally, in a CNN architecture after a series of convolutional layers, there will be several fully-connected layers before the outputs. A fully-connected layer is a normal layer with dense connections (instead of convolutional connections). These layers will process and weave the features from the preceding convolutional layers to make accurate prediction. One of the advantages of CNN is that it can process the raw pixel values from images and learn discriminative features in an end-to-end fashion.
					
					\iffalse
					\begin{wraptable}{i}{0.3\textwidth}
						\vskip -2.5cm
						\begin{center}
							%\begin{minipage}{\textwidth}
							\caption{Our CNN Model Structure}\label{CNN_Model_Structure}%
							\resizebox{0.3\textwidth}{!}{%
								\begin{tabular}{@{}l|c|c@{}}
									\toprule
									Layer & Filter & Kernel Size\\
									\hline
									Conv2D& 32 & 3x3 \\
									Batch Normalization& & \\
									Conv2D& 32 & 3x3 \\
									Batch Normalization& & \\
									MaxPool2D& & 8x8 \ \n
									Conv2D& 32 & 3x3 \\
									Batch Normalization& & \\
									Conv2D& 32 & 3x3 \\
									Batch Normalization& & \\
									MaxPool2D& & 8x8 \n
									Flatten & & \\
									% 	\begin{tabular}{@{}l@{}} Linear (Plant Branch) \\ Linear (Disease Bracnch)  \end{tabular}& & \begin{tabular}{@{}c@{}} Plant Amount  \\ Disease Amount  \end{tabular}\\
									Softmax & & \# of classes \\
									\bottomrule
								\end{tabular} %
							}
							%	\end{minipage}
					\end{center}
					% \vskip -0.5cm
				\end{wraptable}
				\fi
				
				Based on our study on neural networks, we design a custom CNN which can be applied to plant identification or leaf disease classification. \cmt{The structure of our CNN is similar to \cite{8871173}. However, different from it, we apply our structure to image data}. The input size of the our CNN is set as $256\times 256\times 3$ and it has 4 convolutional layers in total. There is a batch normalization layer that can normalize the inputs after each convolutional layer. After 2 convolutional layers and 2 batch normalization layers, we place a max-pooling layer (this combination is repeated twice). The activation functions for all units are ReLU. On top of these layers, depending on various tasks, we add output layers to perform prediction. The units in these layers are constrained together as a softmax group. Besides the custom CNN, in what follows, we will present the most common off-the-self CNNs which have been used for image classification in general and plant pathology in specific.
				
				\textbf{AlexNet.} This CNN architecture starts with five convolutional layers, and there are two max-pooling layers between the first three convolution layers. In the later stage, AlexNet has three fully connected layers. An interesting feature of AlexNet is its activation functions are designed as non-saturating ReLU. AlexNet was one of the early CNN models that made a breakthrough in image classification, notably being the first CNN to win the  ImageNet challenge in 2012.
				
				\textbf{VGG16}. This CNN architecture has 16 layers with multiple $3\times 3$ kernel-size filters for the convolution. This is different from the first and second large kernel-sized filters in AlexNet. VGG was designed to increase the depth of CNNs where it has several max-pooling layers. In VGG16, there are three large fully connected layers, one with $4,096$ units and another with $1,000$ units in the later stage of its architecture. In image classification, VGG16 achieved 92.7\% top-5 test accuracy in the 2014 ImageNet challenge.
				
				\textbf{ResNet101}. This is a powerful structure where we can design and train the model with a lot of layers to gain performance superiority. The key component of ResNet is its "skip connections" which will skip one or several layers before rejoining to connect to the following layer. This idea can help mitigate the vanishing gradient issue or to deal with the degradation issue. ResNet can help reduce the training error when adding more layers to the CNN, hence providing a good structure for scalable learning \cite{7780459}. ResNet was the winner of ILSVRC 2015 challenge (a subset of ImageNet).
				
				\textbf{InceptionV3}. Inception is a class of CNNs that utilises Inception modules for deeper structure with more efficient computation. The motivation of Inception is to prevent the number of parameters from being too large while building deeper neural networks \cite{7298594}. Inception consists of asymmetric and symmetric construction blocks. Different layers are employed, including convolution layers, average and max-pooling layers, concatenate layers, dropout layers and fully connected layers. Each Inception module in this architecture consists of four operations in parallel. The modules will be linked by concatenate layers. The batch Normalization method has been applied to the output of convolutional layers and is widely used in the whole model. In this study, we use the most popular version of Inception, i.e., InceptionV3.
				
				\textbf{MobileNetV2}. MobileNet is one of the most popular light-weight CNN architectures. It aims to significantly reduce the size of the parameters and to increase the computational speed while maintaining accuracy. It was designed based on the inverted residual structure. However, different from other residual models, its residual block’s input and output are thin bottleneck layers. Also, the light-weight depthwise convolution operator is used in its intermediate expansion layer to reduce the number of parameters \cite{Sandler_2018_CVPR}. Interestingly enough, the lightweight depthwise convolution can not only reduce the complexity of the model, i.e. size of the model, but also greatly reduce computational cost. MobileNet, therefore, is popular for low-resource devices, especially for mobile devices. In this paper, we use MobileNetV2.
				
				\textbf{EfficientNet}. Similar to MobileNet, this is one of the light-weight CNN architectures. EfficientNet is based on a scaling approach which employs fixed and compound coefficients to scale all depth/ width/ resolution dimensions. In this study, we employ EfficientNet with two core parts, one is inverted bottleneck residual blocks (adopted from MobileNetV2) and the other is Squeeze-and-Excitation blocks (SENet). EfffientNet achieved 77.3\% top-1 accuracy in the ImageNet dataset.
				
				\textbf{Vision Transformer (ViT)}.
				\ST{Vision Transformer (ViT) model \cite{dosovitskiy2021an} was released in 2021, based on the idea of Transformers \cite{NIPS2017_3f5ee243} developed from the natural language processing. ViTs can handle a wide range of image sizes  without requiring architectural changes. ViTs have a mechanism to capture global context information. They can attend to all image patches simultaneously, allowing them to understand the relationships between distant parts of an image. ViTs leverage the self-attention mechanism, which can capture complex relationships between image patche. They have shown advantages over CNNs in several tasks, for example ViTs perform better than ResNet in image classification on ImageNet and CIFAR-10 \cite{dosovitskiy2021an}.}
				
				We have presented popular CNN models used in this study. Those  models can work alone to predict plant species or disease types, or they can be the backbone in a multi-prediction model to  predict these two labels simultaneously, as shown in what follows.
				\subsubsection{Multi-model CNNs}
\begin{figure}[ht]
					%\vskip -0.5cm
					\centering
					% \begin{figure}[h!]
						\centering
						\includegraphics[width=0.85\textwidth]{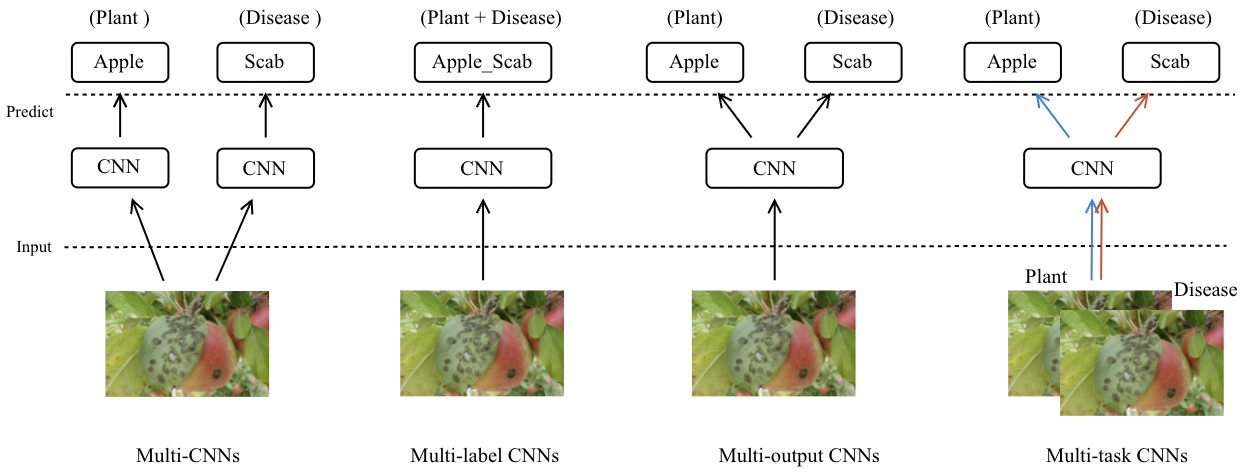}
						\vskip -0.3cm
						\caption{CNNs Approaches for Plant Identification and Disease Classification.}
						\label{fig:3_classes_CNN}
						% \end{figure}
					\vskip -0.5cm
				\end{figure}

				This is the most straightforward application of CNNs for multi-prediction. A deep learning model can consist of two independent CNNs, each for a task. In this study we use two independent CNNs, one for predicting plant species and the other for predicting disease types, as shown in  Figure \ref{fig:3_classes_CNN}. The two CNNs have the same architecture but each has different set of parameters.
				\subsubsection{Multi-label (power-set) CNNs}
				The second approach for multi-prediction is multi-label where the two tasks (plant prediction and disease prediction) are encoded in a single output layer. The most feasible way for it is to join the labels, as known as power-set labelling. This would help transfer a multi-label task to a multi-class task where we can directly apply the backbone CNNs above. In particular, the plant species label and the disease type label will be combined, making a joint label representing both plants and diseases. For example, a power-set label “apple\_scab” can be created from the plant label “apple” and the disease label “scab”. This is the most common method for plant identification and disease classification in the literature. However, our paper is the first to apply and compare different backbones CNNs. Despite being simple, this multi-label approach has a scalable issue when facing a large number of classes for each label. In the worst case, the power-set label will consist of $|\mathcal{P}|\times|\mathcal{D}|$ classes, where $|\mathcal{P}|$ is the number of plants and $|\mathcal{D}|$ is the number of diseases. It may lead to the growth in computational complexity.

				%	In this study, we explored three common types of multi-output deep learning approaches for leaf disease classification (see Figure \ref{fig:3_classes_CNN}), i.e, Multi-model, Multi-Label and Multi-task.
				
				%\begin{wrapfigure}{i}{0.6\textwidth}
				%\end{wrapfigure}
				
				\subsubsection{Multi-output CNNs} 
				This is a class of CNN models in which we have an input layer and multiple output layers, each for a task as shown in Figure \ref{fig:3_classes_CNN} (third model). Theoretically, compare to the multi-label CNNs, multi-output CNNs have fewer parameters because the latter have fewer connections to the output layer(s). The number of total output units in multi-output CNNs for plant \& diseaes prediction is $|\mathcal{P}| + |\mathcal{D}|$. The learning in these CNNs is done by optimising the model for all tasks simultaneously. This is also different from the multi-task learning we will discuss below.
				
				\subsubsection{Multi-task Deep Learning}
				Finally, we can adapt multi-task deep learning architectures for leaf disease and plant type classification. Originally, a multi-task learning problem is to learn a model from different (related) domains, where each task $i$ is associate a dataset $\mathcal{D}_{i}=\{(x_i^{(n)},y_i^{(n)}), n=1, ..., N_i\}$. Multi-task learning is very suitable for deep learning models,  the features learned from one task may benefit another task learning \cite{9392366}.  
				We can utilise those structures for plant identification (task 1) and disease classification (task 2) by sharing input data among different tasks, i.e. $x_1^{(n)}=x_2^{(n)}=x^{(n)}$ and $y_1^{(n)}=p^{(n)}$,  $y_2^{(n)}=d^{(n)}$, where $\{(x^{(n)},p^{(n)},d^{(n)})|n=1,...,N\}$ is a plant leaf dataset in our problem statement.
				Although having the same outputs as multi-output CNNs, the learning in multi-task models is different in which for each data point (an image), they only optimise for a task. In this study, instead of directly using the backbone CNNs (as we will need to implement the learning strategies), we employ the state-of-the-art multi-task models, as shown below.
				
				\textbf{Cross-stitch Network}. Cross-stitch is a multi-task approach to learning shared and task-specific representations \cite{Misra_2016_CVPR}. To this end, cross-stitch units are designed with a soft-parameter sharing mechanism. These units integrate the features
				from outputs of multiple networks, each can represent different patterns of the tasks. In other words, they provide soft feature fusions among multiple single-task networks through a linear combination of every layer’s activations. As the result, cross-stitch networks would be able to fuel discriminative features across multiple tasks to improve performance, even with a small number of training examples. In \cite{Vandenhende_2021}, the authors found that we should pre-train each single-task network first before stitching these networks for better performance. In our study, the cross-stitch units have been deployed in the middle and the end of two CNNs, one for plant prediction and the other for disease prediction.
				
				\textbf{Multi-Task Attention Network (MTAN)}. Different from the parallel learning of shared and task-specific features in cross-stitch networks, MTAN will learn shared (global) features first from the images, and then, allow task-specific features to be learned from those global features via soft-attention modules \cite{Liu_2019_CVPR}. MTAN is built upon a single shared network with a global feature pool and associate each task with a task-specific soft-attention module. Compared to cross-stitch networks, as MTAN aims to share a general feature pool among different single-task networks it will not be affected by the scalability issue. However its limitation would be the lack of diversity in task-specific features are they are all produced from the shared pool \cite{Vandenhende_2021}.
				
				\textbf{Task Switching Network (TSN)}. TSN is based on a task-conditional single-encoder-single-decoder architecture which works with one task at a time while switching between the tasks  \cite{10.1007/978-3-319-24574-4_28}. TSN’s decoder is based on a U-Net architecture and its encoder is ResNet-based backbone (ResNet-18). In TSN, different tasks will be switched by a small task embedding network, one task after another. Meanwhile, the single encoder-decoder would pair all parameters for sharing among the tasks.
				
				\textbf{Model-Contrastive Learning (MOON)}. MOON is a recent model from federated learning paradigm that aims to leverage the similarity of different tasks' representations to enhance local training of each task \cite{Li_2021_CVPR}. In \cite{Li_2021_CVPR}, it is shown that the global features are more useful than the local features learned from each task's dataset. Therefore, MOON proposes a contrastive learning strategy to fine-tune the local representations at model level by maximizing the agreement between representations of the
				local model and the global model. The advantages of MOON are its simplicity, effectiveness, and ability to deal with the non-iid data issue. Compared to state-of-the-art approaches, MOON achieves a significant improvement on various image classification tasks.
				
				\begin{wrapfigure}{i}{0.45\textwidth}
					\vskip -1.3cm
					\centering
					\begin{subfigure}{0.45\textwidth}%
						%	\begin{subfigure}{0.4\linewidth}%		
							\centering
							\includegraphics[width=\textwidth]{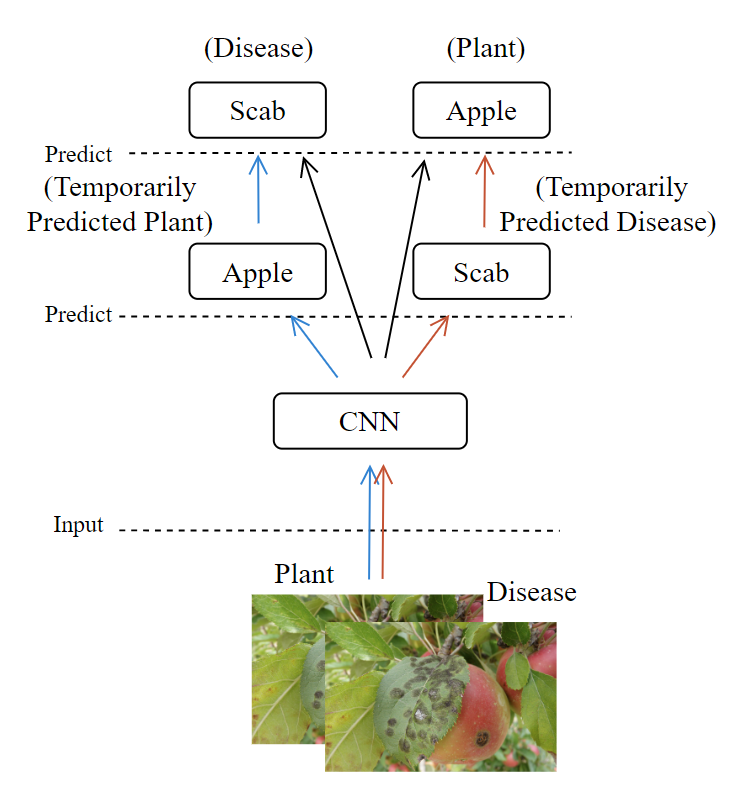}
							% 		\caption{Model Process}
							\label{fig:new_model}
						\end{subfigure}
						\quad
						\vskip -0.8cm
						\caption{Our New Model Structure.}
						\label{fig:gsmo_cnn}
						\vskip -0.8cm
					\end{wrapfigure}
					
					\subsubsection{Our model: Generalised Stacking Multi-output CNNs (GSMo-CNNs)} For completeness, we propose a new model for plant identification and disease classification. The CNN architecture of our model is inspired by the relationship between plant species and disease types. It is commonly known that some diseases may only appear in some particular plants and, therefore, the information about diseases can be useful for the prediction of plants. This reasoning can be applied contrariwise where plant information can be used to predict diseases. We realise this idea by stacking the output (softmax) layers for plant identification and disease classification one on top of another. We generalise the effect of the relationships between plant species and disease types by adding weights on different loss functions at each level of the stack for each output. In what follows, we will show the details of our model.
					
					% \begin{figure*}[h!]%
						% 	\centering
						
						% 	\begin{subfigure}{0.6\textwidth}%
							% %	\begin{subfigure}{0.4\linewidth}%		
								% 		\centering
								% 		\includegraphics[width=0.9\textwidth]{new_model.png}
								% % 		\caption{Model Process}
								% 		\label{fig:new_model}
								% 	\end{subfigure}
							% \quad
							% % 	\begin{subfigure}{0.48\textwidth}%
								% % %	\begin{subfigure}{0.4\linewidth}%
									% % 		\centering
									% % 		\includegraphics[width=0.9\textwidth]{model_structure.png}
									% % 		\centering
									% % 		\caption{Model Structure}
									% % 		\label{fig:model_structure}
									% % 	\end{subfigure}
								% 	\caption{Our New Model Structure.}
								% 	\label{fig:gsmo_cnn}
								% \end{figure*}

							\textbf{Architecture}. The structure of our model is shown in Figure \ref{fig:gsmo_cnn}. The model is based on the multi-output approaches with all convolutional layers that can be reused from the backbone CNNs we presented earlier. The changes here, as we can see, are (1) the split of dense layers for different tasks; and (b) the stacking of prediction layers. The motivation behind the split of layers is we can use the convolutional layers to learn global features while the dense layers will learn task-specific features. For the stacking strategy of prediction layers, GSMo-CNNs will temporarily infer the probability of plant species and the probability of diseases in the first prediction level. After that, we apply cross connection, i.e. we concatenate the probability of predicted plants and the CNN features to make the final prediction of diseases, and similarly,  we concatenate the probability of predicted diseases with the CNN features to make the final prediction of plants.

							In particular, the proposed model has two sets of fully connected layers (called here as "branches") after the flatten layer, each branch will connect to a concatenate layer in the later stage. These two branches will produce a pair of predicted plant and disease results first, named here as $p^{temp}$ (plant temporary) and $d^{temp}$ (disease temporary). We did not make an actual prediction at this level, instead, we extract the prediction probability from the two softmax layers for the next step. At the concatenate layer, the image features from the flatten layer of the CNN will be combined with the features from softmax layers (probability) in the temporary prediction level ($p^{temp}$ or $d^{temp}$, depending on the branch used). These two soft-max layers will connect to two different concatenate layers, where they will join with the shared CNN's flatten layer. On top of each concatenate layer, we have another softmax layer for cross prediction. In other words, our GSMo-CNN uses the prediction probability of plant species (from $p^{temp}$) to predict leaf diseases ($d$) and uses the predicted disease ($d^{temp}$) to predict the plant species ($p$) from leaf images. Note that we can use any of the four output layers ($p$, $d$, $p^{temp}$ and $d^{temp}$ for prediction but we will show in the experiments that by stacking the softmax layers the final output layers would give better performance. The use of probability in the temporary prediction layer (with the $softmax$ function), instead of the predicted values (with the $argmax$ function), will help smooth the propagation of gradients in the learning. Our model is inspired by classifier chain \cite{Read_2009}, but the difference here is that the "chain" in GSMo-CNNs is implemented in a stacking fashion with parallel inference instead of sequential inference.
							
							\textbf{Training}. For training, as we have four outputs, we aim to optimise the prediction at every output layer. This is because a good estimation of plant species in the temporary prediction layer will help improve the prediction of diseases in the final prediction (top) layer. Similarly, a good estimation of diseases in the temporary prediction layer will help the prediction of plants in the final layer. As the result, we will need four loss functions to minimise. Normally, we can train our model to minimise the losses simultaneously to take advantage of the underlying relationships between plant species and leaf disease. This would also allow us to impose the logical negation constraint that some diseases will not appear on specific plants and vice versa. Furthermore, we are interested in how much relationships and constraints affect the learning. However, there will be an issue associated with such relationships and constraints in the case where the temporary layers do not learn well and make the wrong prediction for the top layers. This will definitely happen during the beginning of the training where, after being initialised, the model will have very low performance (and many mistakes will be made) as it just starts to learn. Therefore, in order to control this impact, we associate each loss with a balance weight. In particular, we train our model by minimising the following total loss function:
							\begin{equation}
								\mathcal{C} = \beta_1 \mathcal{L}(P_1,\hat{P}_1) + \beta_2 \mathcal{L}(P_2,\hat{P}_2) + \delta_1 \mathcal{L}(D_1,\hat{D}_1) + \delta_2 \mathcal{L}(D_2,\hat{D}_2)
							\end{equation}
							where $\mathcal{L}$ is a cross-entropy loss function;  $P_1$, $P_2$ are the ground truth for the plant species at level 1 and level 2 in the stack respectively; $P_1$, $P_2$, are the corresponding predicted plant species; $D_1$, $D_2$ are the ground truth for the disease types at level 1 and level 2 in the stack respectively; $D_1$, $D_2$, are the corresponding predicted diseases. $\beta_1$, $\beta_2$, $\delta_1$, $\delta_2$ are the balance weights. In the experiment $\beta_1$, $\beta_2$, $\delta_1$, and $\delta_2$ are treated as hyper-parameters. The reason for using balance weights for the losses in each level is that we want to fine-tune the learning in the upper level based on what we have in the lower level.  We also apply different balance weights to the losses on different outputs to see how they influence each other. We would anticipate that these balance weights would be similar as the prediction of plants would help the prediction of diseases, and vice versa.
							
							\textbf{Inference} Our model performs two stages of inference for plant species and disease types but only the outcomes of the $2^{nd}$ stage will be used for the final prediction. However, by using the balance weights in the training, we generalise the multi-output architecture, allowing control of interactions between different outputs and between outputs in different levels. As the result, the multi-model and multi-output approaches are just special cases of our model, as follows. 
							\vskip -1.5cm
							\begin{itemize}
								\item  If $\beta_1=1$ and $\beta_2=\delta_1 = \delta_2 = 0$ then our model becomes a single CNN for plant identification. In this case the layer for plant identification at the first level will be used as the output.
								\item   If $\delta_1=1$ and $\beta_1=\beta_2 = \delta_2 = 0$ then our model becomes a single CNN for disease classification.  In this case the layer for disease classification at the first level will be used as the output.
								\item If $\beta_1=\delta_1=1$ and $\beta_2$ = $\delta_2$=0 our model becomes a regular multi-output CNNs, as shown earlier. The outputs are the two layers in the first prediction level.
							\end{itemize}
							\vskip -1.5cm
							\subsection{Evaluation Metrics}
							\label{EvaluationMetrics}
							
							Based on the study of related work and classification tasks, we found that Accuracy and F1-score are the most common evaluation metrics for plant identification and disease classification. Both metrics can be calculated from a confusion matrix \cite{edsdoj.8f0d61eb104a67a1e887ad1111fbd220200101}. The accuracy metric indicates how accurate a machine learning model is. A prediction is correct if its output is the same as the actual label (ground truth) and accuracy is the ratio of the number of correct predictions to the total number of samples. Generally, accuracy is a sensible metric for classification tasks, however, when facing the data imbalance issue accuracy will become more biased towards the class with the most number of samples. For example, if 95 out of 100 images are apple leaves, then we can make a simple guess to achieve 95\% accuracy. Although not all datasets are imbalanced and sometimes the imbalance issue is not severe, for completeness, we also employ F1-score as another evaluation metric. Accuracy and F1-score are calculated as follows:
							$ \text{Recall} = \frac{TP}{(TP+FN)}$, $\text{Precision} = \frac{TP}{(TP+FP)}$, $\text{Accuracy} = \frac{(TP+TN)}{(TP+FN+FP+TN)}$, $\text{F1\_score} = \frac{(2 \ast Precision\ast  Recall)}{(Precision+ Recall)}$, 
							$\text{FPR} =  \frac{FP}{(FP+TN)}$.
							
							Here, TP, TN, FP,FN, and \ST{FPR} denote True Positives, True Negatives, False Positives, False Negatives, and \ST{False Positive Rate} respectively. As we can see, F1-score is based on both Recall and Precision. It balances the weights of precision and recall, making it a more reliable metric. In this study, F1-score will be the key evaluation metric, e.g. used for model selection. Together with Accuracy, they will provide a comprehensive view of the performance of a model. We will evaluate a model based on its performance (Accuracy \& F1-score) on plant prediction, disease prediction, and (combined) plant-disease prediction.
							\ST{False Positive Rates will be used to assist in evaluating the optimal models for all approaches. FPR represents the number of false positive predictions, and the sum of FP and TN constitutes the total number of true negatives.}

							\subsection{Setup}
							
							% \begin{wraptable}{i}{5.5cm}
								% 		\begin{center}
									% 			\begin{minipage}{\linewidth}
										%             % \vskip -.5cm
										% 			\caption{Experimental Setup}\label{Experimental_Setup}%
										% 			\resizebox{\textwidth}{!}{%
											% 				\begin{tabular}{@{}ll@{}}
												% 					\toprule
												% 					Items & Values \\
												% 					\midrule
												% 					Batch Size	&16\\					
												% 					Colour Model  & RGB \\
												% 					Early Stopping Monitor 	&Validation Loss\\
												% 					Max Epoch 	&10000\\	
												% 					Input Size	&$256 \times 256 \times 3$\\
												% 					Learning Rate	& [0.01, 0.001, 0.0001]\\
												% 					Balance Weight ( $\beta_1$, $\beta_2$,$\delta_1$, $\delta_2$)	&[0.2,0.4,0.6,0.8] + fine-tuning\\
												% 					Optimizer	&Adamax\\
												% 					\bottomrule
												% 				\end{tabular} %
											% 					}
										% 				\end{minipage}
									% 	\end{center}
								% \end{wraptable}
							
							With the models are ready, we are now in a position to prepare for the empirical study. A  collection of common and recent CNN models for plant species and diseases with various learning approaches (multi-model, multi-label, multi-output, multi-task, and GSMo-CNNs) will be tested.  Backbone models (i.e., CNN, AlexNet, VGG16, ResNet101, EfficientNet, InceptionV3 \& MobileNetV2) are implemented in Tensorflow 2.6, and multi-task models are in Pytorch (downloaded and cited from related authors’ papers and GitHub). We perform the experiments on three public datasets, including Plant Village, Plant Leaves and PlantDoc, as detailed in Section \ref{Datasets}. The ratio of training and test sets is 80: 20 and 10\% of the training set will be used as a validation set. \cmt{All hyper-parameters in the experiments were set uniformly across the models and approaches}. %, as shown in Table \ref{Experimental_Setup}}. 
						During the training, we adopt an early stopping method to avoid the problem of over-fitting. In particular, if a model's validation loss has not been decreasing in 50 epochs we stop the training. The input size of leaf images is set to $256\times 256 \times 3$ pixels and all samples are in RGB format. Our model selection is done by measuring F1-score on the validation set. For GSMo-CNNs, we search the balance weights using a coarse grid [0.2, 0.4, 0.6, 0.8] and then narrow down the search to find the best combination for all datasets. We find that
						$\beta_1=0.1$, $\beta_2=0.4$, $\delta_1=0.1$, $\delta_2=0.5$ are generally good for all three datasets and we report the results of GSMo-CNNs under this configuration. We run each experiment 10 times and report the average accuracy \& F1-score with standard deviation.

						\section{Experimental Results}
						\label{sec:experiment}
						In this section, we report the results of different models on the benchmark datasets and analyse their performance. We divide the results into "Plant Prediction", "Disease Classification", and both (Plant Identification \& Disease Classification). In multi-model, multi-label (power-set), multi-output, and our GSMo-CNNs, we evaluate different CNNs backbones, including AlexNet, VGG16, ResNet101, EfficientNet, InceptionV3, MobileNetV2, and our custom CNN. We test two versions of our GSMo-CNNs, one without balance weights ($\beta_1=\beta_2=\delta_1=\delta_2=1$), and the other with balance weights. To save time for model selection, we choose the best backbone CNN found in the former version (GSMo-CNNs without balance weights) for the latter. For completeness, we apply transfer learning to our GSMo-CNNs to see if improvement can be achieved. We implement the transfer learning idea by using pre-trained backbone CNNs on the ImageNet dataset to initialise our model and fine-tune it on a leaf image dataset.
						\subsection{Plant Identification}
						\label{plant_identification}
						% \begin{figure*}[h!]
							% 	\centering
							% 	\begin{subfigure}{0.25\textwidth}
								% 		\centering
								% 		\includegraphics[width=0.9\textwidth]{plantvillage_apple_health_image__12_.JPG}
								% 		\caption{Apple Leaf (Healthy)}
								% 		\label{fig:plantvillage_apple_health}
								% 	\end{subfigure}
							% 	\begin{subfigure}{0.25\textwidth}
								% 		\centering
								%     \includegraphics[width=0.9\textwidth]{plantvillage_grape_Black_rot_image__8_.JPG}
								% 		\centering
								% 		\caption{Grape Leaf (Black Rot)}
								% 		\label{fig:plantvillage_grape_Black_rot}
								% 	\end{subfigure}
							%     \vskip -.5cm
							% 	\caption{Samples of Plant Prediction}
							% 	\label{fig:Plant_Prediction}
							%  \vskip -.5cm
							% \end{figure*}
						
						Let us discuss the plant identification task first to evaluate the models and approaches on their ability to predict plant species from leaf images. This is a classification task where we identify plant types among a fixed set of categories (species). Different from previous work on identifying plants using healthy leaves only, the challenge of this task in our study is we have both healthy and diseased leaves images. It is worth noting that identifying plant species from corrupted or damaged leaves is non-trivial. %Figure \ref{fig:plantvillage_apple_health} shows a healthy apple leaf from Plant Village dataset and Figure \ref{fig:plantvillage_grape_Black_rot} shows a diseased grape leaf (Black  Rot), from the same dataset. 
						Nevertheless, we expect that an effective classification model should distinguish them accurately. Table \ref{Type} shows all the plant type results from the experiment. The results contain the accuracy and F-1 scores of multi-model, multi-label (power-set), multi-output, multi-task models, and GSMo-CNNs on the benchmark datasets with different CNN backbones.
						
						\iffalse
						\begin{wrapfigure}{r}{.15\textwidth}
							\vskip -0.5cm
							\centering
							\begin{minipage}{0.8\linewidth}
								\centering\captionsetup[subfigure]{justification=centering}
								\includegraphics[width=\linewidth]{plantvillage_apple_health_image__12_.JPG}
								\subcaption{Apple Leaf (Healthy)}
								\label{fig:plantvillage_apple_health}\par\vfill
								\includegraphics[width=\linewidth]{plantvillage_grape_Black_rot_image__8_.JPG}
								\subcaption{Grape Leaf (Black Rot)}
								\label{fig:plantvillage_grape_Black_rot}
								% \includegraphics[width=\linewidth]{image3}
								% \subcaption{}
								% \label{}
							\end{minipage}
							\caption{Samples of Plant Prediction}\label{fig:Plant_Prediction}
							\vskip -0.5cm
						\end{wrapfigure}
						\fi
						
\begin{figure}[h!]
							\centering
							\begin{subfigure}{0.35\textwidth}
								\centering
								\includegraphics[width=\textwidth]{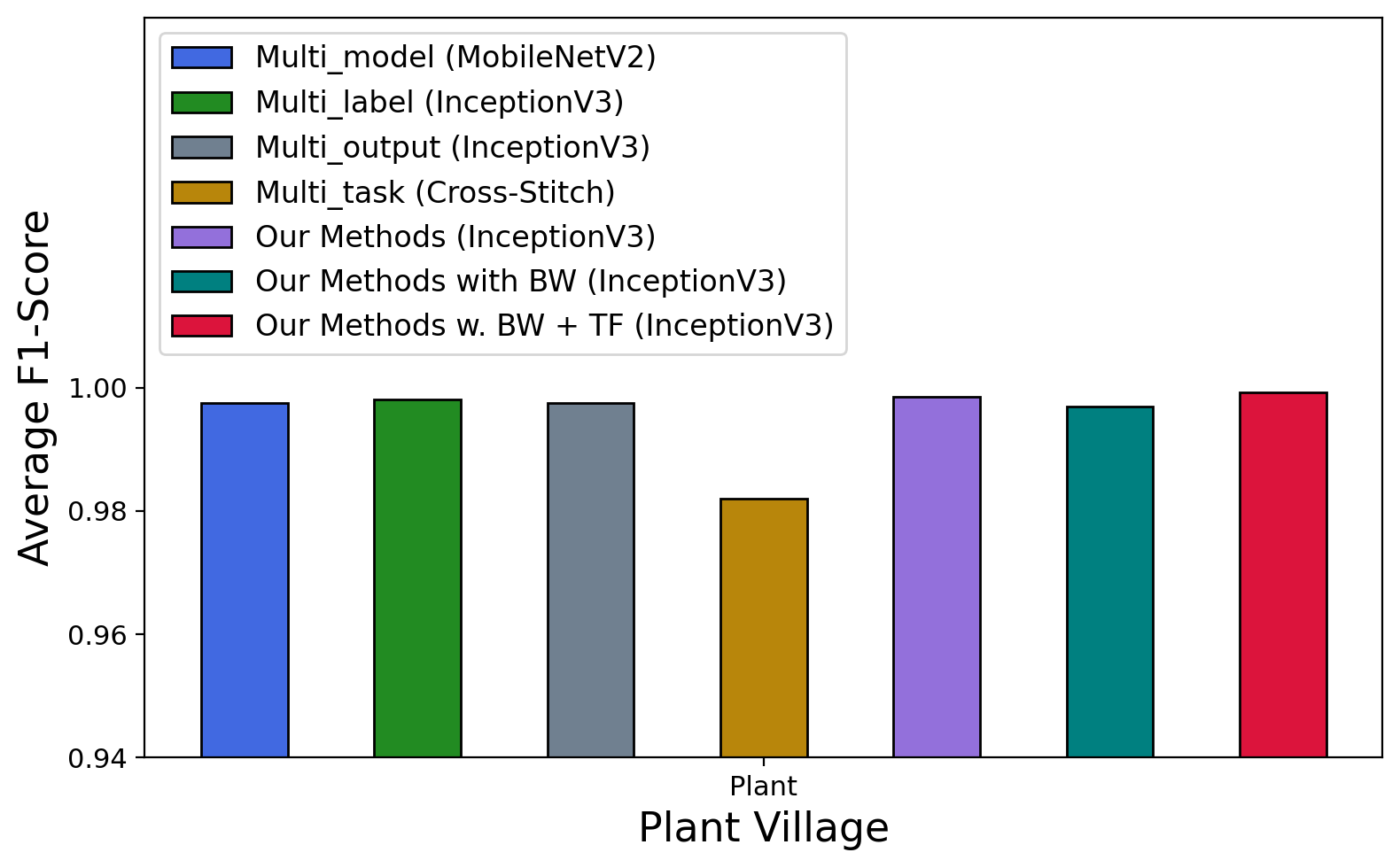}
								\caption{}
								\label{fig:BarChat_pv_plant}
							\end{subfigure}
							\begin{subfigure}{0.35\textwidth}
								\centering
								\includegraphics[width=\textwidth]{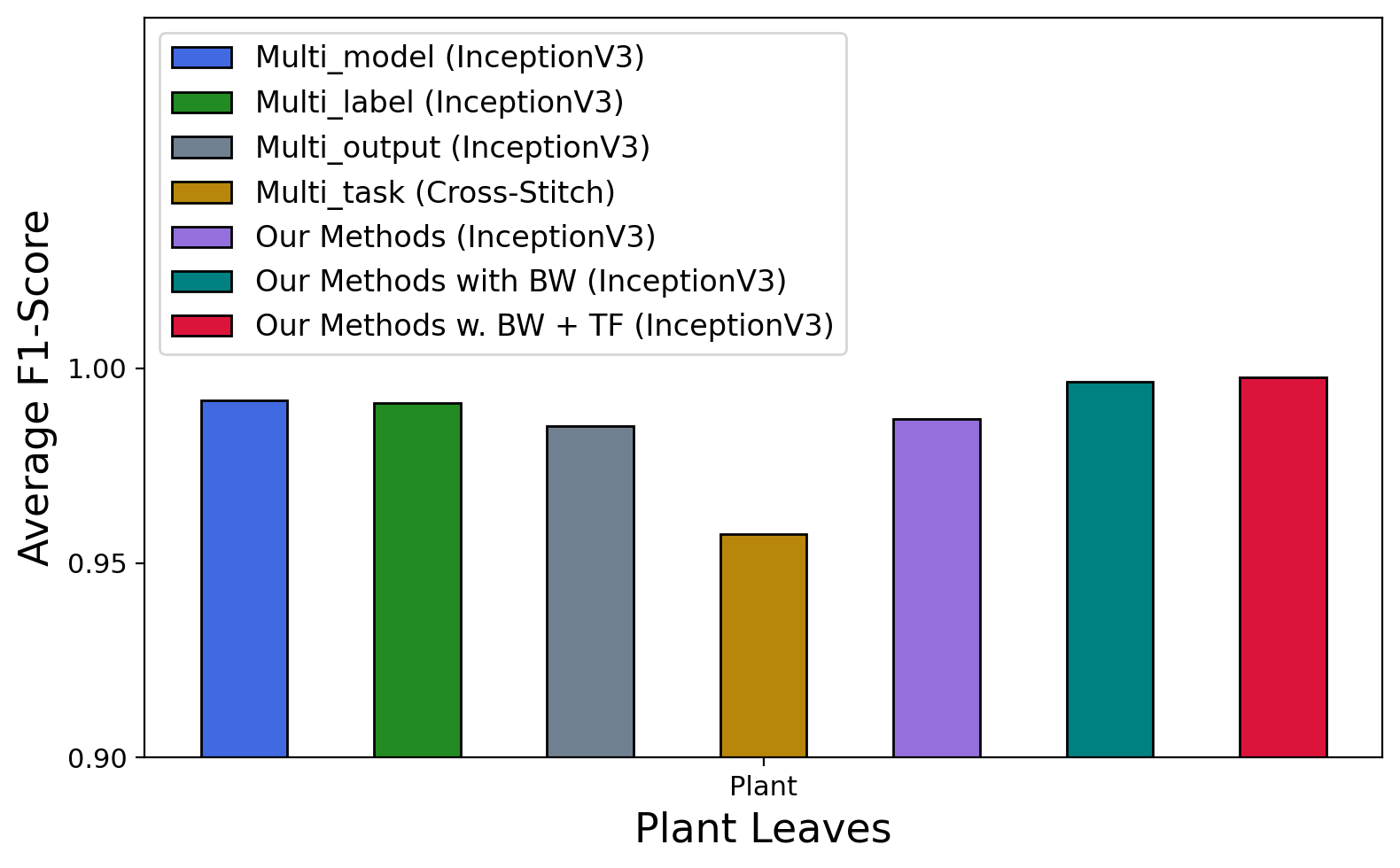}
								\centering
								\caption{}
								\label{fig:BarChat_pl_plant}
							\end{subfigure}\\
							\begin{subfigure}{0.35\textwidth}
								\centering
								\includegraphics[width=\textwidth]{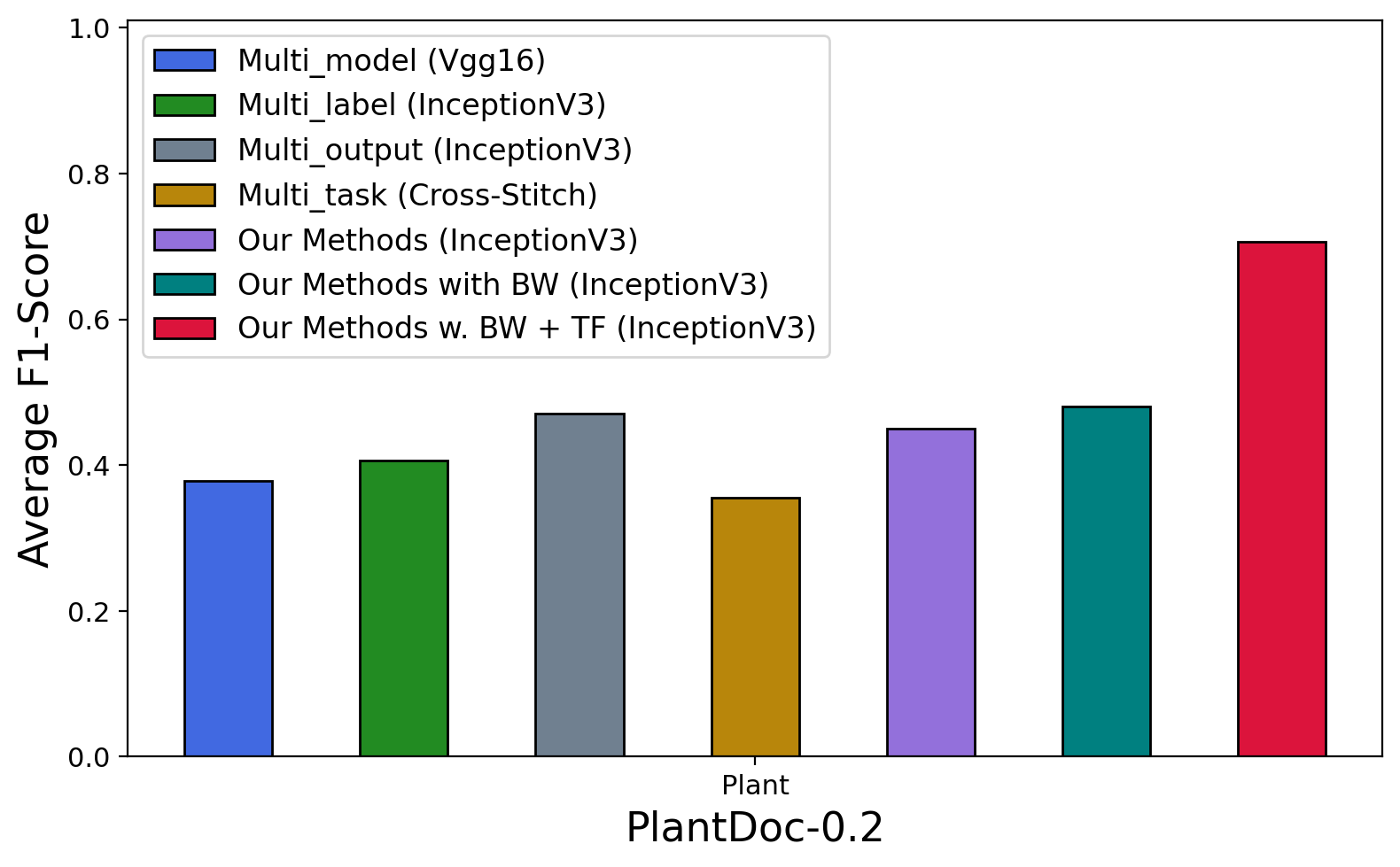}
								\caption{}
								\label{fig:BarChat_pd_plant}
							\end{subfigure}
							\begin{subfigure}{0.35\textwidth}
								\centering
								\includegraphics[width=\textwidth]{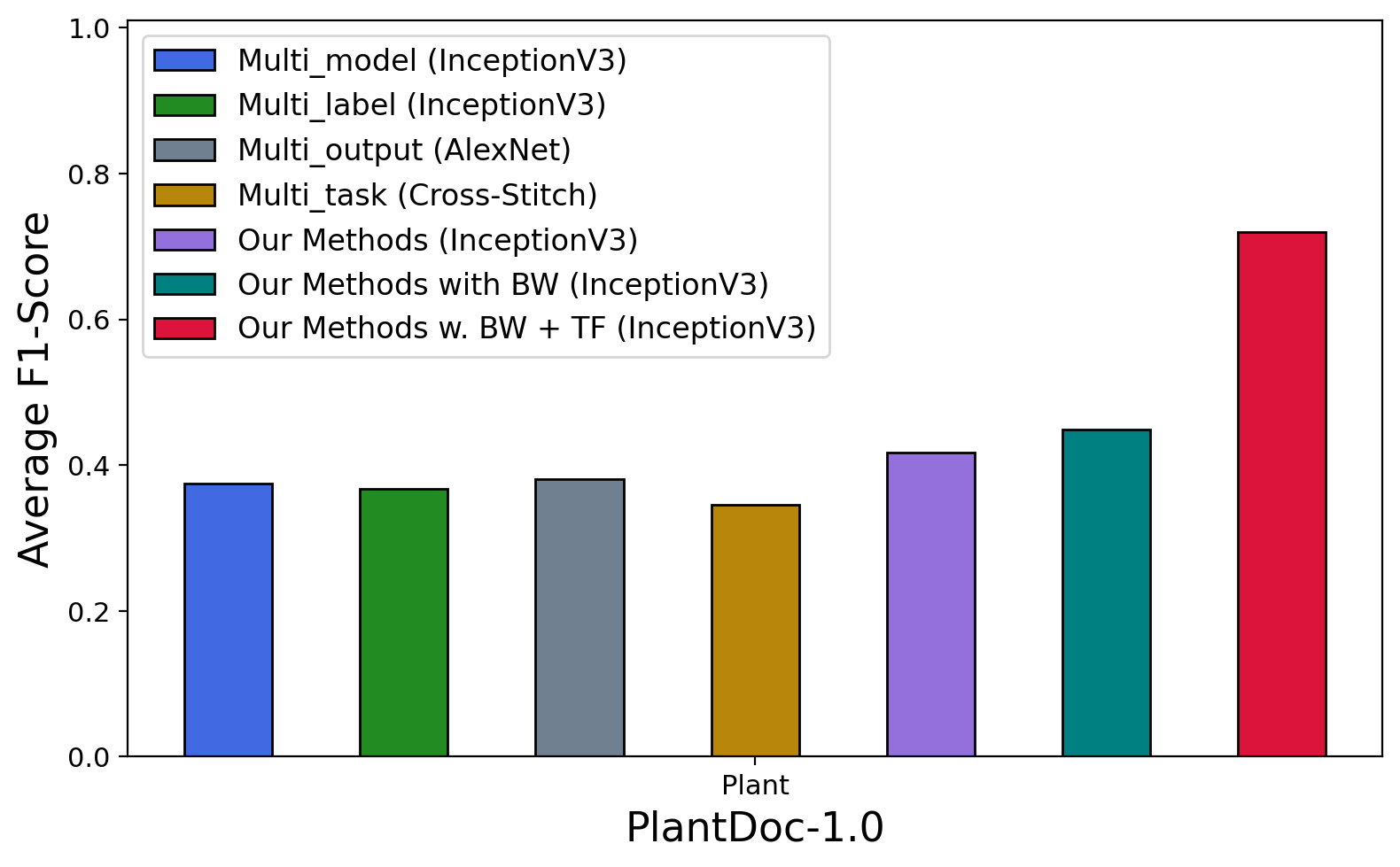}
								\centering
								\caption{}
								\label{fig:BarChat_pdo_plant}
							\end{subfigure}						
							\vskip -0.3cm   
							\caption{\ST{Comparison of The Approaches on Plant Identification.}}
							\label{fig:Comparison_of_BarChat_plant}
							\vskip -0.3cm   
						\end{figure}

						Table \ref{Type}  and Figure \ref{fig:Comparison_of_BarChat_plant} shows that all approaches achieve good performance (more than $90\%$ accuracy \& $90\%$ F1-score) on Plant Village and Plant Leaves. These results confirm the effectiveness of deep learning approaches for classification tasks with image inputs. For PlantDoc (both PlantDoc-0.2 and PlantDoc-1.0), the performance is lower. We anticipate this outcome because the images in Plant Doc are more complex with noisy backgrounds and there can be multiple leaves in one image. Note that, for the sake of the fair evaluation, we don't apply any data processing and augmentation techniques, even though previous research shows that they can improve the performance greatly \cite{9418013,S004579061930002320190601,8974752}.

						In terms of backbone CNNs, as we can see, in multi-model approach, there is an inconsistency of which backbone CNNs has the best performance in all datasets. In particular, MobileNetV2 has the highest accuracy and F1-score in Plant Village, VGG16 is the best in PlantDoc-0.2, while InceptionV3 has the best results in both Plant leaves and PlantDoc-1.0. Different from the multi-model approach, in multi-label and multi-output approaches InceptionV3 clearly shows its advantages with higher performance than other backbones. In PlantDoc-0.2 and PlantDoc-1.0,  InceptionV3 achieves the best performance and its advantages are overwhelming. In particular, its accuracy/F1-score on PlantDoc-0.2 are $51.437\%$/$0.47071$  and on  PlantDoc-1.0 are $43.093\%$/$0.338037$, respectively. In the case of GSMo-CNN, InceptionV3 is also better than other backbone CNNs in Plant Village, Plant Leaves, PlantDoc-2.0, and PlantDoc-1.0.
						
						In terms of common approaches (multi-model, multi-label, multi-output, multi-task), multi-label (power-set) achieves the best results in Plant Village (Accuracy: 99.469\% \& F1-score: 0.99469) and multi-model has the highest average accuracy and F1-score (99.175\% \& 0.99175) in Plant Leaves dataset. Multi-output outperforms multi-model, multi-label, and multi-task in PlantDoc-0.2 (Accuracy: $51.437\%$ , F1-score: $0.4707$) and PlantDoc-1.0 (Accuracy: $43.093\%$, F1-score: $0.38037$ ). Multi-task is inferior to other approaches. This is not surprised as both tasks (plant identification and disease classification) shared the same input data, which is different from original multi-task learning paradigm where different tasks have different sets of data.
						
						For our proposed model GSMo-CNN (denoted as "Our methods (Plant)"  in Table \ref{Type}, we applied 7 backbone CNNs for our new model structure without the balance weight (BW) of each loss (i.e. set $\beta_1$:$\beta_2$:$\delta_1$:$\delta_2$ as 1:1:1:1). InceptionV3 is also the best backbone in this case. This is the backbone we use for GSMo-CNN with balance weights. Without the balance weights, the performance of the GSMo-CNNs is already promising as it achieves better results than multi-model, multi-label, multi-output, and multi-task on Plant Village (Accuracy:$99.850\%$, F1-score: $0.99850$) and on PlantDoc-1.0 (Accuracy: $44.492\%$, F1-score: $0.41757$). With the balance weights, GSMo-CNNs achieves the best performance in three cases: Plant Leaves, PlantDoc-0.2, and PlantDoc-1.0. The accuracy and F1-score are as follows, Plant Village: $99.687\%$ \& $0.99688$); Plant Leaves: $99.646\%$ \& $0.99647$;  PlantDoc-0.2: $49.068\%$, $0.47960$; and PlantDoc-1.0: $46.864\%$ \& $0.44804$.
						
						Finally, when transfer learning is applied we can see improvement in all cases, especially for PlantDoc-0.2 and PlantDoct-1.0. 
						%In particular, plant identification accuracy and F1-score increase from the $99.850\%$ \& $0.99850$ (best result without transfer learning) to $99.928\%$ \& $0.99927$. 
						In Plant Village, the best results without transfer learning are $99.850\%$ accuracy and $0.99850$ F1-score while transfer learning achieves $99.928\%$ accuracy and $0.99927$ F1-score. Huge improvement can be seen in PlantDoc sets. In particular,  our model with transfer learning increase the performance from $51.437\%$ accuracy \& $0.48051$ F1-score to $71.262\%$ \& $0.70602$ on PlantDoc-0.2, and  from $46.992\%$ accuracy \& $0.44934$ F1-score to $73.644\%$ \& $0.71980$ on PlantDoc-1.0.

						\subsection{Disease Classification}
						\label{disease_classification}

						The disease classification task aims to predict the diseases of plants from leaf images. Note that, "healthy" is also treated as a category of plant diseases. The task will not only identify whether a plant has a disease or not but also need to accurately categorise the disease types from different plants. %For example, the leaf in Figure \ref{fig:Corn__Common_rust} is infected with "Common rust" and the leaf in Figure \ref{fig:Corn_Northern_Leaf_Blight} is infected with "Northern Leaf Blight", both in Corn leaf images, but Figure \ref{fig:plantvillage_Apple_scab} shows another disease (Scab) from another type of plant (Apple). 
						A model should be able to focus on the diseases and not be confused by the common patterns from leaves of the same type. Table \ref{Disease} shows the results of different models and approaches, including all the disease classification results from multi-model CNNs, multi-label (power-set) CNNs, multi-output CNNs, multi-task learning approaches and our new model.
						
\begin{figure}[h!]
							\centering
							\begin{subfigure}{0.35\textwidth}
								\centering
								\includegraphics[width=\textwidth]{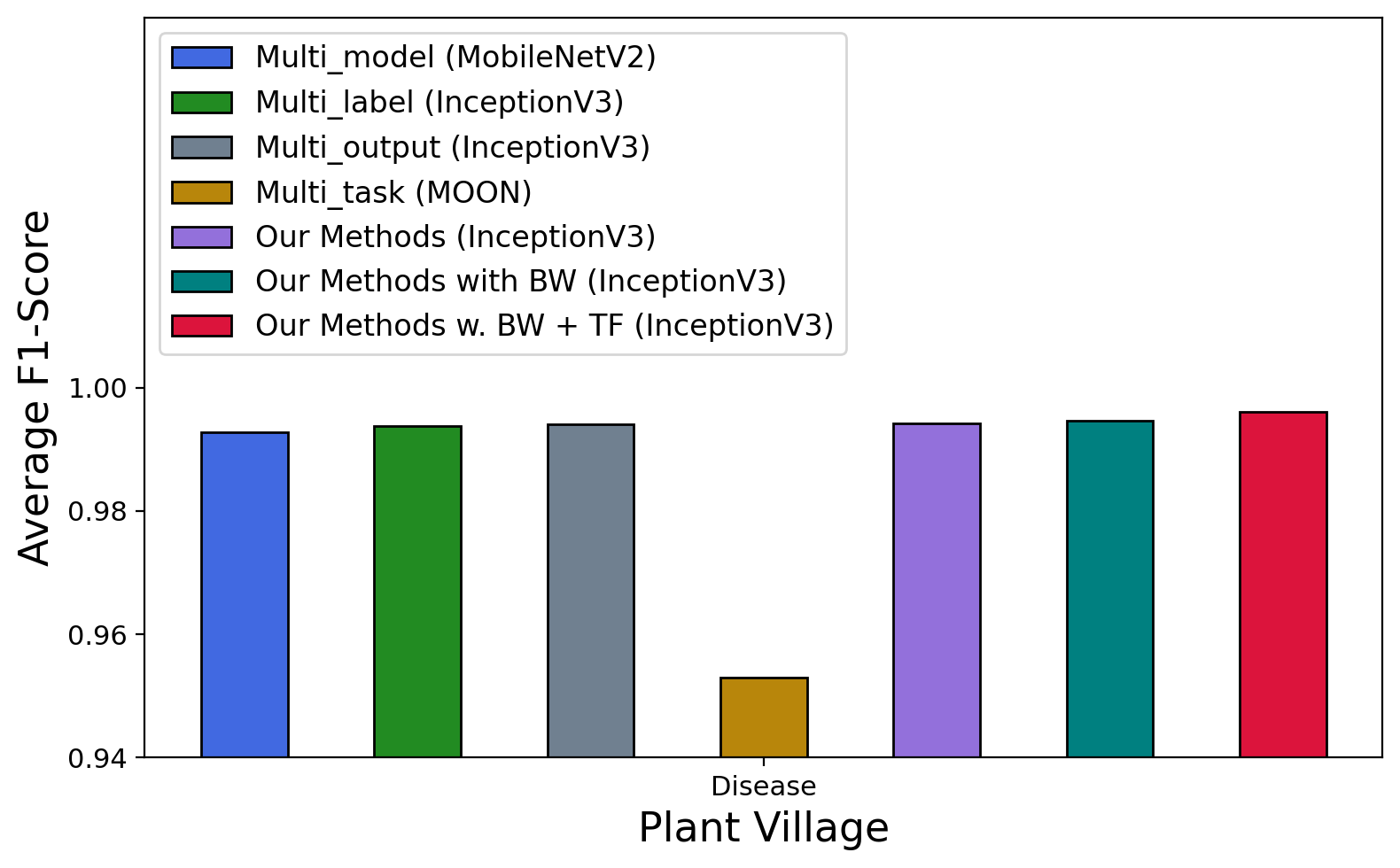}
								\caption{}
								\label{fig:BarChat_pv_disease}
							\end{subfigure}
							\begin{subfigure}{0.35\textwidth}
								\centering
								\includegraphics[width=\textwidth]{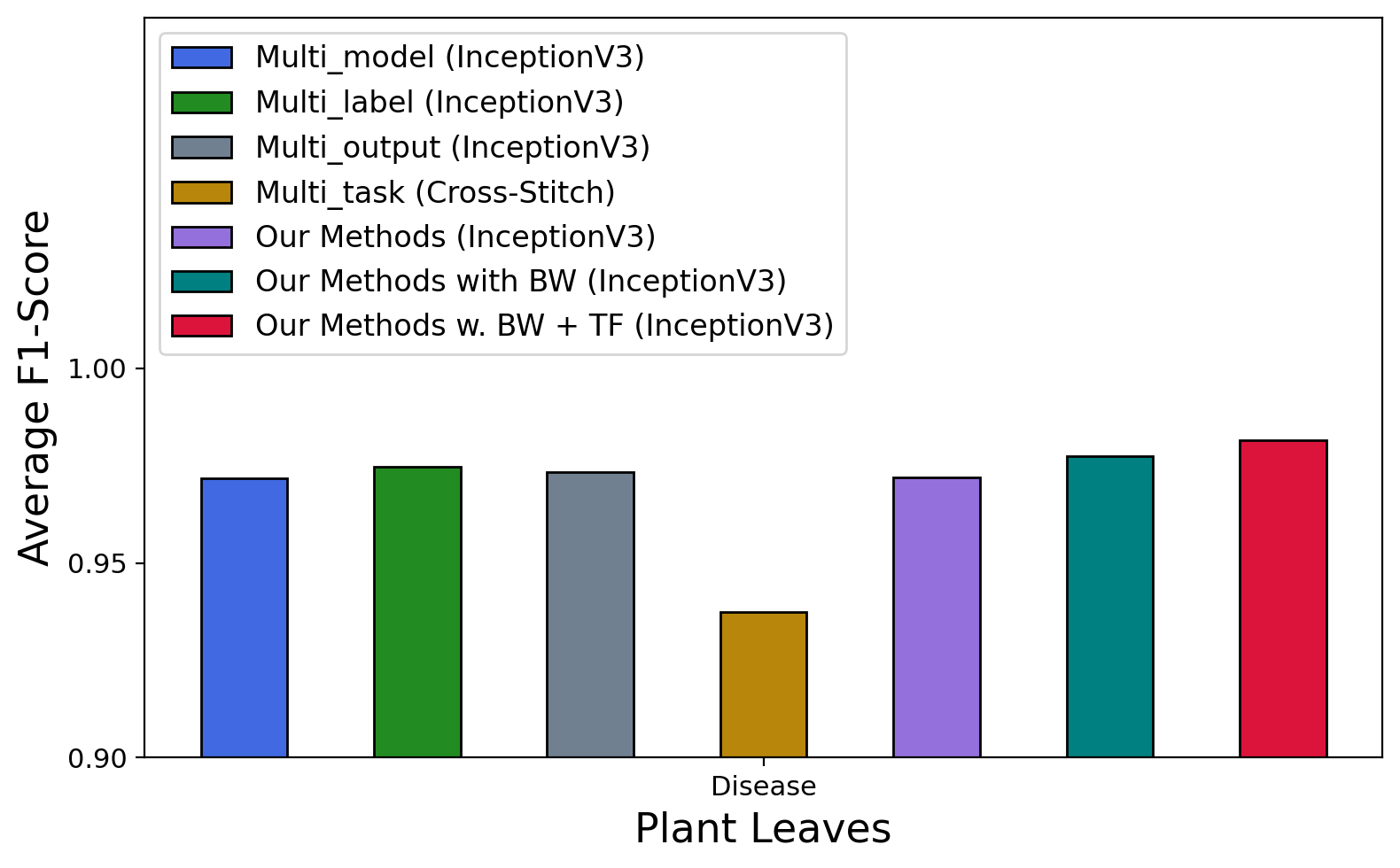}
								\centering
								\caption{}
								\label{fig:BarChat_pl_disease}
							\end{subfigure}\\
							\begin{subfigure}{0.35\textwidth}
								\centering
								\includegraphics[width=\textwidth]{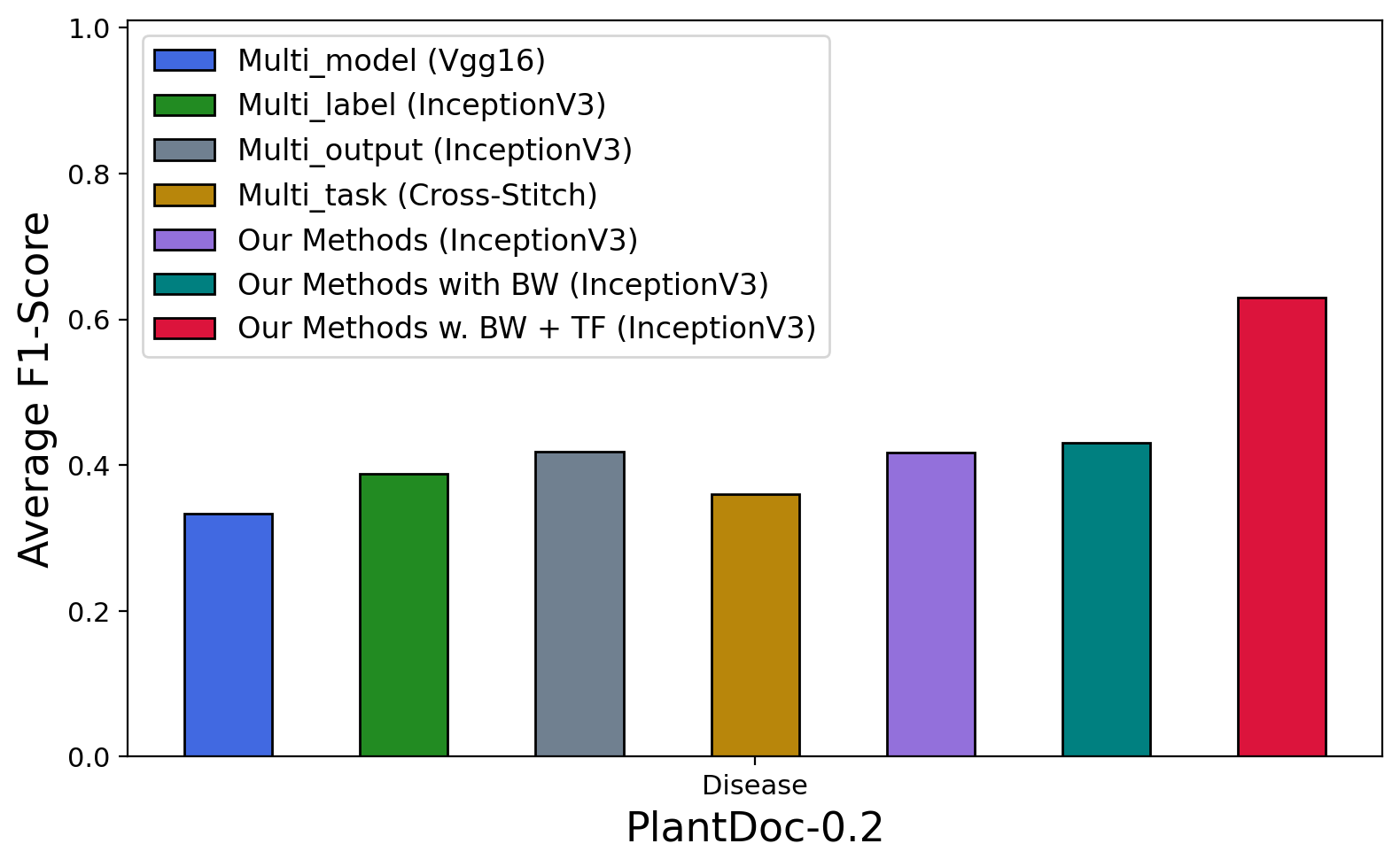}
								\caption{}
								\label{fig:BarChat_pd_disease}
							\end{subfigure}
							\begin{subfigure}{0.35\textwidth}
								\centering
								\includegraphics[width=\textwidth]{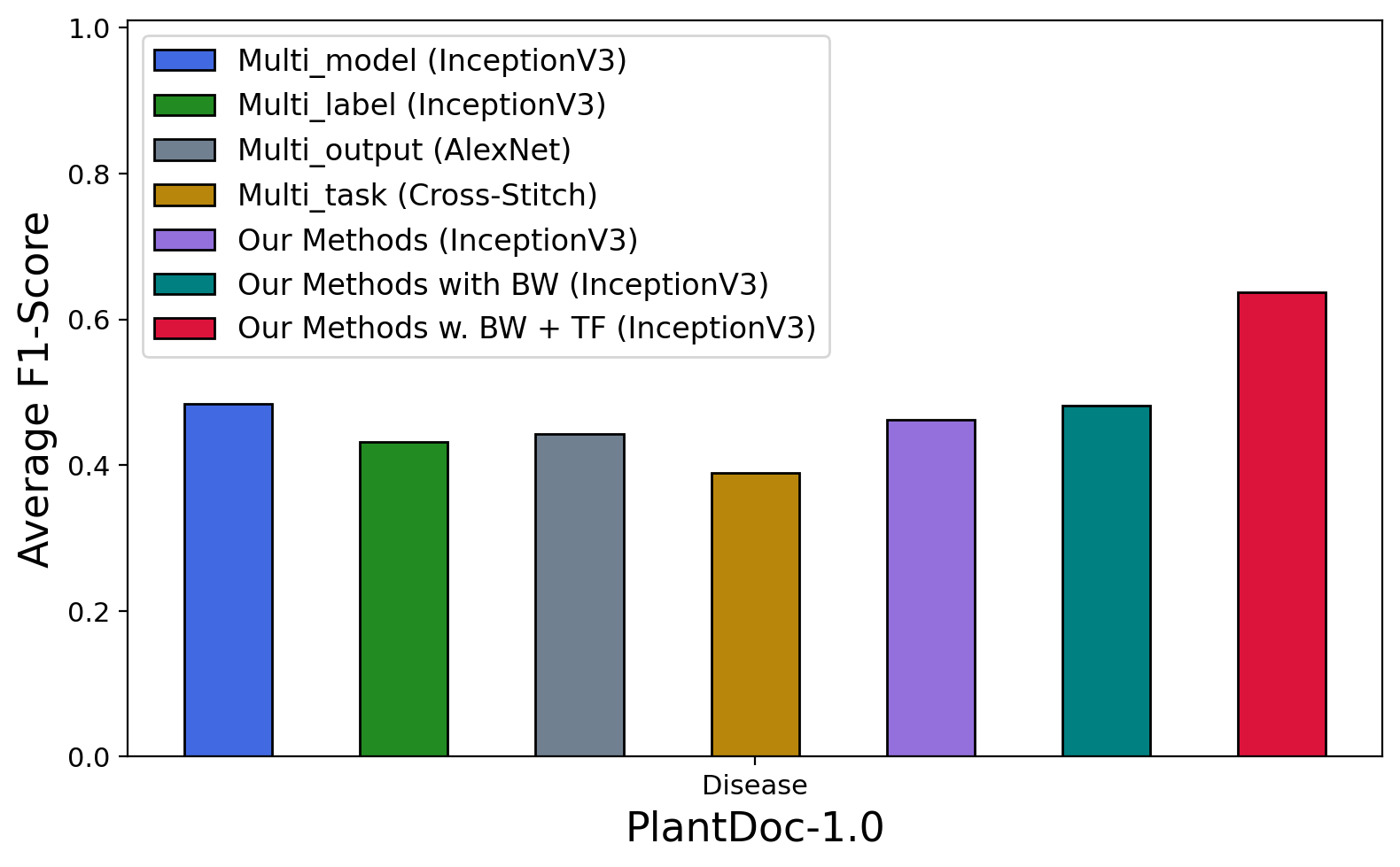}
								\centering
								\caption{}
								\label{fig:BarChat_pdo_disease}
							\end{subfigure}		
							\vskip -0.3cm   
							\caption{\ST{Comparison of The Approaches on Disease Classification.}}
							\label{fig:Comparison_of_BarChat_disease}
							\vskip -0.3cm   
						\end{figure}

						As we can see in Table \ref{Disease} and Figure \ref{fig:Comparison_of_BarChat_disease},  all approaches achieve good performance on Plant Village and Plant Leaves datasets (most accuracy \& F1-score over 80\% \& 0.80). In terms of backbone CNNs, we observe a similar trend as in Plant identification, where InceptionV3 performs better than other CNNs in most of the approaches and datasets. The exception can only be seen in the multi-model approach where MobileNet got the best results in Plant Village ($99.275\%$ accuracy \& $0.99274$ F1-score) and VGG16 achieved $40.291\%$ accuracy and $0.33305$ F1-score in PlantDoc-0.2. In terms of common approaches (multi-model, multi-label, multi-output, multi-task), multi-model (with InceptionV3) achieves the best accuracy ($51.992\%$) and F1-score ($0.48344$) on PlantDoc-1.0;  multi-output (with InceptionV3)  is better than the others on Plant Village ($99.414\%$ accuracy and $0.99413$ F1-score) and on PlantDoc-0.2 ($43.165\%$ accuracy and $0.41892$ F1-score); multi-label (with InceptionV3) achieves the highest accuracy ($97.547\%$) and F1-score ($0.97468$) on Plant Leaves.
						Our proposed model also performs well in the case of disease classification. Without the balance weights, GSMo-CNN achieves better results than multi-model, multi-label, multi-output, and multi-task approaches in PlantVillage and PlantDoc-0.2. The accuracy and F1-score of our new model without balance weights and with InceptionV3 as the backbone are as follows: Plant Village (99.418\% \& 0.99417), Plant Leaves (97.292\% \& 0.97201), PlantDoc-0.2 (45.029\%, 0.41716) and PlantDoc-1.0 (47.542\% \&  0.46156). When balance weights are used, with InceptionV3 as the best backbone GSMo-CNNs achieves the best results in 3 out of 4 cases. In particular, it has $99.466\%$ accuracy and $0.994466$ F1-score on Plant Village dataset;  $97.759\%$ accuracy and $0.97738$ F1-score on Plant Leaves; and $45.301\%$ accuracy and $0.43095$ F1-score  on PlantDoc-0.2. Only in PlantDoc-1.0 where GSMo-CNN ranks second, here, multi-model wins the best accuracy ($51.992\%$) and F1-score ($0.48344$). If we use InceptionV3 pre-trained weights from ImageNet as the backbone for GSMo-CNN, we can even achieve much higher performance. As we can see, the accuracy and F1-score of our model in all datasets are improved significantly as follows, Plant Village (99.615\% accuracy \& 0.99615 F1-score), Plant Leaves (98.149\% accuracy \& 0.98146 F1-score), PlantDoc-0.2 (64.544\% accuracy \& 0.62968 F1-score) and PlantDoc-1.0 (63.305\% accuracy \& 0.63757 F1-score).
						
						%\JY{Figure \ref{fig:Comparison_of_BarChat_disease} visualise the best average F1-Scores results of each approach, all approaches archived good performance in Plant Village (See Subfigure \ref{fig:BarChat_pv_disease}) and Plant Leaves (See Subfigure \ref{fig:BarChat_pl_disease}), however, our methods (red) has a slight margin of victory over the other approaches. Its performance advantages are more evident in the two datasets of PlantDoc (See Subfigures \ref{fig:BarChat_pd_disease} \& \ref{fig:BarChat_pdo_disease}). Both their F1-Scores are more than 0.6. The performance of our three hybrid methods increases from left to right (purple, bluish-green \& red), which shows the effectiveness of our stacked methods (i.e., classifier chain, balance weights \& TF) is verified in the Figure.}

						%Table \ref{Disease} shows the results of this study’s Power Set tasks. InceptionV3 had the best results in Plant Village, Plant leaves and PlantDoc-1.0. And it is one of the top three in PlantDoc-0.2. Other models have their strengths in different datasets, InceptionV3 showed its robustness in all four datasets. All tests would run 10 times and calculate the mean and standard deviation of them.

						\subsection{Plant Identification \& Disease Classification}
						% \begin{figure*}[h!]
							% 	\centering
							% 	\begin{subfigure}{0.3\textwidth}
								% 		\centering
								% 		\includegraphics[width=0.9\textwidth]{plantvillage_Potato_Early_blight_image__16_.JPG}
								% 		\caption{Potato Leaf with Early Blight}
								% 		\label{fig:Potato_Early_blight}
								% 	\end{subfigure}
							% 	\begin{subfigure}{0.3\textwidth}
								% 		\centering
								% 		\includegraphics[width=0.9\textwidth]{plantvillage_Tomato_Septoria_leaf_spot_image__12_.JPG}
								% 		\centering
								% 		\caption{Tomato Leaf with Septoria Leaf Spot}
								% 		\label{fig:Tomato_Septoria}
								% 	\end{subfigure}
							% 	\caption{Samples of Plant Identification \& Disease Prediction}
							% 	\label{fig:Plant_Disease Prediction}
							% \end{figure*}
						
						\iffalse
						\begin{wrapfigure}{r}{.2\textwidth}
							\vskip -0.5cm
							\centering
							\begin{minipage}{0.8\linewidth}
								\centering\captionsetup[subfigure]{justification=centering}
								\includegraphics[width=\linewidth]{plantvillage_Potato_Early_blight_image__16_.JPG}
								\subcaption{Potato Leaf with Early Blight}
								\label{fig:Potato_Early_blight}\par\vfill
								\includegraphics[width=\linewidth]{plantvillage_Tomato_Septoria_leaf_spot_image__12_.JPG}
								\subcaption{Tomato Leaf with Septoria Leaf Spot}
								\label{fig:Tomato_Septoria}
								% \includegraphics[width=\linewidth]{image3}
								% \subcaption{}
								% \label{}
							\end{minipage}
							\caption{Samples of Plant Identification \& Disease Prediction}\label{fig:Plant_Disease Prediction}
							\vskip -0.3cm
						\end{wrapfigure}
						\fi
						For completeness, we will evaluate the models and approaches in predicting plant species and disease types altogether. This is the combination of the plant identification task and the disease classification task, as discussed earlier. However, instead of evaluating each task separately, we are interested in studying how deep learning models can perform accurate predictions for both tasks. A prediction is accurate if and only if both plant species and disease type are inferred correctly. This would be the key criteria for users to select a model for their applications, as usually we want to have information about both plants and diseases to find a suitable treatment. %For example, Figure \ref{fig:Plant_Disease Prediction} demonstrates two different diseases of two different plants. Figure  \ref{fig:Potato_Early_blight} shows a potato leaf with "Early Blight" disease and Figure \ref{fig:Tomato_Septoria} shows a tomato leaf with "Septoria Leaf Spot".
						For evaluation, plant species and disease type will be combined to the joint species-disease labels for the calculation of average accuracy and F1-score.
						
\begin{figure*}[h!]
							\centering
							\begin{subfigure}{0.35\textwidth}
								\centering
								\includegraphics[width=\textwidth]{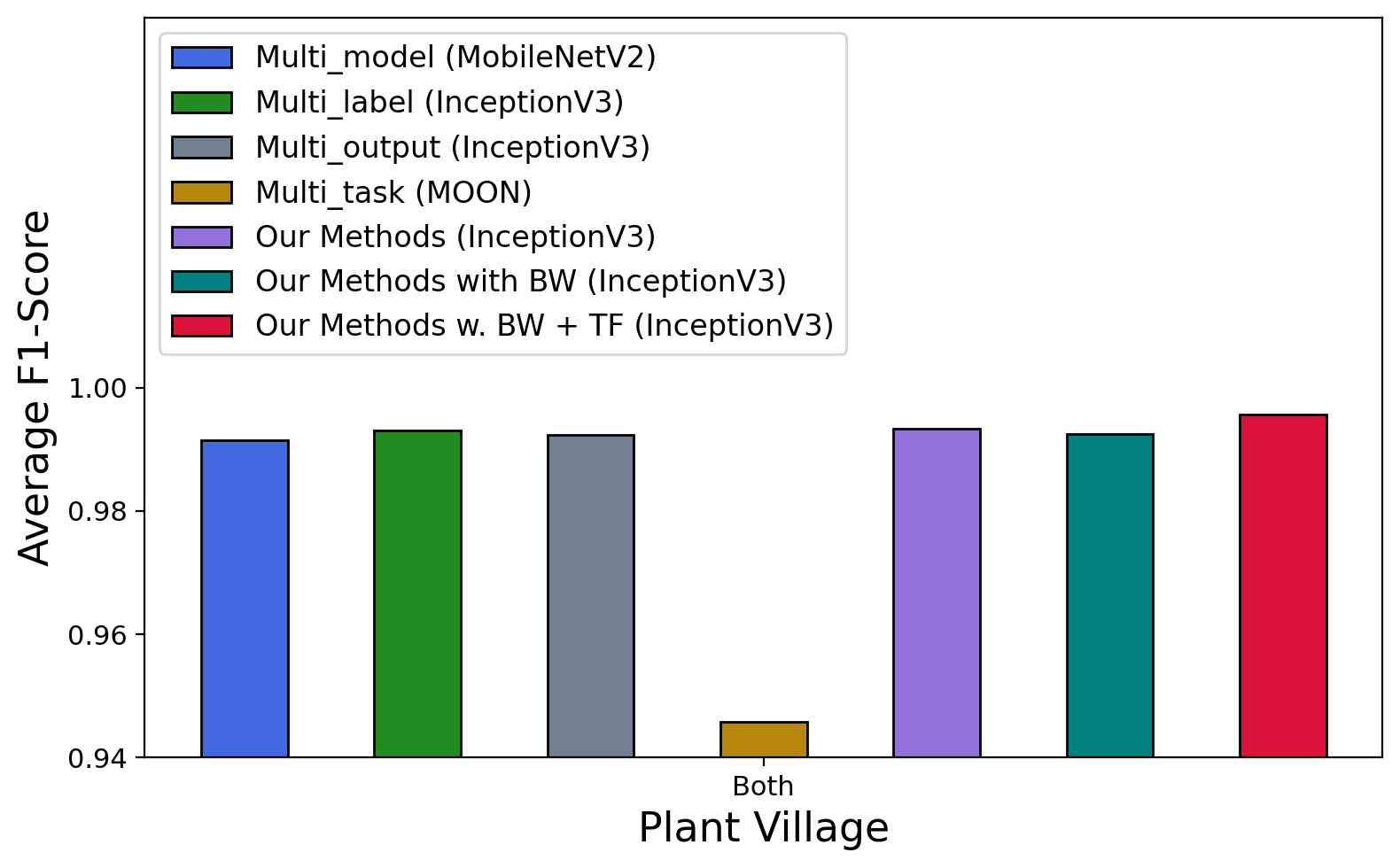}
								\caption{}
								\label{fig:BarChat_pv_both}
							\end{subfigure}
							\begin{subfigure}{0.35\textwidth}
								\centering
								\includegraphics[width=\textwidth]{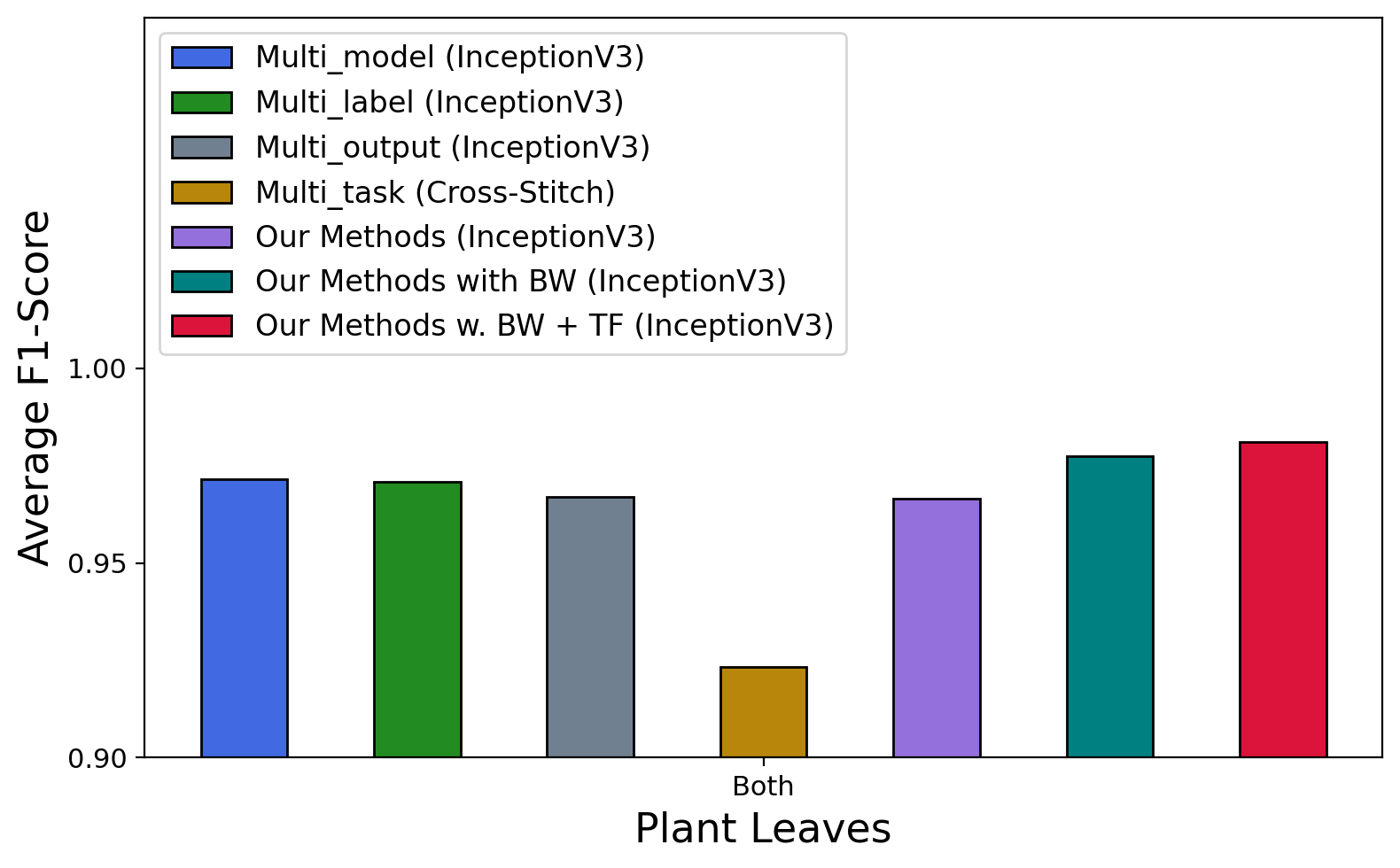}
								\centering
								\caption{}
								\label{fig:BarChat_pl_both}
							\end{subfigure}\\
							\begin{subfigure}{0.35\textwidth}
								\centering
								\includegraphics[width=\textwidth]{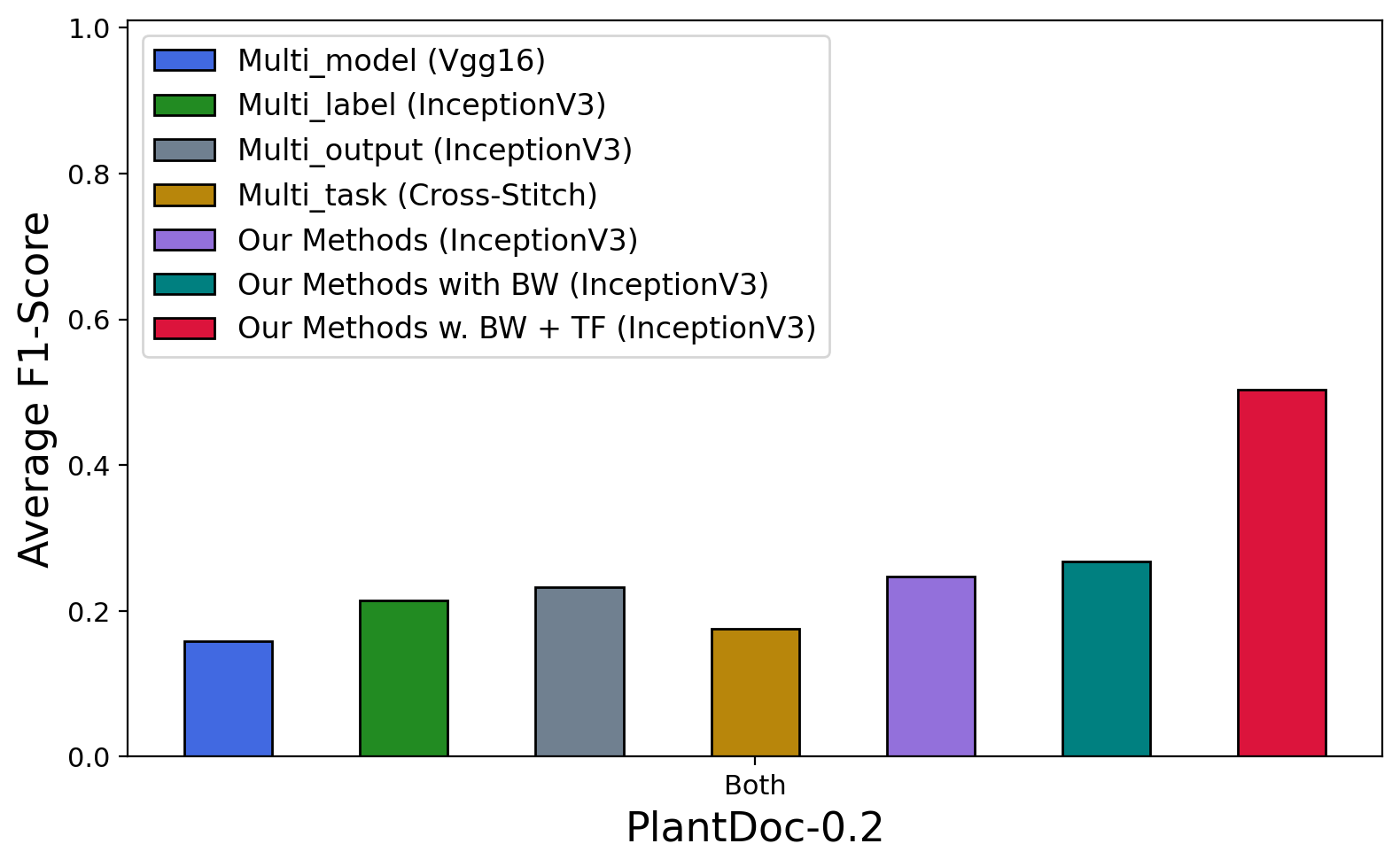}
								\caption{}
								\label{fig:BarChat_pd_both}
							\end{subfigure}
							\begin{subfigure}{0.35\textwidth}
								\centering
								\includegraphics[width=\textwidth]{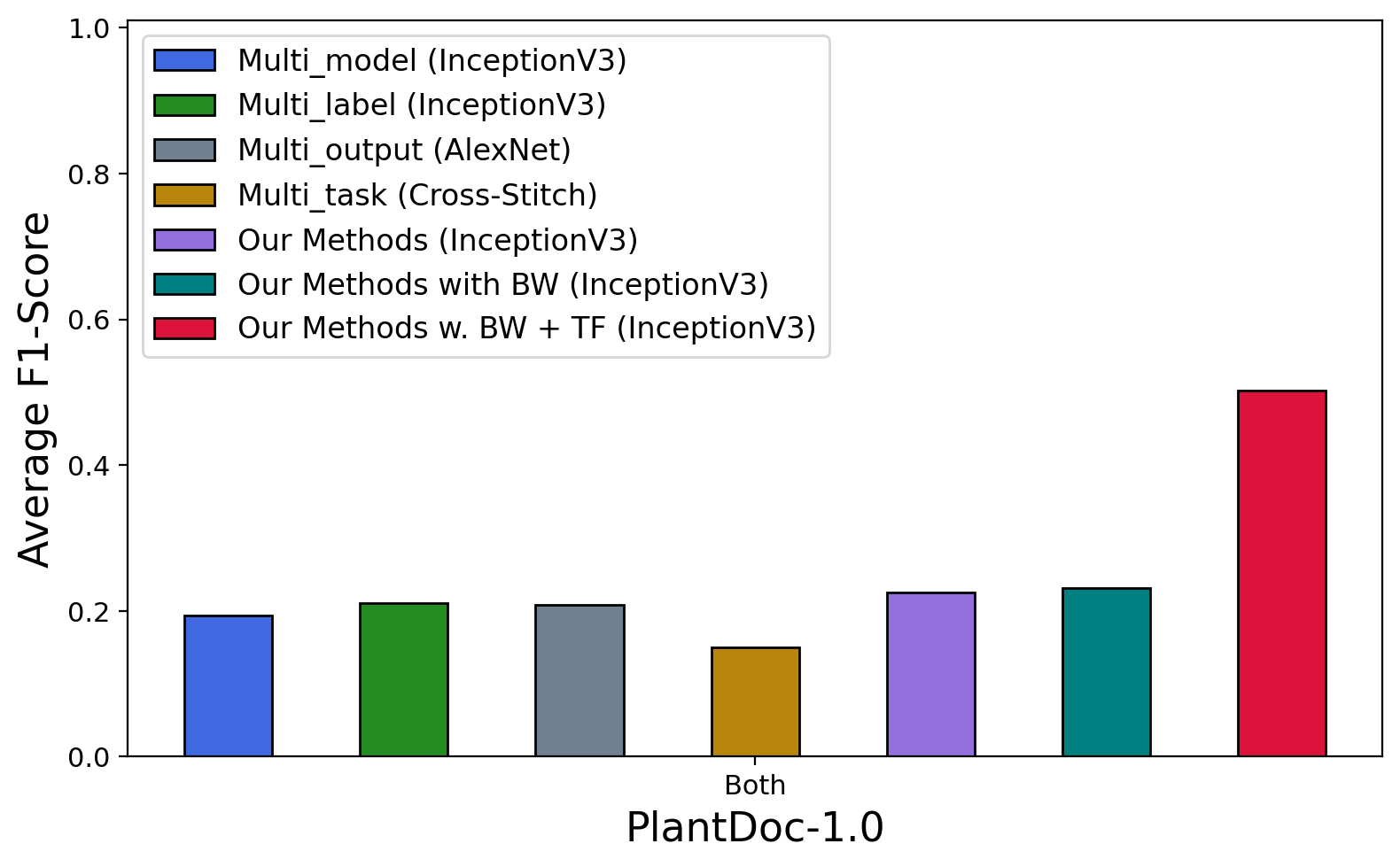}
								\centering
								\caption{}
								\label{fig:BarChat_pdo_both}
							\end{subfigure}		
							\vskip -0.3cm   
							\caption{\ST{Comparison of The Approaches on Plant Identification and Disease Classification.}}
							\label{fig:Comparison_of_BarChat_both}
							\vskip -0.3cm   
						\end{figure*}

						Table \ref{Total} and  Figure \ref{fig:Comparison_of_BarChat_both} shows the results of all models and approaches. Although the combined prediction of plant species and disease types is more complex than the previous two tasks, all approaches achieved promising results with more than $85\%$ average accuracy and $0.85$ F1-score on Plant Village and Plant Leaves. In terms of backbone CNNs, the multi-model approach demonstrates that MobileNet achieves the best performance ($99.05\%$ accuracy and $0.99148$ F1-score) on Plant Village and VGG16 achieves the highest results on PlantDoc-0.2 ($16.252\%$ accuracy \& $0.15821$ F1-score). The multi-output approach shows that AlexNet has the highest accuracy of $22.458\%$ and F1-score of $0.20815$ on PlantDoc-1.0. Except for such three cases, InceptionV3 achieves higher performance than other backbone CNNs in the other cases.
						
						Among the common multi-prediction approaches (multi-model, multi-label, multi-output, and multi-task), multi-label (power-set) performs the best on Plant Village, Plant Leaves, and PlantDoc-1.0 while multi-output has the best results on PlantDoc-0.2. It makes sense as the power-set labelling in the multi-task approach combines the two labels together, therefore, the learning would directly optimise the prediction of the combined label (plant \& disease). Interestingly, multi-output performs well in PlantDoc-0.2 even though there is no communication between the two labels during the learning. Having said that, it can be the back-propagation of gradients from the two branches to the shared CNNs that plays a role in regularisation for the shared features in the learning. Such regularisation would help the model to learn more generalised features rather than task-specific features. The multi-task learning models (TSNs, MOON, Cross Stitch \& MTAN) show good results in Plant Village and Plant Leaves, however, they are not comparable to other approaches.
						
						Our proposed model GSMo-CNN achieves the best performance on all datasets, except accuracy on PlantDoc-1.0 where multi-label with InceptionV3 backbone tops the list with $25.466\%$. In the case where balance weights are not applied ($\beta_1:\delta_1:\beta_2:\delta_2=1:1:1:1$), GSMo-CNN is better than multi-model, multi-label, multi-output, and multi-task. Its best performance is: $99.315\%$ accuracy and  $0.99333$ F1-score on Plant Village; $96.593\%$ accuracy and $0.96641$ F1-score on Plant Leaves; $25.359\%$ accuracy and  $0.24692$ F1-score on PlantDoc-0.2; and $23.814\%$ accuracy and $0.22487$ F1-score on PlantDoc-1.0.  Again, InceptionV3 is the best backbone CNN for GSMo-CNN. When balance weights are employed, the improvement can be seen on Plant Leaves, PlantDoc-0.2, and PlantDoc-1.0. As mentioned earlier, we apply a grid search and narrow down the grid to find a combination of the weights that perform well on all datasets. The reported results in Tables \ref{Type}, \ref{Disease}, \ref{Total} are from the balance weights $\beta_1:\delta_1:\beta_2:\delta_2=0.1:0.1:0.4:0.5$. In particular, GSMo-CNN with BW achieves $99.208\%$ accuracy and $0.99245$ F1-score on Plant Village; $97.524\%$ accuracy and $0.97746$ F1-score on Plant Leaves; $26.175\%$ accuracy and  $0.26760$ F1-score on PlantDoc-0.2; and $24.153\%$ accuracy and $0.23191$ F1-score on PlantDoc-1.0.
						
						Similar to the independent prediction of plant and disease as we shown in \ref{plant_identification} and \ref{disease_classification}, when we initialise InceptionV3 from the weights pre-trained on ImageNet, GSMo-CNN can improve the performance in all datasets, producing the highest accuracy and F1-scores in this study. The accuracy and F1-scores are as follows, Plant Village: 99.558\% \& 0.99565; Plant Leaves: 98.042\% \& 0.98116; PlantDoc-0.2: 50.291\% \& 0.50396;  and PlantDoc-1.0: 50.000\% \& 0.50191.
						
						It is worth noting that our proposed method does not have the problem of computational overhead because (1) we do not need to combine the labels during the learning; and (2) stacking the labels is efficient and its structure only needs a single CNN as the backbone. The only complexity users might find from our approach is the search for balance weights. Fortunately, as we showed in our intensive experiment that it can be optional as GSMo-CNN without balance weights can already achieve better performance than other approaches. Together with the transfer learning results the balance weights even bolster the effectiveness of GSMo-CNN. This shows the flexibility of our approach where users are provided with different options, including the choice of backbone CNNs, the choice of the balance weights for the losses, and/or the choice of pre-training weights. All these options have proved to be effective in improving the performance of plant identification and disease classification. We have shown in the experiment that the idea of the proposed deep learning structure, where two labels (Plant \& Disease) can help each other to eliminate some wrong options to improve the prediction accuracy. In the next section, let us summarise the findings and provide further analysis of the results.
						%Table \ref{Multi-Total} shows the results of this study’s multi-output task on Plant Village, InceptionV3 was the best baseline model. And compare to Table \ref{Power_Set}, most baselines’ results of multi-output are all better than those in Power Set (i.e. VGG16, ResNet101, EfficientNet, InceptionV3 and MobileNetV2). 
						%Table \ref{Multi-Output_PlantLeaves} shows the results of this study’s multi-output task on Plant leaves. In this dataset, baselines in Power Set are better than in multi-output. InceptionV3 had the best results. In the two tasks of PlantDoc, there is not much difference between the two methods, but overall Power Set is better. InceptionV3 was the best in PlantDoc-0.2 test set) and CNN was the best in PlantDoc (original test set).
						
						%In summary, InceptionV3 still maintains good performance and robustness against different data sets. In future studies, it should be paid more attention to conduct further studies to extract key advantages that may benefit the proposed model in leaf disease classification.

						\clearpage
						\newgeometry{margin=1cm}
						\thispagestyle{empty}
						
						\begin{table}[h!]
							%\sidewaystablefn%
							\centering
							\begin{minipage}{1.3\textwidth}
								%	\begin{center}
									
									%	\resizebox{\textwidth}{!}{%
										\caption{Plant Type Results.}\label{Type}
										%			\begin{minipage}{1.2\textwidth}
											%				\resizebox{\textwidth}{!}{%
												\begin{sideways}
													\begin{tabular*}{\textwidth}{@{\extracolsep{\fill}}ll|cc|cc|cc|cc@{\extracolsep{\fill}}}
														\toprule
														&& \multicolumn{2}{@{}c@{}}{Plant Village} & \multicolumn{2}{@{}c@{}}{Plant leaves}& \multicolumn{2}{@{}c@{}}{PlantDoc-0.2} & \multicolumn{2}{@{}c@{}}{PlantDoc-1.0}\\%
														\cmidrule{3-4}\cmidrule{5-6}\cmidrule{7-8}\cmidrule{9-10}%
														& & Acc & F1 & Acc & F1 & Acc & F1& Acc & F1 \\
														\midrule
														%%%%%%%%%%%%%%%%%%%%%%%%%%%%%%%%%%%%%%%%%%%%%%%%%%%%
														\multirow{8}{*}{\begin{tabular}{@{}l@{}} Multi-model \\ (Plant) \end{tabular}}
														
														& CNN & $97.448\% \pm 00.148\%$	&$0.97441 \pm 0.00152$	&$94.451\% \pm00.888\%$	&$0.94416 \pm 0.00872$	&$35.515\% \pm 00.984\%$ & $ 0.31791 \pm 0.00757$	& $37.712\% \pm 00.000\%$	&$0.33216 \pm 0.00000$ \\
														& AlexNet & $98.655\% \pm 00.321\%$	& $0.98654 \pm 0.00322$ & $96.085\% \pm 00.840\%$	& $0.96089 \pm 0.00852$	& $40.019\% \pm 02.968\%$ &	$0.32573 \pm 0.03386 $	& $36.398\% \pm 01.449\%$	& $ 0.34042 \pm 0.03723 $ \\
														&VGG16	& $99.465\% \pm 00.319\%$	& $0.99465 \pm 0.00320 $	& $96.792\% \pm 00.755\%$	& $0.96789 \pm 0.00758 $	& $45.553\% \pm 02.971\%$	& $ 0.37827 \pm 0.03285 $	& $42.119\% \pm 02.919\%$	& $ 0.36066 \pm 0.04843 $\\
														&ResNet101	& $99.427\% \pm 00.095\%$	& $0.99428 \pm 0.00095$	& $98.467\% \pm 00.000\%$	& $0.98465 \pm 0.00000$	& $26.369\% \pm 03.394\%$	& $0.18736 \pm 0.01701$	& $29.958\% \pm01.444\%$	& $0.21007 \pm 0.02095$ \\
														&EfficientNet	& $99.389\% \pm 00.074\%$	& $0.99388 \pm 0.00075 $	& $94.717\% \pm 00.689\%$	& $0.94707 \pm 0.00674 $ &	$34.369\% \pm 00.000\%$	&$ 0.28476 \pm 0.00829$	& $31.780\% \pm 01.705\%$	& $0.23555 \pm 0.01912$ \\
														&InceptionV3 & $99.389\% \pm 00.074\%$	& $0.99388 \pm 0.00075$	& $99.175\% \pm 00.000\%$	& $0.99175 \pm 0.00000$	&$45.068\% \pm 01.495\%$	& $0.36668 \pm 0.02233$	& $40.297\% \pm 03.301\%$	& $0.37522 \pm 0.03833$ \\
														&MobileNetV2	& $99.750\% \pm 00.000\%$	& $0.99750 \pm 0.00000$	& $96.050\% \pm 00.260\%$	& $0.96065 \pm 0.00253$	& $13.592\% \pm 00.000\%$	& $0.03253 \pm 0.00000$	& $25.593\%
														\pm 07.288\%$	& $0.11020 \pm 0.04417$\\
														& ViT &$ 92.045\% \pm  00.875\%$ &$ 0.91833\pm 0.00980 $ &$94.617\% \pm  00.166\%$	&$  0.94635\pm  0.00153$&$ 30.874 \% \pm 00.793\%$	&$  0.15409 \pm 0.01404$ &$ 29.661\% \pm 00.346\%$	&$  0.14268 \pm  0.00854$ \\	
														\hline
														\hline
														\multirow{8}{*}{\begin{tabular}{@{}l@{}} Multi-label\\ (Plant) \end{tabular}}
														&CNN	& $98.645\% \pm00.160\%$	&$0.98643 \pm0.00159$	&$94.384\% \pm01.338\%$&	$0.94404 \pm0.01318$&	$38.252\% \pm00.000\%$&	$0.36807 \pm0.00239$&	$34.746\% \pm00.000\%$&	$0.32100 \pm 0.00000$\\
														&AlexNet&	$98.578\% \pm00.092\%$&	$0.98581 \pm 0.00089$&	$96.404\% \pm00.489\%$&	$0.96405 \pm 0.00501$&	$33.981\% \pm02.044\%$&	$0.32072 \pm 0.02660$&	$32.500\% \pm06.460\%$&	$0.31231 \pm 0.06649$\\
														&VGG16&	$97.114\% \pm 00.185\%$& $0.97110 \pm0.00187$&	$92.175\% \pm00.055\%$&	$0.92139 \pm0.00031$&	$30.252\% \pm01.738\%$&	$0.28354 \pm 0.01460$&	$23.771\% \pm03.301\%$&	$0.22337 \pm0.02978$\\
														&ResNet101&	$99.704\% \pm00.056\%$&	$0.99704 \pm 0.00056$&	$98.257\% \pm00.051\%$&	$0.98261 \pm0.00048$&	$27.340\% \pm02.315\%$&	$0.24259 \pm 0.02912$&	$23.390\% \pm03.187\%$&	$0.21037 \pm 0.01491$\\
														&EfficientNet	&$ 99.314\% \pm 00.089\%$ &$ 0.99315 \pm 0.00089$ & $95.538\% \pm00.932\%$&	$0.95510 \pm 0.00902$&	$23.903\% \pm01.097\%$&	$0.21702 \pm0.01132$&	$21.102\% \pm03.559\%$&	$0.17609 \pm0.05528$ \\
														&InceptionV3&	$99.810\% \pm00.118\%$&	$0.99810 \pm0.00118$&	$99.112\% \pm00.000\%$&	$0.99111 \pm0.00000$& $41.981\% \pm02.608\%$& $0.40630 \pm 0.01026$&	$38.771\% \pm00.636\%$&	$0.36693 \pm 0.00783$ \\
														&MobileNetV2&	$99.469\% \pm00.019\%$&	$0.99469 \pm 0.00019$&	$95.450\% \pm00.778\%$&	$0.95464 \pm0.00786$&	$13.592\% \pm00.000\%$&	$0.03253 \pm 0.00000$&	$11.017\% \pm00.000\%$&	$ 0.02187 \pm0.00000$  \\
														& ViT &$  92.238\% \pm 00.175\%$ &$ 0.92285\pm 0.00168 $ &$ 95.560\% \pm 00.333\%$	&$  0.95563 \pm 0.00350 $	&$ 22.718 \% \pm 00.583\%$	&$  0.20821\pm 0.00993$	&$ 17.797\% \pm 01.695\%$	&$ 0.16317 \pm 0.02967$ \\	
														%%%%%%%%%%%%%%%%%%%%%%%%%%%%%%%%%%%%%%%%%%%%%%%%%%%%%%%%%%%%%%%%%%%
														\hline
														\hline
														\multirow{8}{*}{\begin{tabular}{@{}l@{}} Multi-output\\ (Plant) \end{tabular}}
														&CNN	&$97.843\% \pm00.154\%$	&$0.97838 \pm0.00150$	&$92.963\% \pm00.997\%$	&$0.92930 \pm0.01038$	&$39.398\% \pm02.940\%$	&$0.37328 \pm 0.02085$	&$35.805\% \pm00.636\%$	&$0.33279 \pm0.00302$ \\
														&AlexNet	&$99.356\% \pm00.056\%$	&$0.99357 \pm 0.00056$	&$95.871\% \pm00.435\%$	&$0.95862 \pm0.00473$	&$39.456\% \pm02.032\%$	&$0.37112 \pm 0.01990$	&$40.254\% \pm00.000\%$	&$0.36676 \pm 0.00000$ \\
														&VGG16	&$99.407\% \pm00.035\%$	&$0.99406 \pm0.00034$	&$96.659\% \pm00.812\%$	&$0.96652 \pm 0.00803$	&$41.592\% \pm00.117\%$	&$0.33072 \pm 0.00733$	&$42.585\% \pm03.178\%$	&$0.35115 \pm0.01855$ \\
														&ResNet101	&$99.681\% \pm00.019\%$	&$0.99681 \pm0.00019$	&$97.658\% \pm00.033\%$	&$0.97651 \pm0.00031$	&$34.951\% \pm00.000\%$	&$0.25710 \pm 0.00000$	&$34.746\% \pm00.000\%$	&$0.27496 \pm0.00000$ \\
														&EfficientNet	&$ 99.506\% \pm 00.106\%$ &$ 0.99506 \pm 0.00106$ &$95.183\% \pm01.100\%$	&$0.95161 \pm0.01112$	&$33.126\% \pm02.228\%$	&$0.25029 \pm 0.01327$	&$31.695\% \pm01.525\%$	&$0.21581 \pm 0.01123$ \\
														&InceptionV3 &$ 99.752\% \pm 00.044\%$ &$ 0.99753 \pm 0.00044$ &$98.524\% \pm00.222\%$	&$0.98521 \pm0.00223$	&$\textbf{51.437\%} \pm01.513\%$	&$0.47071 \pm 0.02533$	&$43.093\% \pm03.902\%$	&$0.38037 \pm0.04974$ \\
														&MobileNetV2 &$ 99.634\% \pm 00.103\%$ &$ 0.99634
														\pm 0.00104$ &$94.750\% \pm01.129\%$	&$0.94731 \pm 0.01103$	&$25.010\% \pm07.474\%$	&$0.10613 \pm0.04818$	&$20.127\% \pm09.110\%$	&$0.07708 \pm 0.05521$ \\
														& ViT &$  93.352\% \pm 00.837\%$ &$ 0.93246 \pm 0.00906 $ &$ 90.035\% \pm 01.002\%$ &$ 0.89969 \pm 0.00947$	&$ 29.900 \% \pm 00.000\%$	&$  0.13770 \pm 0.00000 $	&$ 0.29237\% \pm 00.000\%$	&$ 0.13229\pm 0.00000$ \\							
														% & ViT &$  \% \pm  \%$ &$ \pm  $ &$ \% \pm \%$	&$  \pm  $	&$ \% \pm \%$	&$  \pm $	&$ \% \pm \%$	&$  \pm  $ \\			
														\hline
														\hline
														\multirow{4}{*}{\begin{tabular}{@{}l@{}} Multi-task \\ (Plant) \end{tabular} }
														& Cross-Stitch & $98.197\% \pm 00.155\%$ & $ 0.98196 \pm 0.00153$ &$ 95.743\% \pm 00.318\%$	&$ 0.95737 \pm 0.00324$	&$ 39.845\% \pm 00.720\%$	&$ 0.35536 \pm 0.01282$	&$ 38.814\% \pm 02.734\%$	&$ 0.34577 \pm 0.03165$ \\								
														&MTAN & $95.461\% \pm  00.390\%$ & $ 0.95464 \pm 0.00395$ &$ 90.377\% \pm 02.631\%$	&$ 0.90345 \pm 0.02651$	&$ 27.825\% \pm 02.394\%$	&$ 0.19536 \pm 0.02203$	&$ 30.169\% \pm 01.180\%$	&$ 0.18824 \pm 0.01435$ \\
														&TSNs & $91.704\% \pm  01.452\%$ & $  0.91686 \pm 0.01468$ &$ 85.749\% \pm 03.536\%$	&$ 0.85800 \pm 0.03514$	&$ 27.786\% \pm 02.775\%$	&$ 0.18716 \pm 0.02020$	&$ 32.881\% \pm 01.989\%$	&$ 0.21219 \pm 0.03381$ \\
														
														&MOON & $ 97.589\% \pm   00.410\%$ & $ 0.97586 \pm  0.00411$& $ 95.627\% \pm 01.013\%$& $ 0.95646 \pm 0.01016$& $ 36.350\% \pm 01.437\%$& $ 0.25105 \pm 0.01992$& $ 35.339\% \pm 01.542\%$ & $ 0.25096 \pm 0.01692$ \\								
														\hline 
														\hline
														%						\multirow{2}{*}{Multi-task}& MTL 1 & xxx\% & xxx\% & xxx\% &xxx\% &xxx\% &xxx\% &xxx\% &xxx\% \\
														%						&MTL 2 &xxx\% &xxx\% &xxx\% &xxx\% &xxx\% &xxx\% &xxx\% & \\
														\multirow{8}{*}{\begin{tabular}{@{}l@{}} Our Methods\\w.o BW\\ (Plant) \end{tabular}} 
														&CNN	&$ 98.550\% \pm 00.000\%$ &$ 0.98548\pm 0.00000$	&$ 97.780\% \pm 00.000\%$	&$ 0.97781 \pm 0.00000$	&$ 42.252\% \pm 00.621\%$	&$  0.39688\pm  0.00666$	&$ 38.390\% \pm 02.330\%$	&$  0.35370\pm 0.02055$ \\
														&AlexNet	&$ 98.925\% \pm 00.245\%$	&$  0.98928\pm 0.00243$	&$ 95.716\% \pm 00.805\%$	&$  0.95688\pm 0.00816$	&$ 37.942\% \pm 01.963\%$	&$  0.36431\pm  0.01388$	&$ 38.856\% \pm 03.014\%$	&$  0.35983\pm 0.03556$ \\
														&VGG16	&$ 99.511\% \pm 00.084\%$	&$  0.99512\pm 0.00085$	&$ 96.249\% \pm 00.583\%$	&$  0.96236\pm 0.00579$	&$ 44.621\% \pm 00.666\%$	&$  0.36913\pm  0.00923$	&$ 43.898\% \pm 03.495\%$	&$  0.35444\pm 0.02802$ \\
														&ResNet101	&$ 99.714\% \pm 00.052\%$	&$  0.99714\pm 0.00052$	&$ 98.069\% \pm 00.217\%$	&$  0.98066\pm 0.00217$	&$ 36.039\% \pm 04.150\%$	&$  0.29830\pm  0.03925$	&$ 36.610\% \pm 01.356\%$	&$ 0.27054 \pm 0.00194$ \\
														&EfficientNet	&$ 99.625\% \pm 00.037\%$	&$  0.99625\pm 0.00038$	&$ 95.816\% \pm 00.592\%$	&$  0.95798\pm 0.00596$	&$ 31.631\% \pm 04.450\%$	&$ 0.23771 \pm  0.03248$	&$ 33.898\% \pm 01.271\%$	&$  0.24964\pm 0.01032$ \\
														&InceptionV3	&$   \textbf{99.850\%} \pm  00.019\%$	&$ \textbf{0.99850}\pm 0.00019$	&$  98.690\% \pm   00.097\%$	&$ 0.98692\pm 0.00098$	&$ 46.757\% \pm 04.038\%$	&$ 0.45049 \pm  0.03376$	&$44.492\% \pm   06.196\%$	&$ 0.41757 \pm  0.05976$ \\
														&MobileNetV2	&$ 99.594\% \pm 00.119\%$	&$  0.99593\pm 0.00120$	&$ 95.627\% \pm 00.761\%$	&$  0.95682\pm 0.00727$	&$ 13.592\% \pm 00.000\%$	&$  0.03253\pm 0.20127 $	&$ 00.000\% \pm 09.110\%$	&$  0.07708\pm 0.05521$ \\
														& ViT &$ 90.282\% \pm 00.290\%$ &$ 0.90219\pm 0.00281 $ &$ 94.458\% \pm 00.472\%$	&$0.94426\pm 0.00470 $	&$ 32.136\% \pm 01.650\%$	&$ 0.25778\pm 0.02330$	&$30.085\% \pm 01.271\%$	&$  0.24128\pm 0.00673 $ \\
														
														%                 	       \hline
														%                 	       \hline 
														
														%    \multirow{2}{*}{\begin{tabular}{@{}l@{}} 
																%    %Our Methods\\ w. balance weights \\ + transfer learning\\
																%    0.4:0.5:0.1:0.1 \\
																% 	(Plant) \end{tabular}}				&InceptionV3	&$  99.687\% \pm  00.000\%$	&$   0.99688\pm 0.0000 $	&$  99.646\% \pm  00.000\%$	&$   0.99647\pm 0.0000 $	&$  49.068\% \pm  02.712\%$	&$  0.47960 \pm 0.01894  $	&$  46.864\% \pm  03.714\%$	&$  0.44804 \pm  0.03107$ \\
														
														\hline
														\hline
														\multirow{3}{*}{\begin{tabular}{@{}l@{}} Our Methods\\ with BW \\ (Plant) \end{tabular}} &&&&&&&&&\\
														&InceptionV3	&$  99.687\% \pm  00.000\%$	&$   0.99688\pm 0.0000 $	
														&$  \textbf{99.646\%}\pm  00.000\%$	&$  \textbf{0.99647} \pm 0.00000$	
														&$  49.262\% \pm  02.442\%$	&$  \textbf{0.48051} \pm 0.01716$	
														&$  \textbf{46.992}\% \pm  03.637\%$	&$  \textbf{0.44934} \pm 0.03012$ \\
														
														&&&&&&&&&\\
														
														\hline
														\hline 
														\multirow{3}{*}{\begin{tabular}{@{}l@{}}Our Methods\\ w. BW 
																+ TF\\
																(Plant) 			\end{tabular}}	&&&&&&&&&\\
														&InceptionV3 &$ 99.928\% \pm 00.012\%$ &$ 0.99927\pm 0.00012$	&$ 99.776\% \pm 00.227\%$	&$0.99777\pm 0.00226$ &$ 71.262\% \pm 03.962\%$ &$ 0.70602\pm 0.04275$ &$ 73.644\% \pm 02.548\%$ &$ 0.71980 \pm 0.02900$ \\			
														&&&&&&&&&\\
														
														\bottomrule%
													\end{tabular*}
												\end{sideways}
												%						}

											%		\end{center}
									\end{minipage}
								\end{table}
								
								\clearpage
								\restoregeometry

								\clearpage
								\newgeometry{margin=1cm}
								\thispagestyle{empty}
								
								\begin{table}[h!]
									%\sidewaystablefn%
									\centering
									\begin{minipage}{1.3\textwidth}
										%	\begin{center}
											
											%	\resizebox{\textwidth}{!}{%
												\caption{Plant Disease Results.}\label{Disease}
												%			\begin{minipage}{1.2\textwidth}
													%				\resizebox{\textwidth}{!}{%
														\begin{sideways}
															\begin{tabular*}{\textwidth}{@{\extracolsep{\fill}}ll|cc|cc|cc|cc@{\extracolsep{\fill}}}
																\toprule
																&& \multicolumn{2}{@{}c@{}}{Plant Village} & \multicolumn{2}{@{}c@{}}{Plant leaves}& \multicolumn{2}{@{}c@{}}{PlantDoc-0.2} & \multicolumn{2}{@{}c@{}}{PlantDoc-1.0}\\%
																\cmidrule{3-4}\cmidrule{5-6}\cmidrule{7-8}\cmidrule{9-10}%
																& & Acc & F1 & Acc & F1 & Acc & F1& Acc & F1 \\
																\midrule
																%%%%%%%%%%%%%%%%%%%%%%%%%%%%%%%%%%%%%%%%%%%%%%%%%%%%
																
																\multirow{8}{*}{\begin{tabular}{@{}l@{}} Multi-model\\ (Disease) \end{tabular}}
																&CNN	&$95.374\% \pm00.323\%$	&$0.95364 \pm0.00312$	 &$89.345\% \pm02.442\%$	&$0.89046 \pm0.02561$	&$33.146\% \pm01.573\%$	&$0.30879 \pm0.01138$	&$42.797\% \pm00.000\%$	&$0.36857 \pm0.00000$ \\
																&AlexNet	&$97.399\% \pm00.322\%$	&$0.97400 \pm0.00318$	&$90.165\% \pm00.543\%$	&$0.89938 \pm0.00537$	&$33.864\% \pm00.303\%$	&$0.30133 \pm0.01534$	&$41.059\% \pm00.898\%$	&$0.35199 \pm0.02495$ \\
																&VGG16	&$99.105\% \pm00.715\%$	&$0.99104 \pm0.00716$	&$93.042\% \pm00.472\%$	&$0.92978 \pm0.00452$	&$40.291\% \pm01.226\%$	&$0.33305 \pm0.02199$	&$44.237\% \pm01.887\%$	&$0.36279 \pm0.01651$ \\
																&ResNet101	&$98.828\% \pm00.058\%$	&$0.98825 \pm0.00056$	&$92.642\% \pm00.094\%$	&$0.92414 \pm0.00113$	&$31.204\% \pm01.139\%$	&$0.24170 \pm0.01724$	&$41.314\% \pm00.434\%$	&$0.29068 \pm0.00875$ \\
																&EfficientNet	&$98.561\% \pm00.074\%$	&$0.98558 \pm0.00076$	&$88.974\% \pm00.921\%$	&$0.88555 \pm0.01084$ &$31.320\% \pm00.990\%$& $0.25177\pm 0.00279$&	$39.873\% \pm00.481\%$	&$0.31227 \pm0.00996$ \\
																&InceptionV3	&$98.561\% \pm00.074\%$	&$0.98558 \pm0.00076$	&$97.170\% \pm00.000\%$	&$0.97164 \pm0.00000$	&$35.786\% \pm02.346\%$	&$0.30539 \pm0.02399$	&$\textbf{51.992\%} \pm03.272\%$	&$\textbf{0.48344} \pm0.05286$ \\
																&MobileNetV2	&$99.275\% \pm00.000\%$	&$0.99274 \pm0.00000$	&$87.807\% \pm01.868\%$	&$0.87907 \pm0.01975$	&$07.029\% \pm00.078\%$	&$0.00923 \pm0.00020$	&$08.475\% \pm00.000\%$	&$0.01324 \pm0.00000$ \\
																& ViT &$  90.314\% \pm  01.031\%$ &$ 0.89702\pm 0.01010 $ &$ 89.789\% \pm00.888\%$ &$ 0.89693 \pm 0.00833$&$30.680\% \pm 02.220\%$&$  0.18810\pm 0.03184$ &$ 38.701\% \pm 00.528\%$ &$ 0.25129 \pm 0.03408$ \\	
																\hline
																\hline
																\multirow{8}{*}{\begin{tabular}{@{}l@{}} Multi-label \\ (Disease) \end{tabular}}
																
																&CNN	&$96.748\% \pm00.023\%$	&$0.96752 \pm0.00025$	&$91.632\% \pm00.977\%$	&$0.91562 \pm 0.01059$	 &$35.184\% \pm00.534\%$	&$0.33953 \pm0.00126$	&$46.610\% \pm00.000\%$	&$0.42350 \pm0.00000$ \\
																&AlexNet	&$96.438\% \pm00.115\%$	&$0.96441 \pm0.00121$	&$92.431\% \pm00.381\%$	&$0.92444 \pm0.00312$	&$32.000\% \pm02.120\%$	&$0.30747 \pm0.01745$	&$41.864\% \pm04.797\%$	&$0.38276 \pm0.05842$ \\
																&VGG16	&$94.680\% \pm00.444\%$	&$0.94659 \pm0.00436$	&$88.380\% \pm00.736\%$	&$0.88040 \pm0.00741$	&$32.757\% \pm00.815\%$	&$0.29959 \pm0.00604$	&$39.958\% \pm03.137\%$	&$0.35054 \pm0.01112$ \\
																&ResNet101	&$98.926\% \pm00.083\%$	&$0.98927 \pm0.00083$	&$95.572\% \pm00.402\%$	&$0.95467 \pm0.00394$	&$30.155\% \pm03.440\%$	&$0.28404 \pm0.02025$	&$38.220\% \pm03.048\%$	&$0.34683 \pm0.01162$ \\
																&EfficientNet	&$ 98.087\% \pm 00.183\%$ &$ 0.98083 \pm 0.00185$ &$91.188\% \pm00.089\%$	&$0.90920 \pm0.00027$	&$26.777\% \pm01.800\%$	&$0.26184 \pm0.01083$	&$32.203\% \pm05.932\%$	&$0.30440 \pm0.04113$ \\
																&InceptionV3	&$99.383\% \pm00.099\%$	&$0.99384 \pm0.00099$	&$97.547\% \pm00.033\%$	&$0.97468 \pm0.00029$	&$39.573\% \pm01.218\%$	&$0.38842 \pm0.00837$	&$46.483\% \pm00.381\%$	&$0.43180 \pm0.00194$ \\
																&MobileNetV2	&$98.571\% \pm00.064\%$	&$0.98569 \pm0.00064$	&$91.787\% \pm01.372\%$	&$0.91625 \pm0.01531$	&$07.146\% \pm00.078\%$	&$0.00953 \pm0.00020$	&$05.424\% \pm01.017\%$	&$0.00575 \pm0.00250$ \\
																& ViT &$  90.052\% \pm 00.668 \%$ &$ 0.89733\pm 0.00664 $ &$ 92.675\% \pm 00.222\%$&$ 0.92528\pm 0.00239$&$26.796\% \pm 00.000\%$&$0.24090\pm 0.00985$&$ 38.983\% \pm 00.847\%$&$0.32884 \pm 0.01774$ \\	
																\hline
																\hline
																\multirow{8}{*}{\begin{tabular}{@{}l@{}} Multi-output \\ (Disease) \end{tabular}}
																&CNN	&$94.898\% \pm00.493\%$	&$0.94882 \pm0.00487$	&$89.467\% \pm00.948\%$	&$0.89130 \pm0.01004$	&$39.592\% \pm01.452\%$	&$0.35808 \pm0.01346$	&$41.314\% \pm00.636\%$	&$0.39240 \pm0.00270$ \\
																&AlexNet	&$98.058\% \pm00.127\%$	&$0.98056 \pm0.00127$	&$92.508\% \pm00.114\%$	&$0.92443 \pm0.00163$	&$36.621\% \pm01.807\%$	&$0.33630 \pm0.02018$	&$47.034\% \pm00.000\%$	&$0.44287 \pm0.00000$ \\
																&VGG16	&$98.643\% \pm00.042\%$	&$0.98642 \pm0.00044$	&$94.839\% \pm01.049\%$	&$0.94706 \pm0.01153$	&$42.252\% \pm00.816\%$ &$0.35863 \pm0.01492$	&$48.517\% \pm01.907\%$	&$0.39686 \pm0.03644$ \\
																&ResNet101	&$99.021\% \pm00.049\%$	&$0.99020 \pm0.00049$	&$95.627\% \pm00.466\%$	&$0.95588 \pm0.00429$	&$35.730\% \pm00.000\%$	&$0.29591 \pm0.00000$	&$41.949\% \pm00.000\%$	&$0.32134 \pm0.00000$ \\
																&EfficientNet &$ 98.539\% \pm 00.103\%$ &$ 0.98539 \pm 0.00103$ &$91.831\% \pm01.184\%$	&$0.91649 \pm0.01190$	&$30.583\% \pm00.558\%$ &$0.24967 \pm0.00594$	&$38.644\% \pm01.017\%$  &$0.26158 \pm0.01855$ \\
																&InceptionV3 &$ 99.414\% \pm 00.056\%$ &$ 0.99413 \pm 0.00056$ &$97.414\% \pm00.494\%$	&$0.97330 \pm0.00540$	&$43.165\% \pm00.089\%$	&$0.41892 \pm0.00977$	&$46.441\% \pm03.239\%$	&$0.40202 \pm0.03922$ \\
																&MobileNetV2 &$ 98.581\% \pm 00.207\%$ &$ 0.98577 \pm 0.00206$ &$91.931\% \pm00.740\%$	&$0.91873 \pm0.00750$	&$05.553\% \pm01.068\%$	&$0.00604 \pm0.00235$	&$07.881\% \pm01.780\%$	&$0.01204 \pm0.00359$ \\
																& ViT &$  90.553\% \pm 00.928\%$ &$ 0.90458 \pm 0.00712$ &$ 86.733\% \pm 01.946\%$	&$  0.86373 \pm 0.02181 $	&$ 31.456\% \pm 00.000\%$	&$  0.15054\pm 0.00000$	&$ 0.38136
																0.0
																\% \pm \%$	&$ 0.21056 \pm 0.00000 $ \\	
																\hline 
																\hline
																\multirow{4}{*}{\begin{tabular}{@{}l@{}} Multi-task \\ (Disease) \end{tabular} }
																& Cross-Stitch & $95.241\% \pm 00.169\%$ & $0.95237 \pm 0.00179$ &$ 93.785\% \pm 00.248\%$	&$ 0.93739 \pm  0.00276$	&$ 40.466\% \pm 00.995\%$	&$ 0.35972 \pm 0.00859$	&$ 44.492\% \pm 01.431\%$	&$ 0.38920 \pm 0.01102$ \\								
																&MTAN & $91.410\% \pm  00.710\%$ & $  0.91411 \pm 0.00702$ &$ 87.148\% \pm 02.273\%$	&$ 0.86893 \pm 0.02296$ &$ 30.971\% \pm 01.075\%$ &$ 0.20637 \pm 0.01803$	&$ 38.941\% \pm 01.079\%$	&$ 0.25224 \pm 0.01672$ \\
																
																&TSNs & $79.926\% \pm  02.681\%$ & $  0.79900 \pm   0.02653$ &$ 81.953\% \pm 03.986\%$	&$ 0.81046 \pm 0.04324$	&$ 30.039\% \pm 01.139\%$	&$ 0.19396 \pm 0.01471$	&$ 37.627\% \pm 01.254\%$	&$ 0.25399 \pm 0.02069$ \\
																&MOON & $95.307\% \pm  00.889\%$ & $  0.95296 \pm   0.00886$ &$ 92.986\% \pm 01.060\%$	&$ 0.92804 \pm 0.01191$	&$ 34.757\% \pm 01.074\%$ &$ 0.22978 \pm 0.02701$	&$ 41.568\% \pm 01.485\%$	&$ 0.31160 \pm 0.02726$ \\

																%&TSNs & $\% \pm  \%$ & $  \% \pm   \%$ &$ \% \pm \%$	&$ \% \pm \%$	&$ \% \pm \%$	&$ \% \pm \%$	&$ \% \pm \%$	&$ \% \pm \%$ \\
																\hline 
																\hline
																\multirow{8}{*}{\begin{tabular}{@{}l@{}} Our Methods\\ w.o BW\\ (Disease) \end{tabular}} 
																&CNN	&$ 96.363\% \pm 00.000\%$	&$  0.96352\pm 0.00000$	&$ 95.006\% \pm 00.000\%$	&$  0.94992\pm 0.0000$	&$ 42.058\% \pm 00.932\%$	&$  0.38199\pm  0.00462$	&$ 46.229\% \pm 01.359\%$	&$  0.41789\pm 0.01399$ \\
																&AlexNet	&$ 97.380\% \pm 00.496\%$	&$  0.97376\pm 0.00496$	&$ 92.519\% \pm 00.928\%$	&$  0.92387\pm 0.00963$	&$ 37.631\% \pm 00.756\%$	&$  0.30592\pm  0.03019$	&$ 44.619\% \pm 02.218\%$	&$ 0.39379 \pm 0.02078$ \\
																&VGG16	&$ 98.716\% \pm 00.139\%$	&$  0.98713\pm 0.00142$	&$ 95.061\% \pm 00.259\%$	&$  0.94922\pm 0.00299$	&$ 42.718\% \pm 00.951\%$	&$  0.34025\pm 0.01214 $	&$ 45.636\% \pm 00.194\%$	&$ 0.37864 \pm 0.01234$ \\
																&ResNet101	&$ 98.943\% \pm 00.056\%$	&$  0.98943\pm 0.00056$	&$ 95.461\% \pm 00.674\%$	&$  0.95442\pm 0.00664$	&$ 32.893\% \pm 01.054\%$	&$  0.23341\pm  0.00812$	&$ 38.729\% \pm 00.339\%$	&$  0.29411\pm 0.01903$ \\
																&EfficientNet	&$ 98.639\% \pm 00.034\%$	&$  0.98642\pm  0.00035$	&$ 92.786\% \pm 00.880\%$	&$  0.92687\pm 0.00850$	&$ 30.194\% \pm 02.639\%$	&$  0.24793\pm  0.02109$	&$ 40.466\% \pm 00.212\%$	&$ 0.33290 \pm 0.00744$ \\
																&InceptionV3	&$   99.418\% \pm   00.029\%$	&$ 0.99417\pm 0.00030  $	&$   97.292\% \pm 00.546\%$	&$ 0.97201\pm 0.00608  $ & $ 45.029\% \pm 03.134\%$ &$ 0.41716\pm 0.03430 $ &$ 47.542\% \pm  04.664\%$	&$ 0.46156\pm 0.03985$ \\
																&MobileNetV2	&$ 98.645\% \pm 00.335\%$	&$  0.98648\pm 0.00329$	&$ 91.720\% \pm 01.414\%$	&$  0.91710\pm 0.01342$	&$ 07.010\% \pm 00.058\%$	&$  0.00918\pm 0.00015 $	&$ 06.356\% \pm 01.271\%$	&$  0.00787\pm 0.00294$ \\
																& ViT &$ 86.728\% \pm 00.087\%$ &$ 0.86416\pm 0.00183 $ &$ 90.330\% \pm 00.354\%$	& $0.90116\pm 0.00410 $	&$ 32.913\% \pm 00.291\%$ &$ 0.24160\pm 0.01754$	&$ 38.983\% \pm 00.424\%$ &$ 0.30192\pm 0.00490$ \\				
																\hline
																\hline
																% 	\rule{0pt}{25pt}Our Methods w. BW (Disease)		
																\multirow{3}{*}{\begin{tabular}{@{}l@{}} Our Methods\\ with BW \\ (Disease) \end{tabular}} &&&&&&&&&\\
																&InceptionV3 &$  \textbf{99.466\%} \pm  00.000\%$	&$   \textbf{0.99466}\pm  0.0000$	
																&$  \textbf{97.759\%} \pm  00.000\%$	&$   \textbf{0.97738}\pm 0.0000 $	
																&$  \textbf{45.301\%} \pm 03.939 \%$	&$   \textbf{0.43104}\pm 0.03197 $	
																&$  50.212\% \pm  02.584\%$	&$   0.48154\pm 0.03108 $ \\
																
																&&&&&&&&&\\
																
																\hline
																\hline 
																\multirow{3}{*}{\begin{tabular}{@{}l@{}}Our Methods\\ w. BW 
																		+ TF\\
																		(Disease) 			\end{tabular}}	&&&&&&&&&\\
																&InceptionV3 &$ 99.615\% \pm 00.049\%$ &$ 0.99615\pm 0.00049  $	&$ 98.149\% \pm 00.167\%$ &$ 0.98146\pm 0.00173 $ &$ 64.544\% \pm 03.089\%$ &$ 0.62968\pm 0.03476 $ &$ 63.305\% \pm 00.786\%$	&$ 0.63757\pm 0.00953$ \\
																
																&&&&&&&&&\\
																\hline
																\bottomrule%
															\end{tabular*}
														\end{sideways}
														%						}

													%		\end{center}
											\end{minipage}
										\end{table}
										
										\clearpage
										\restoregeometry
										
										\clearpage
										\newgeometry{margin=1cm}
										\thispagestyle{empty}
										
										\begin{table}[h!]
											%\sidewaystablefn%
											\centering
											\begin{minipage}{1.3\textwidth}
												%	\begin{center}
													
													%	\resizebox{\textwidth}{!}{%
														\caption{Both (Type \& Disease) Results. }\label{Total}
														%			\begin{minipage}{1.2\textwidth}
															%				\resizebox{\textwidth}{!}{%
																\begin{sideways}
																	\begin{tabular*}{\textwidth}{@{\extracolsep{\fill}}ll|cc|cc|cc|cc@{\extracolsep{\fill}}}
																		\toprule
																		&& \multicolumn{2}{@{}c@{}}{Plant Village} & \multicolumn{2}{@{}c@{}}{Plant leaves}& \multicolumn{2}{@{}c@{}}{PlantDoc-0.2} & \multicolumn{2}{@{}c@{}}{PlantDoc-1.0}\\%
																		\cmidrule{3-4}\cmidrule{5-6}\cmidrule{7-8}\cmidrule{9-10}%
																		& & Acc & F1 & Acc & F1 & Acc & F1& Acc & F1 \\
																		\midrule
																		%%%%%%%%%%%%%%%%%%%%%%%%%%%%%%%%%%%%%%%%%%%%%%%%%%%%
																		\multirow{8}{*}{\begin{tabular}{@{}l@{}} Multi-model \\ (Both) \end{tabular}}
																		
																		&CNN	&$93.273\% \pm00.396\%$	&$0.94202 \pm0.00310$	&$85.283\% \pm03.019\%$	&$0.86703 \pm0.02708$	&$11.456\% \pm00.583\%$	&$0.12928 \pm0.00427$	&$12.712\% \pm00.000\%$	&$0.11760 \pm0.00000$ \\
																		&AlexNet & $96.275\% \pm00.582\%$ 
																		&$0.96831 \pm0.00447$	&$88.007\% \pm00.971\%$	&$0.88788 \pm0.00804$	&$12.117\% \pm01.836\%$	&$0.12140 \pm0.03146$	&$14.746\% \pm02.678\%$	&$0.12720 \pm0.02971$ \\
																		&VGG16	&$98.634\% \pm00.978\%$	&$0.98852 \pm0.00827$	&$90.778\% \pm00.283\%$	&$0.92730 \pm0.00409$	&$16.252\% \pm01.648\%$	&$0.15821 \pm0.02069$	&$14.746\% \pm01.150\%$	&$0.12209 \pm0.01727$ \\
																		&ResNet101	&$98.295\% \pm00.065\%$	&$0.98524 \pm0.00052$	&$91.698\% \pm00.094\%$	&$0.92189 \pm0.00087$	&$05.612\% \pm00.566\%$	&$0.03379 \pm0.00847$	&$05.805\% \pm00.601\%$	&$0.03731 \pm0.00309$ \\
																		&EfficientNet	&$98.034\% \pm00.000\%$	&$0.98301 \pm0.00029$	&$85.814\% \pm00.592\%$	&$0.87085 \pm0.00535$	&$07.553\% \pm00.058\%$	&$0.06758 \pm0.00492$	&$07.839\% \pm01.820\%$	&$0.05904 \pm0.00971$ \\
																		&InceptionV3	&$98.034\% \pm00.000\%$	&$0.98301 \pm0.00029$	& $96.4623\% \pm00.000\%$	&$0.97142 \pm0.00000$	&$14.194\% \pm00.421\%$	&$0.13135 \pm0.00396$	&$20.212\% \pm02.924\%$	&$0.19346 \pm0.04271$ \\
																		&MobileNetV2	&$99.050\% \pm00.000\%$	&$0.99148 \pm0.00000$	&$85.307\% \pm01.726\%$	&$0.87486 \pm0.01687$	&$04.078\% \pm01.553\%$	&$0.00361 \pm0.00301$	&$00.847\% \pm01.695\%$	&$0.00069 \pm0.00138$ \\
																		& ViT &$  84.481\% \pm  01.864\%$ &$ 0.85548\pm 0.01771 $ &$ 86.238\% \pm 00.999\%$	&$  0.87549\pm 0.00578$	&$ 03.883\% \pm 01.040\%$ &$ 0.02103\pm 0.00969$	&$04.802\% \pm 00.871\%$	&$  0.01652\pm 0.00460 $ \\	
																		\hline
																		\hline
																		\multirow{8}{*}{\begin{tabular}{@{}l@{}} Multi-label \\ (Both) \end{tabular}}							
																		&CNN	&$96.537\% \pm00.057\%$	&$0.96542 \pm0.00062$	&$89.523\% \pm01.655\%$	&$0.89448 \pm0.01606$	&$20.680\% \pm00.445\%$	&$0.18933 \pm0.00913$	&$20.763\% \pm00.000\%$	&$0.16217 \pm0.00000$ \\
																		&AlexNet	&$96.150\% \pm00.172\%$	&$0.96156 \pm0.00175$	&$91.099\% \pm00.652\%$	&$0.90945 \pm0.00583$	&$18.000\% \pm00.772\%$	&$0.15703 \pm0.01071$	&$19.195\% \pm04.383\%$	&$0.15931 \pm0.04658$ \\
																		&VGG16	&$94.350\% \pm00.433\%$	&$0.94316 \pm0.00444$	&$85.527\% \pm00.743\%$	&$0.85292 \pm0.00795$	&$16.854\% \pm00.986\%$	&$0.14534 \pm0.00831$	&$13.432\% \pm01.748\%$	&$0.09539 \pm0.02077$ \\
																		&ResNet101	&$98.841\% \pm00.070\%$	&$0.98841 \pm0.00071$	&$94.684\% \pm00.230\%$	&$0.94643 \pm0.00240$	&$16.000\% \pm00.956\%$	&$0.12609 \pm0.02183$	&$11.864\% \pm00.709\%$	&$0.07009 \pm0.00548$ \\
																		&EfficientNet & $ 97.923\% \pm00.204\%$ &	$ 0.97908 \pm 0.00207$ &$89.789\% \pm00.444\%$	&$0.89457 \pm0.00349$	&$12.913\% \pm00.806\%$	&$0.09489 \pm0.00599$	&$08.983\% \pm00.678\%$	&$0.04257 \pm0.00800$ \\
																		&InceptionV3	&$99.310\% \pm00.171\%$	&$0.99311 \pm0.00173$	&$97.114\% \pm00.000\%$	&$0.97077 \pm0.00000$	&$22.971\% \pm01.129\%$	&$0.21467 \pm0.00871$	&$\textbf{25.466\%} \pm00.127\%$	&$0.21130 \pm0.00654$ \\
																		&MobileNetV2	&$98.483\% \pm00.023\%$	&$0.98481 \pm0.00021$	&$90.122\% \pm01.469\%$	&$0.90013 \pm0.01574$	&$06.408\% \pm01.553\%$	&$0.00813 \pm0.00301$	&$05.000\% \pm00.254\%$	&$0.00477 \pm0.00044$ \\
																		& ViT &$87.699\% \pm 00.571\%$ &$ 0.87090\pm  0.00500$ &$ 0.90899\% \pm 00.333\%$ &$ 0.90881\pm 0.00392 $ &$ 13.592\% \pm 00.000\%$	&$  0.10092 \pm 0.00978$ &$ 11.864 \% \pm 00.000\%$	&$ 0.07795\pm 0.01533$ \\	
																		\hline
																		\hline
																		\multirow{8}{*}{\begin{tabular}{@{}l@{}} Multi-output \\ (Both) \end{tabular}}	
																		& CNN	&$93.208\% \pm00.569\%$	&$0.93766 \pm0.00483$	&$85.094\% \pm01.141\%$	&$0.86082 \pm0.01036$	&$14.078\% \pm00.701\%$	&$0.15213 \pm0.00644$	&$12.288\% \pm01.271\%$	&$0.11036 \pm0.01495$ \\
																		&AlexNet	&$97.719\% \pm00.131\%$	&$0.97855 \pm0.00121$	&$90.777\% \pm00.578\%$	&$0.90887 \pm 0.00518$	&$16.272\% \pm01.319\%$	&$0.16022 \pm0.01750$	&$22.458\% \pm00.000\%$	&$0.20815 \pm0.00000$ \\
																		&VGG16	&$98.258\% \pm00.085\%$	&$0.98359 \pm0.00066$	&$92.941\% \pm01.441\%$	&$0.93384 \pm0.01352$	&$17.767\% \pm01.456\%$	&$0.15111 \pm0.01682$	&$16.737\% \pm02.331\%$	&$0.11670 \pm0.02365$ \\
																		&ResNet101	&$98.789\% \pm00.071\%$	&$0.98843 \pm0.00053$	&$94.295\% \pm00.466\%$	&$0.94471 \pm0.00399$	&$09.126\% \pm00.000\%$	&$0.06699 \pm0.00000$	&$11.441\% \pm00.000\%$	&$0.05938 \pm0.00000$ \\
																		&EfficientNet &$ 98.195\% \pm 00.162\%$ &$ 0.98290 \pm 0.00135$ &$89.734\% \pm01.825\%$	&$0.89997 \pm0.01753$	&$08.738\% \pm01.387\%$	&$0.06837 \pm0.00493$	&$06.737\% \pm00.127\%$	&$0.03745 \pm0.00025$ \\
																		&InceptionV3 &$ 99.215\% \pm 00.017\%$ &$ 0.99239 \pm  0.00018$ &$96.493\% \pm00.519\%$	&$0.96693 \pm0.00536$	&$23.650\% \pm00.356\%$	&$0.23228 \pm0.01445$	 &$18.136\% \pm04.308\%$	&$0.14635 \pm0.03692$ \\
																		&MobileNetV2 & $98.309\% \pm 00.284\%$ & $ 0.98401 \pm  0.00276$ &$89.234\% \pm01.277\%$	&$0.89726 \pm0.01118$	&$02.155\% \pm03.292\%$	&$0.00289 \pm0.00441$	&$02.119\% \pm02.119\%$	&$0.00172 \pm0.00172$ \\
																		& ViT &$  87.276\% \pm 01.327\%$ &$ 0.87476\pm 0.01048$ &$ 82.842\% \pm 02.300\%$ &$ 0.82853\pm 0.02119$	&$ 01.942\% \pm 00.000\%$	&$0.00074 \pm 0.00000$&$ 03.390\% \pm 00.000\%$ &$ 0.00222\pm 0.00000$ \\	
																		
																		\hline 
																		\hline

																		\multirow{4}{*}{\begin{tabular}{@{}l@{}} Multi-task \\ (Both) \end{tabular} }
																		& Cross-Stitch & $93.870\% \pm 00.310\%$ & $  0.94509 \pm   0.00276$ & $ 91.356\% \pm 00.106\%$	&$ 0.92334 \pm 0.00091$	&$ 17.068\% \pm 00.629\%$	&$ 0.17532 \pm 0.00541$	&$ 16.610\% \pm 02.241\%$	&$ 0.15028 \pm 0.01913$ \\								
																		&MTAN & $89.509\% \pm  00.898\%$ & $  0.90269 \pm 0.00836$ &$ 82.420\% \pm 03.549\%$	&$ 0.83209 \pm 0.03344$	&$ 04.175\% \pm 01.210\%$	&$ 0.02986 \pm 0.01344$	&$ 06.102\% \pm 00.988\%$	&$ 0.02871 \pm 0.00839$ \\
																		&TSNs & $76.808\% \pm  03.190\%$ & $  0.77433 \pm   0.03101$ &$ 74.673\% \pm 05.778\%$	&$ 0.75099 \pm 0.05800$	&$ 05.029\% \pm 01.179\%$	&$ 0.02603 \pm 0.01110$	&$ 05.975\% \pm 01.545\%$	&$ 0.02682 \pm 0.01049$ \\
																		&MOON & $94.414\% \pm  00.997\%$ & $  0.94586 \pm   0.00974$ &$ 90.744\% \pm 01.470\%$	&$ 0.90727 \pm 0.01542$	&$ 08.272\% \pm 01.571\%$	&$ 0.04720 \pm 0.01639$	&$ 09.068\% \pm 01.506\%$	&$ 0.04485 \pm 0.01521$ \\
																		
																		%&TSNs & $\% \pm  \%$ & $  \% \pm   \%$ &$ \% \pm \%$	&$ \% \pm \%$	&$ \% \pm \%$	&$ \% \pm \%$	&$ \% \pm \%$	&$ \% \pm \%$ \\

																		\hline 
																		\hline
																		\multirow{8}{*}{\begin{tabular}{@{}l@{}} Our Methods\\w.o BW\\ (Both) \end{tabular}} 
																		&CNN	&$ 95.500\% \pm 00.000\%$	&$  0.95741\pm 0.00000$	&$ 93.563\% \pm 00.000\%$	&$ 0.93847 \pm 0.00000$	&$ 21.748\% \pm 01.553\%$	&$  0.21419\pm 0.01257 $	&$ 18.178\% \pm 02.524\%$	&$  0.16272\pm 0.02979$ \\
																		&AlexNet	&$ 96.968\% \pm 00.588\%$	&$  0.97061\pm 0.00561$	&$ 90.765\% \pm 01.310\%$	&$  0.90818\pm 0.01255$	&$ 15.553\% \pm 02.203\%$	&$  0.15499\pm  0.02119$	&$ 17.246\% \pm 01.229\%$	&$  0.15248\pm 0.01697$ \\
																		&VGG16	&$ 98.395\% \pm 00.198\%$ &$  0.98445\pm 0.00189$	&$ 93.219\% \pm 00.644\%$	&$  0.93439\pm 0.00628$	&$ 16.000\% \pm 01.332\%$	&$  0.14380\pm  0.01274$	&$ 17.119\% \pm 02.330\%$	&$  0.11884\pm 0.01286$ \\
																		&ResNet101	&$ 98.773\% \pm 00.053\%$	&$  0.98796\pm 0.00056$	&$ 94.550\% \pm 00.737\%$	&$  0.94664\pm 0.00664$	&$ 09.825\% \pm 01.492\%$	&$ 0.07677 \pm  0.01649$	&$ 10.254\% \pm 00.678\%$	&$  0.06014\pm 0.00730$ \\
																		&EfficientNet	&$ 98.445\% \pm 00.052\%$	&$  0.98482\pm 0.00060$	&$ 90.877\% \pm 00.980\%$	&$  0.90985\pm 0.00847$	&$ 08.039\% \pm 01.318\%$	&$  0.06106\pm  0.01903$	&$ 09.322\% \pm 00.424\%$	&$  0.07318\pm 0.00716$ \\
																		&InceptionV3	&$\textbf{99.315\%} \pm 00.037\%$ &$ \textbf{0.99333}\pm 0.00037$	&$   96.593\% \pm   00.447\%$	&$ 0.96641\pm 0.00457 $	&$ 25.359\% \pm 02.170\%$	&$  0.24692\pm  0.02342   $& $ 23.814\% \pm 02.910\%$	&$ 0.22487\pm 0.03326 $ \\
																		&MobileNetV2	&$ 98.408\% \pm 00.404\%$	&$ 0.98445 \pm 0.00392$	&$ 89.989\% \pm 01.468\%$	&$  0.90320\pm 0.01200$	&$ 03.689\% \pm 01.165\%$	&$  0.00286\pm 0.00226$	&$ 04.661\% \pm 00.424\%$	&$  0.00418\pm 0.00074$ \\
																		& ViT &$ 83.003\% \pm 00.064\%$ &$ 0.83316\pm 0.00037 $ &$87.913\% \pm 00.295\%$ &$0.87931\pm 0.00502 $	&$ 08.447\% \pm 01.262\%$	&$ 0.07872 \pm 0.01946$	&$ 06.780\% \pm 00.424\%$	&$  0.05606\pm 0.00594 $ \\				
																		
																		\hline
																		\hline
																		% 	\rule{0pt}{25pt}Our Methods w. BW (Disease)		
																		\multirow{3}{*}{\begin{tabular}{@{}l@{}} Our Methods\\ with BW \\ (Both) \end{tabular}} &&&&&&&&&\\
																		&InceptionV3	&$  99.208\% \pm  00.000\%$	&$   0.99245\pm 0.0000 $	&$ \textbf{97.524\%} \pm  00.000\%$	&$   \textbf{0.97746}\pm 00.000 $	
																		&$  \textbf{26.175\%} \pm 03.008\%$	&$  \textbf{0.26760} \pm 0.03055  $	
																		&$  24.153\% \pm  02.584\%$	&$   \textbf{0.23191}\pm 0.02879 $ \\
																		
																		&&&&&&&&&\\
																		
																		\hline
																		\hline 
																		\multirow{3}{*}{\begin{tabular}{@{}l@{}}Our Methods\\ w. BW 
																				+ TF\\
																				(Both) 			\end{tabular}}	&&&&&&&&&\\
																		&InceptionV3 &$ 99.558\% \pm 00.055\%$ &$ 0.99565\pm 0.00064  $ &$ 98.042\% \pm 00.270\%$ &$ 0.98116 \pm 0.00196 $ &$ 50.291\% \pm   03.895\%$ &$ 0.50396 \pm 0.04018 $ &$ 50.000\% \pm 02.023\%$ &$ 0.50191 \pm 0.02036 $ \\		
																		&&&&&&&&&\\

																		\bottomrule%
																	\end{tabular*}
																\end{sideways}
																%						}

															%		\end{center}
													\end{minipage}
												\end{table}
												
												\clearpage
												\restoregeometry

												\subsection{Accuracy, F1-score and False Positive Rates of the Optimums}			
												\ST{Table \ref{tab:FPR} offers a comprehensive performance evaluation of various approaches' optimal models across diverse datasets. All models leverage Inception V3 backbones, recognised as the top-performing backbone in Table \ref{Type}, \ref{Disease}, and \ref{Total}. Within the table, we present accuracy (Acc), F1-score (F1), and false positive rate (FPR) for each model. In the context of accuracy and F1-score, higher values are indicative of better performance, while for FPR, closer values to 0 will be better performance. Notably, our methods, specifically the classifier chain with optimised balance weights and the transfer learning approach (the last row), demonstrate their superiority in this table. We can see that it achieves accuracy and F1-scores close to 1.0 in both the Plant Village and Plant Leaves datasets. It also performs exceptionally well in the two datasets of PlantDoc, achieving over 0.7 for plant species classification (Plant) and over 0.6 for leaf disease classification (Dis). And the overall classification (Both) exceeds 0.5 in both Accuracy and F1-score. In terms of FPR, each column's values are significantly close to 0 compared to all other rows, highlighting the exceptional robustness of this model. }
												
												\ST{In summary, this supremacy underscores the effectiveness of incorporating the classifier chain structure, optimising weights through grid-search, utilising transfer learning from ImageNet weights, and adopting Inception V3 backbones. These techniques collectively result in a more robust and precise classification system. Furthermore, these methods consistently maintain relatively low FPR, underscoring their capability to minimise misclassifications. As we focus on the rows related to the classifier chain in the table (the last three rows), the effectiveness of various methods becomes evident. Their accuracy and F1-scores exhibit a progressive increase, while the FPR consistently decreases. Ultimately, this amalgamation leads to a heightened level of model effectiveness, that may render them an invaluable choice for practical applications in smart agriculture and plant health management.}
												\begin{table}[H]
													\centering
													\caption{The Optimums' Accuracy, F1-score \& False Positive Rates.}
													\label{tab:FPR}
													\resizebox{\textwidth}{!}{%
														\begin{tabular}{cc|c|c|c|c|c|c|c|c|c|c|c|c}
															\hline
															% \multirow{2}{*}{\textbf{Approaches}} & \multirow{2}{*}{\textbf{Metric}} & \multicolumn{3}{c|}{\textbf{Plant Village}} & \multicolumn{3}{c|}{\textbf{Plant Leaves}} & \multicolumn{3}{c|}{\textbf{Plant Doc 0.2}} & \multicolumn{3}{c}{\textbf{Plant Doc Original}} \\
															% \cline{3-14}
															% &  & \textbf{Plant} & \textbf{Dis} & \textbf{Both} & \textbf{Plant} & \textbf{Dis} & \textbf{Both} & \textbf{Plant} & \textbf{Dis} & \textbf{Both} & \textbf{Plant} & \textbf{Dis} & \textbf{Both} \\
															\multirow{2}{*}{Approaches} & \multirow{2}{*}{Metric} & \multicolumn{3}{c|}{Plant Village} & \multicolumn{3}{c|}{Plant Leaves} & \multicolumn{3}{c|}{Plant Doc 0.2} & \multicolumn{3}{c}{Plant Doc Original} \\
															\cline{3-14}
															&  & Plant & Dis & Both & Plant & Dis & Both & Plant & Dis & Both & Plant & Dis & Both \\
															\hline		
															\multirow{3}{*}{Multi-model} & Acc & 0.99750 & 0.99275 & 0.99050 & 0.99175 & 0.97170 & 0.96462 & 0.44854 & 0.36311 & 0.14563 & 0.44492 & 0.50847 & 0.21186 \\
															& F1 & 0.99750 & 0.99274 & 0.99148 & 0.99175 & 0.97164 & 0.97142 & 0.36914 & 0.30423 & 0.13374 & 0.43324 & 0.47997 & 0.21345 \\
															& FPR & 0.00033 & 0.00039 & 0.00022 & 0.00093 & 0.00373 & 0.00118 & 0.05168 & 0.04737 & 0.01723 & 0.04504 & 0.03634 & 0.01221 \\
															\hline
															\multirow{3}{*}{Multi-label} & Acc & 0.99834 & 0.99558 & 0.99484 & 0.99112 & 0.97558 & 0.97114 & 0.33786 & 0.37282 & 0.21748 & 0.41949 & 0.47034 & 0.24576 \\
															& F1 & 0.99834 & 0.99559
															& 0.99485 & 0.99111 & 0.97477 & 0.97077 & 0.32529 & 0.33752 & 0.17350 & 0.41247 & 0.46042 & 0.21948 \\
															& FPR & 0.00013 & 0.00023 & 0.00014 & 0.00080 & 0.00361 & 0.00137 & 0.05330 & 0.04435 & 0.02964 & 0.04725 & 0.03855 & 0.02834 \\
															\hline
															\multirow{3}{*}{Multi-output} & Acc & 0.99888 & 0.99388 & 0.99288 & 0.98890 & 0.96670 & 0.96004 & 0.52427 & 0.43107 & 0.23883 & 0.44492 & 0.51271 & 0.19915 \\
															& F1 & 0.99887 & 0.99388 & 0.99319 & 0.98893 & 0.96616 & 0.96127 & 0.48730 & 0.42531 & 0.24174 & 0.39350 & 0.45603 & 0.17277 \\
															& FPR & 0.00017 & 0.00032 & 0.00020 & 0.00104 & 0.00437 & 0.00167 & 0.04249 & 0.03991 & 0.01745 & 0.04800 & 0.03861 & 0.02533 \\
															\hline
															\multirow{3}{*}{\begin{tabular}{@{}l@{}} Our Methods\end{tabular}} & Acc & 0.99838 & 0.99400 & 0.99300 & 0.99112 & 0.97336 & 0.96559 & 0.46214 & 0.44854 & 0.28155 & 0.51695 & 0.50424 & 0.25424 \\
															& F1 & 0.99837 & 0.99400 & 0.99319 & 0.99119 & 0.97201 & 0.96599 & 0.44974 & 0.41720 & 0.25822 & 0.47412 & 0.48480 & 0.24085 \\
															& FPR & 0.00025 & 0.00032 & 0.00020 & 0.00081 & 0.00414 & 0.00149 & 0.04403 & 0.03973 & 0.01491 & 0.04185 & 0.03658 & 0.02038 \\
															\hline
															\multirow{3}{*}{\begin{tabular}{@{}l@{}} Our Methods\\ with BW \end{tabular}} & Acc & 0.99725 & 0.99450 & 0.99225 & 0.99646 & 0.97759 & 0.97524 & 0.48932 & 0.50874 & 0.31456 & 0.53814 & 0.53390 & 0.31356 \\
															& F1 & 0.99726 & 0.99449 & 0.99248 & 0.99647 & 0.97738 & 0.97746 & 0.51162 & 0.47363 & 0.33094 & 0.51138 & 0.50913 & 0.28519 \\
															& FPR & 0.00037 & 0.00030 & 0.00022 & 0.00039 & 0.00327 & 0.00107 & 0.04121 & 0.03774 & 0.01443 & 0.03818 & 0.03411 & 0.01660 \\
															\hline
															% \multirow{3}{*}{\begin{tabular}{@{}l@{}}Our Methods\\ w. BW 
																	% 		+ TF\end{tabular}} & Acc & 0.99913 & 0.99650 & 0.99600 & 1.00000 & 0.98231 & 0.98231 & 0.76505 & 0.65049 & 0.55340 & 0.76271 & 0.62712 & 0.51271 \\
															% & F1 & 0.99912 & 0.99650 & 0.99625 & 1.00000 & 0.98237 & 0.98225 & 0.76336 & 0.63742 & 0.54967 & 0.75084 & 0.63908 & 0.50652 \\
															% & FPR & 0.00013 & 0.00019 & 0.00011 & 0.00000 & 0.00217 & 0.00094 & 0.02015 & 0.02508 & 0.01113 & 0.02011 & 0.02545 & 0.01304 \\
															
															% \bfseries \multirow{3}{*}{\begin{tabular}{@{}l@{}}Our Methods\\ w. BW 
																	% + TF\end{tabular}} &  Acc &  0.99913 &  0.99650 &  0.99600 & \bfseries 1.00000 & \bfseries 0.98231 & \bfseries 0.98231 & \bfseries 0.76505 & \bfseries 0.65049 & \bfseries 0.55340 & \bfseries 0.76271 & \bfseries 0.62712 & \bfseries 0.51271 \\
															% & \bfseries F1 & \bfseries 0.99912 & \bfseries 0.99650 & \bfseries 0.99625 & \bfseries 1.00000 & \bfseries 0.98237 & \bfseries 0.98225 & \bfseries 0.76336 & \bfseries 0.63742 & \bfseries 0.54967 & \bfseries 0.75084 & \bfseries 0.63908 & \bfseries 0.50652 \\
															% & \bfseries FPR & \bfseries 0.00013 & \bfseries 0.00019 & \bfseries 0.00011 & \bfseries 0.00000 & \bfseries 0.00217 & \bfseries 0.00094 & \bfseries 0.02015 & \bfseries 0.02508 & \bfseries 0.01113 & \bfseries 0.02011 & \bfseries 0.02545 & \bfseries 0.01304 \\
															
															\multirow{3}{*}{\begin{tabular}{@{}l@{}}Our Methods\\ w. BW 
																	+ TF\end{tabular}} &  Acc &  0.99913 &  0.99650 &  0.99600 & 1.00000 &  0.98231 &  0.98231 &  0.76505 &  0.65049 &  0.55340 &  0.76271 &  0.62712 &  0.51271 \\
															&  F1 &  0.99912 &  0.99650 &  0.99625 &  1.00000 &  0.98237 &  0.98225 &  0.76336 &  0.63742 &  0.54967 &  0.75084 &  0.63908 &  0.50652 \\
															&  FPR &  0.00013 &  0.00019 &  0.00011 &  0.00000 &  0.00217 &  0.00094 &  0.02015 &  0.02508 &  0.01113 &  0.02011 &  0.02545 &  0.01304 \\		
															\hline
														\end{tabular}
													}
												\end{table}								
												
												\clearpage	
												
												% \begin{figure*}[h]
													%         \centering
													%         \begin{subfigure}{0.44\textwidth}
														%                 \centering
														%                 \includegraphics[width=0.9\textwidth]{FPR/FPR_line_p.png}
														%      %								\caption{Plant Village}
														%                 \label{fig:FPR_line_p}
														%             \end{subfigure}
													%         \begin{subfigure}{0.44\textwidth}
														%                 \centering
														%                 \includegraphics[width=0.9\textwidth]{FPR/FPR_line_l.png}
														%                 \centering
														%      %								\caption{Plant Leaves}
														%                 \label{fig:FPR_line_l}
														%             \end{subfigure}
													%         \begin{subfigure}{0.44\textwidth}
														%                 \centering
														%                 \includegraphics[width=0.9\textwidth]{FPR/FPR_line_d.png}
														%      %								\caption{PlantDoc-0.2}
														%                 \label{fig:FPR_line_d}
														%             \end{subfigure}
													%         \begin{subfigure}{0.44\textwidth}
														%                 \centering
														%                 \includegraphics[width=0.9\textwidth]{FPR/FPR_line_o.png}
														%                 \centering
														%      %								\caption{PlantDoc-1.0}
														%                 \label{fig:FPR_line_o}
														%             \end{subfigure}						
													%         \caption{Comparison of False Positive Rates \\(The lower the better).}
													%         \label{fig:Comparison_of_FPR}
													%     \end{figure*}						

												\subsection{Summary and Ablation Study}
												
												\subsubsection{Backbone CNNs and ViT}
												\begin{figure*}[h!]
													\centering
													\begin{subfigure}{0.35\textwidth}
														\centering
														\includegraphics[width=\textwidth]{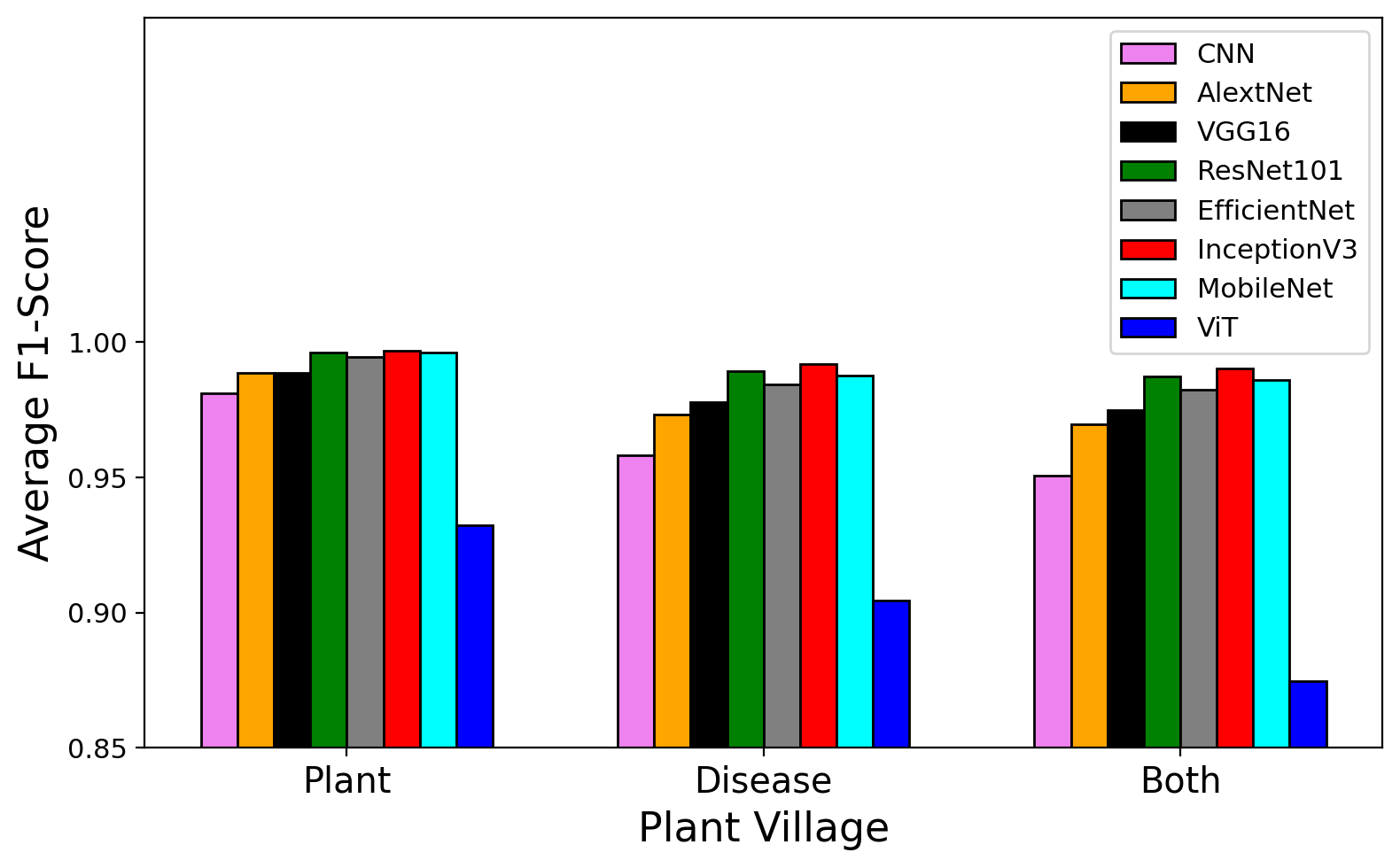}
														\caption{}
														\label{fig:backbone_Plant_village}
													\end{subfigure}
													\begin{subfigure}{0.35\textwidth}
														\centering
														\includegraphics[width=\textwidth]{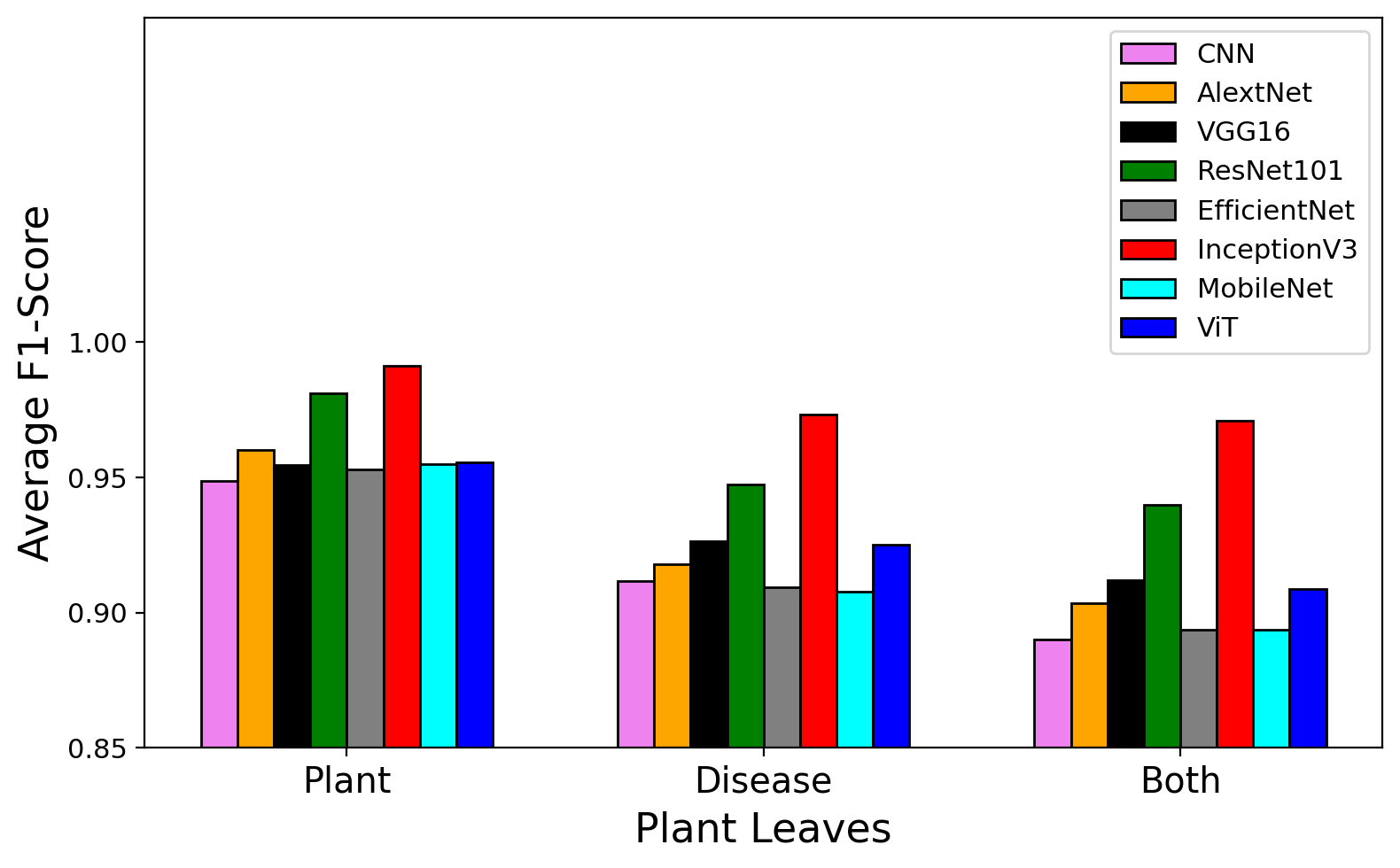}
														\centering
														\caption{}
														\label{fig:backbone_Plant_Leaves}
													\end{subfigure}\\
													\begin{subfigure}{0.35\textwidth}
														\centering
														\includegraphics[width=\textwidth]{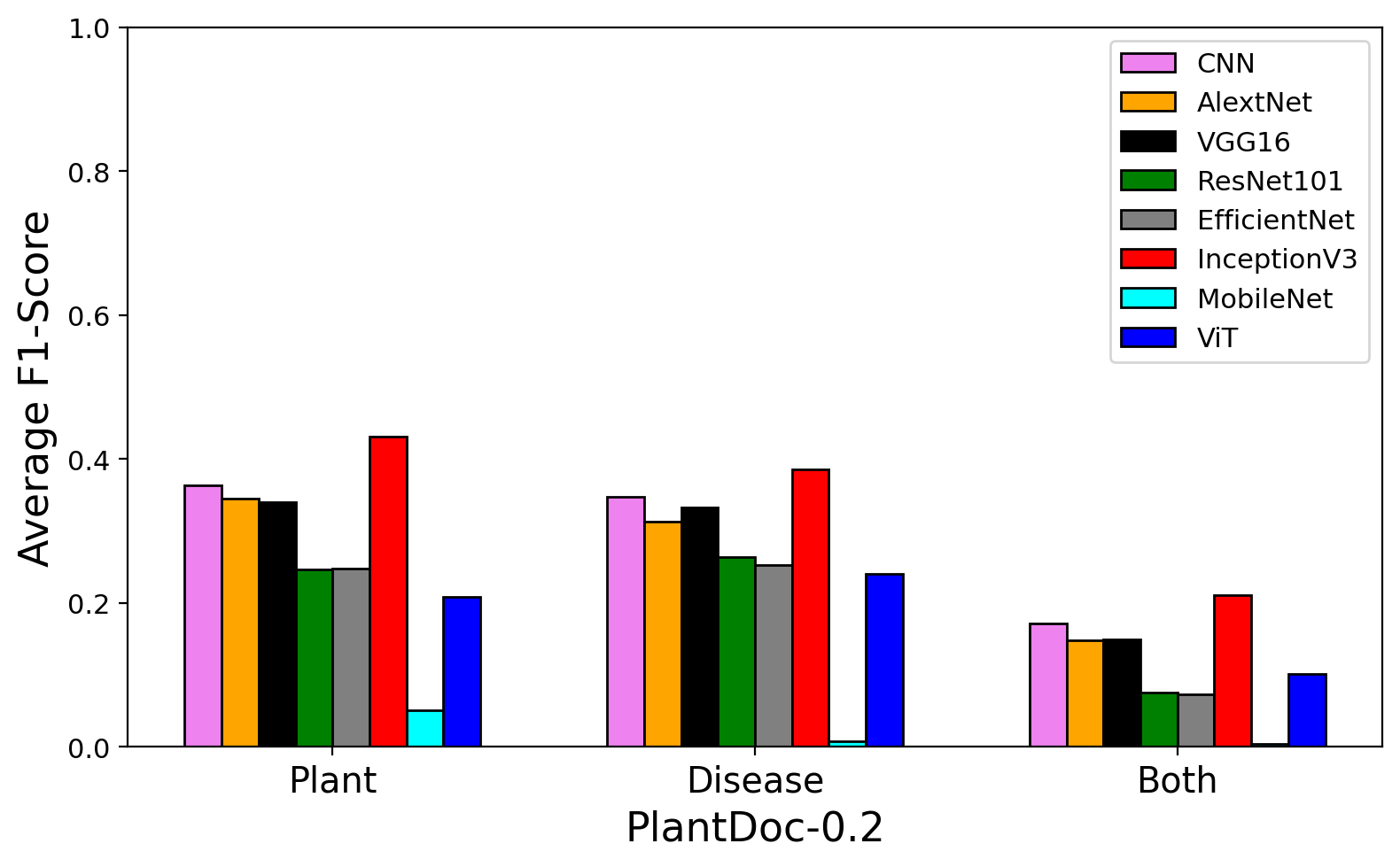}
														\caption{}
														\label{fig:backbone_PlantDoc_20}
													\end{subfigure}
													\begin{subfigure}{0.35\textwidth}
														\centering
														\includegraphics[width=\textwidth]{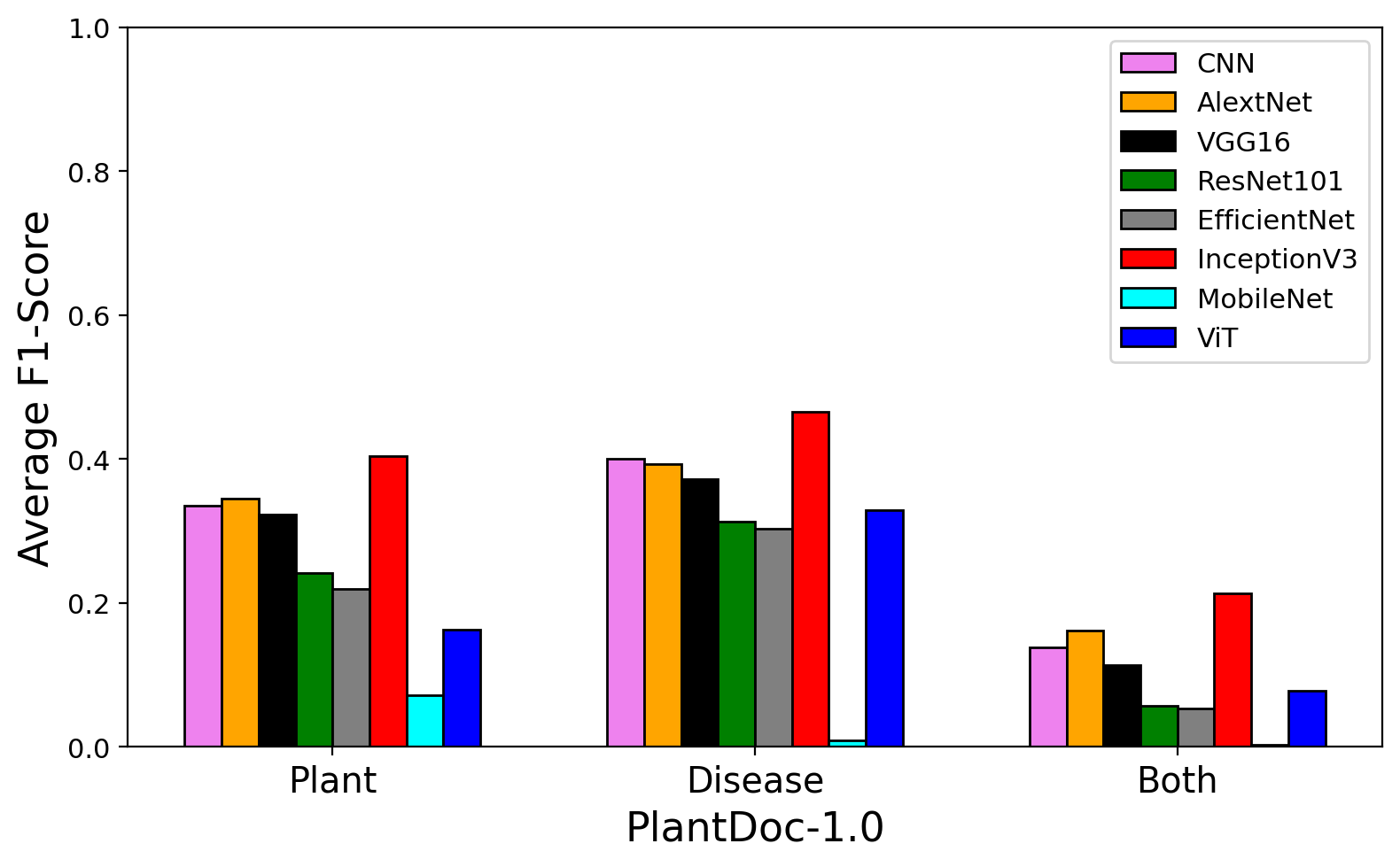}
														\centering
														\caption{}
														\label{fig:backbone_PlantDoc_Original}
													\end{subfigure}		
													\vskip -0.3cm   
													\caption{\ST{Comparison of Backbone CNNs (Including ViT).}}
													\label{fig:Comparison_of_Backbones}
													\vskip -0.3cm   
												\end{figure*}
												
												The very first question we would like to answer in this study is which CNN architectures are most useful for plant identification and disease classification. Although we cannot test all CNNs available in the literature, we selected the most popular ones in image classification and plant pathology. AlexNet, VGG16, ResNet101, EfficientNet, InceptionV3, MobileNetV2, ViT, and our custom CNN are implemented for the study. In summary, we plot four groups of histograms in  Figure \ref{fig:Comparison_of_Backbones}, where each column representing the average F1-score of a backbone CNN in all approaches of this study (multi-model, multi-label, multi-output, and GSMo-CNN). \ST{These related results are elaborated in Tables \ref{Type}, \ref{Disease}, and \ref{Total}. As highlighted in Section \ref{EvaluationMetrics}, the F1 score proves more appropriate for evaluating prediction results which have imbalanced data, as discussed earlier.} This graph aims to compare the performance of all backbone CNNs statistically and provide a view of how different backbone CNNs perform on different datasets. Each plot in Figure \ref{fig:Comparison_of_Backbones} shows average performance in plant identification, disease classification; and the combination of plant \& disease prediction (Both) over different approaches. As we can see, InceptionV3 has a better average F1-score than AlexNet, VGG16, ResNet101, EfficientNet, MobileNet, ViT, and our custom CNN in all datasets. In PlantVillage, although being better, the difference between InceptionV3 and the other backbone CNNs does not stand out, as all backbone CNNs achieve more than $0.95$ F1-score. However, in Plant Leaves, PlantDoc-0.2, and PlantDoc-1.0, InceptionV3 clearly demonstrates its advantage over the other CNNs. The second-best backbone CNN in Plant Village and Plant Leaves is ResNet101. In PlantDoc-0.2 and PlantDoc-1.0, our custom CNN and AlexNet are better than other backbones, just behind InceptionV3. This is different from Plant Village and Plant Leaves where our custom CNN has the lowest performance. \ST{Besides, ViT showcased somewhat satisfactory outcomes in testing across all datasets, but it fell short of attaining optimal results.} We can also notice that, the light-weight CNNs (MobileNetV2 and EfficientNet) perform reasonably well on Plant Village and Plant Leaves and they have similar results. However, for PlantDoc-0.2 and PlantDoc-1.0 both light-weight CNNs have the lowest F1-score with MobileNetV2 trailing behind. It would suggest that light-weights CNNs should only be used for close-up images of a leaf, and not for images with multiple leaves and complex backgrounds.
												
												\ST{The advantage of InceptionV3 can be justified by three factors. First, InceptionV3 can be deeper than VGG with a similar number of parameters. Second, InceptionV3 incorporates techniques like batch normalization and dropout to help mitigate overfitting, making it more robust on smaller datasets compared to VGG. Finally and more importantly, InceptionV3 uses a combination of convolutional layers with different kernel sizes and max-pooling layers to capture multi-scale features efficiently. This allows it to learn features representing disease types from small areas of leaves, thus, achieving competitive performance compared to other backbone CNNs.}

												% using maxima
												\begin{figure*}[h!]
													\centering
													\begin{subfigure}{0.35\textwidth}
														\centering
														\includegraphics[width=\textwidth]{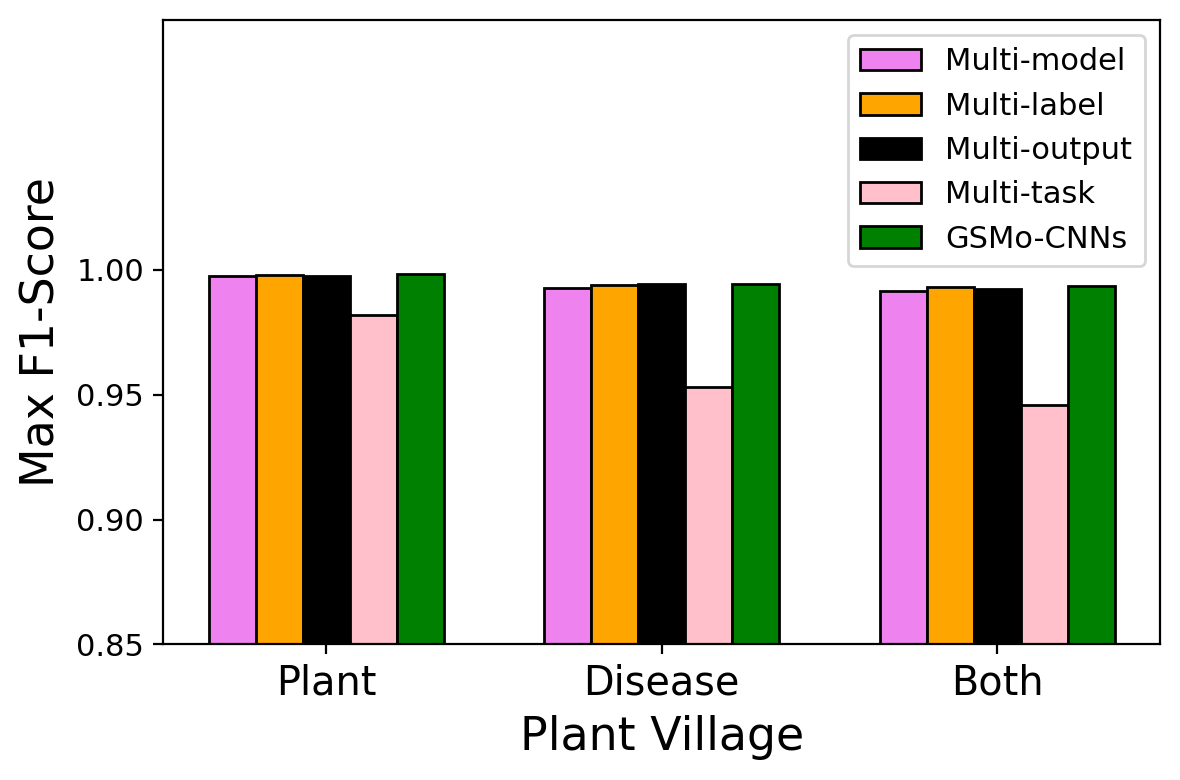}
														\caption{}
														\label{fig:approach_Plant_Village_Max}
													\end{subfigure}
													\begin{subfigure}{0.35\textwidth}
														\centering
														\includegraphics[width=\textwidth]{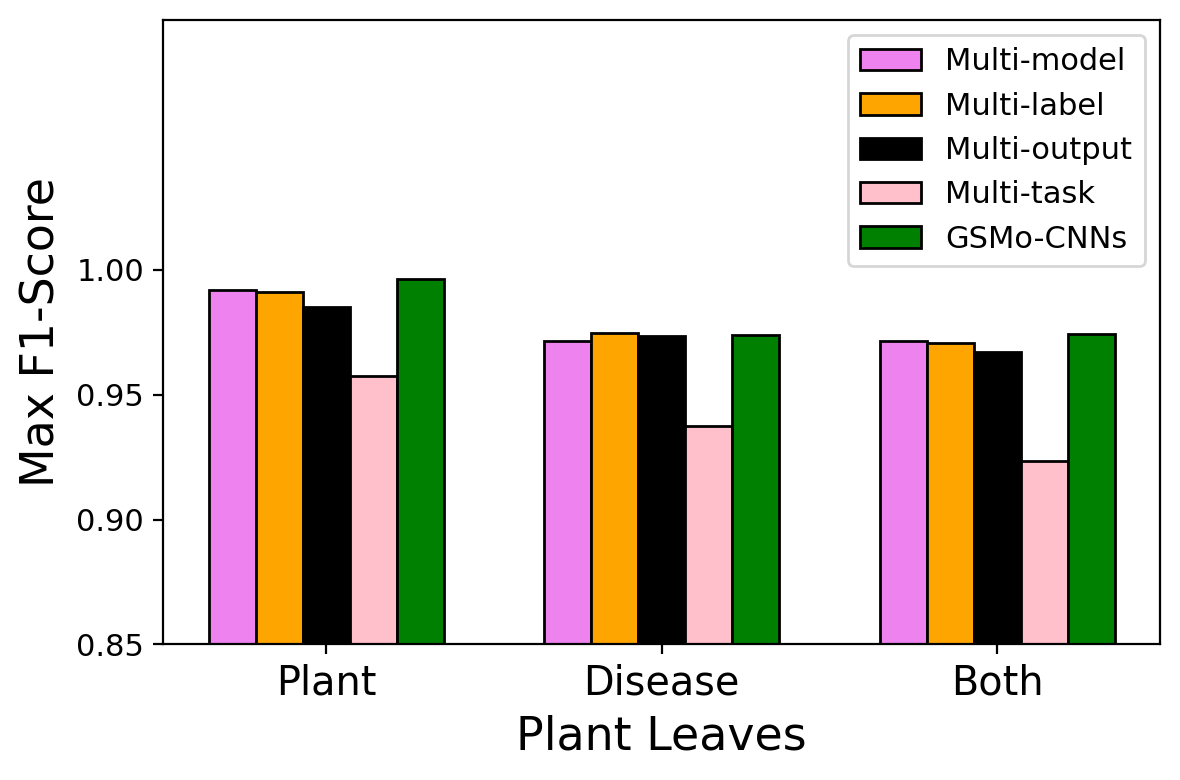}
														\centering
														\caption{}
														\label{fig:approach_Plant_Leaves_Max}
													\end{subfigure}\\
													\begin{subfigure}{0.35\textwidth}
														\centering
														\includegraphics[width=\textwidth]{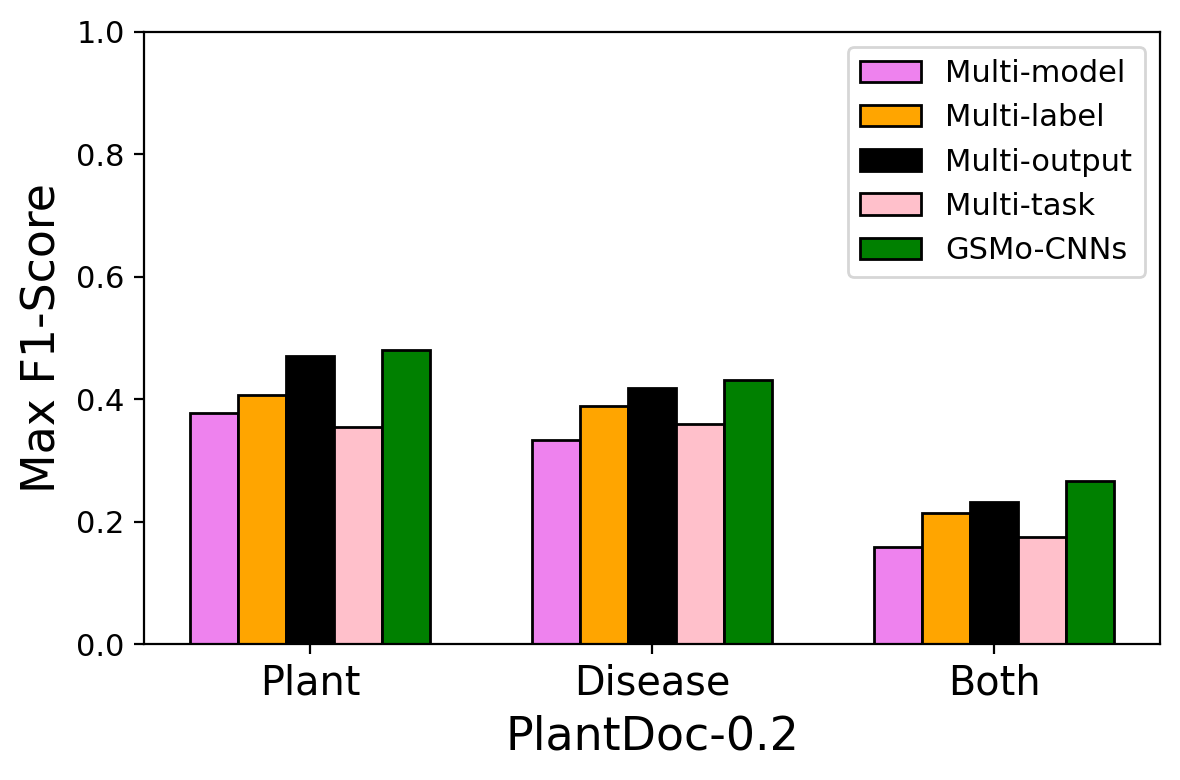}
														\caption{}
														\label{fig:approach_PlantDoc_20_Max}
													\end{subfigure}
													\begin{subfigure}{0.35\textwidth}
														\centering
														\includegraphics[width=\textwidth]{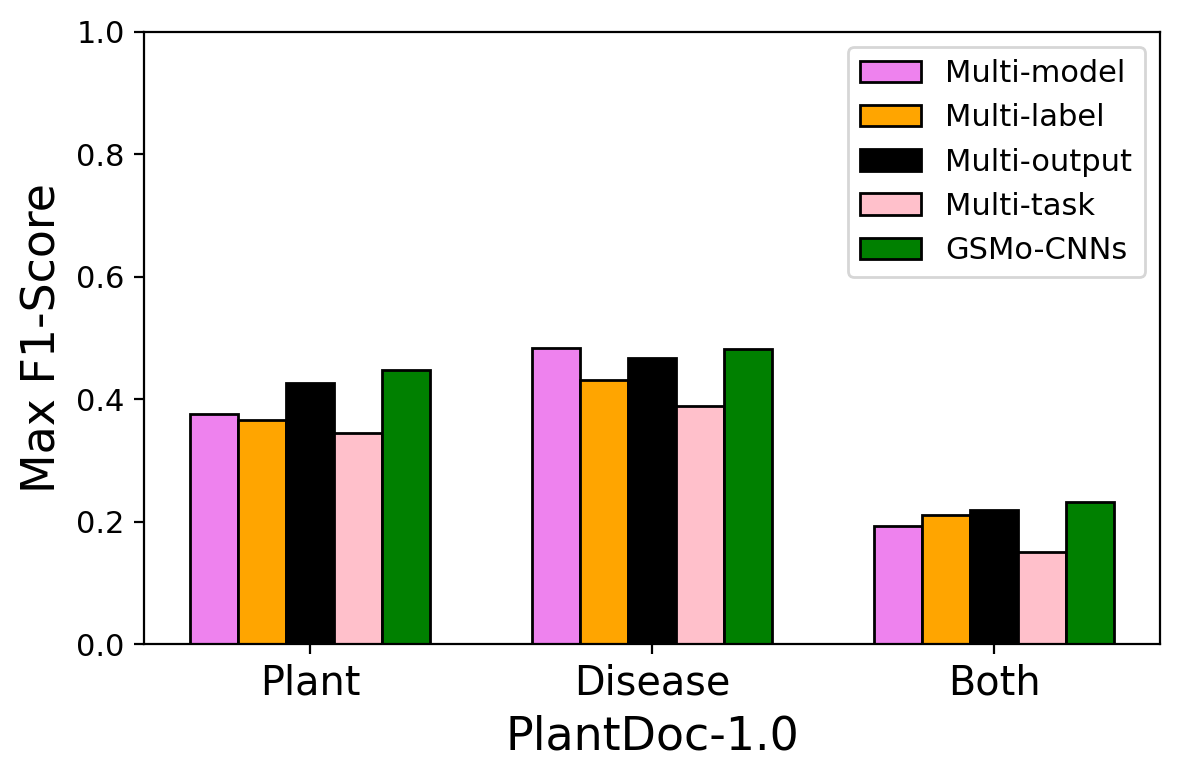}
														\centering
														\caption{}
														\label{fig:approach_PlantDoc_Original_Max}
													\end{subfigure}		
													\vskip -0.3cm   
													\caption{Comparison of Different Approaches: Multi-model, Multi-label, Multi-output, Multi-task, and GSMo-CNNs (Our Methods).}
													\label{fig:Comparison_of_Approaches_Max}
													\vskip -0.5cm   
												\end{figure*}

												\subsubsection{Single Models versus Dual Models}
												In the past, it was common and effective to use a model for a task, such as a plant species identification \cite{9804121, Shelke2022, 9631212} or leaf disease classification \cite{8974752, 9418245, 9137986}. Training an individual model for a particular task would be reasonable and straightforward as ones can optimise the model for that specific task. Multi-task learning has been shown useful, however, it is designed for tasks with different data where the combination of data can be useful for all task. This is not the case in this study where the data for all tasks are the same. That's the reason why multi-task approach does not show its advantage in our experiment. Despite that, we are interested in investigating whether using a single model for the two tasks will be better than using an ensemble of CNNs, each for a task. Also, it would be more efficient, in terms of both computation and memory storage, if a single model can effectively predict both plant species and disease types. 
												
												In Figure \ref{fig:Comparison_of_Approaches_Max} we show the comparison of all approaches in this study, including multi-model, multi-label, multi-output, multi-task, and our methods (GSMo-CNNs with \& without balance weights). Note that, the multi-model approach applies dual models, one for each task. For the sake of comparison, each column bar in the graph represents the highest F1-score of an approach in plant identification (Plant), disease classification (Disease), or combined plant \& disease prediction (Both). \ST{The related results can be checked in Tables \ref{Type}, \ref{Disease} and \ref{Total}. As mentioned in Section \ref{EvaluationMetrics}, the evaluation indicator in this part is F1-score, which is more suitable for evaluating forecasts with imbalanced data.} We can see that using dual models, one for each task, is not really advantageous as it has lower performance than approaches using single CNNs for both tasks in most cases. Compared to multi-label, multi-model is only better in Plant Leaves for plant identification, and in PlantDoc-1.0 for plant identification and combined prediction. Multi-model is shown to beat multi-output in Plant Leaves but is inferior in Plant Village, PlantDoc-0.2, and PlantDoc-1.0. Compared to our methods, multi-model achieves lower performance in all datasets. \ST{The above results demonstrate the potential and superiority of the single-model approaches, where using shared backbone CNNs for multi-predictions can be more effective than training individual CNNs, each for prediction of a label. We attribute the advantages of single models over a dual models are they can learn the common features of plant species classification and disease classification, which can influence each other efficiently. A single model have an end-to-end structure where the backbone can help to learn common features while the prediction layers will be optimised to learn task-specific features. By doing this, a single model can achieve better generalisation.}
												
												%\cmt{Add theoretical and technical justification for the results. Why single models  are better than multi-models?}
												\subsubsection{Learning Approaches}
												Figure \ref{fig:Comparison_of_Approaches_Max} also shows F1-score of different approaches \ST{(the related results have been detailed in Tables \ref{Type}, \ref{Disease} and \ref{Total}). As mentioned in Section \ref{EvaluationMetrics}, the F1 score is better suited for evaluating prediction results on imbalanced data compared to Accuracy.} Overall, the proposed GSMo-CNNs achieve the best results in all datasets, followed by multi-output and multi-label. Multi-output is comparable to multi-label in Plant Village and slightly better in Plant Leaves, however, multi-label is better in PlantDoc-0.2 and PlantDoc-1.0. Compared to existing methods, as summarised in Table \ref{DL_Comparison} we achieve state-of-the-art results. The only approach achieves $99.98\%$ validation accuracy on Plant Village is AlexNet+SVM in \cite{9250885}. However, this is a hybrid approach and the result is for a validation set, and it is not clear how the data is partitioned for evaluation. 
												
												\ST{The remarkable performance of GSMo-CNNs can be attributed to several factors. First, it uses a chain of prediction  to jointly predict species and diseases. This would enhance feature learning and relationships of plant species and disease types, and facilitates feature sharing, enabling the model to extract general image characteristics effectively; Second, the balance weights were employed to address class imbalances, ensuring accurate predictions for all classes. These theoretical and technical advantages collectively make GSMo-CNNs an advantegous approach for plant species identification and disease classification.}

												%\cmt{Add theoretical and technical justification for the results. Why GSMo-CNN work well?}
												
												% \clearpage
												\subsubsection{Balance Weights}
\begin{figure}[h!]
													\centering
													\begin{tabular}{c c}
														\begin{subfigure}{0.26\textwidth}
															\centering
															\includegraphics[width=\textwidth]{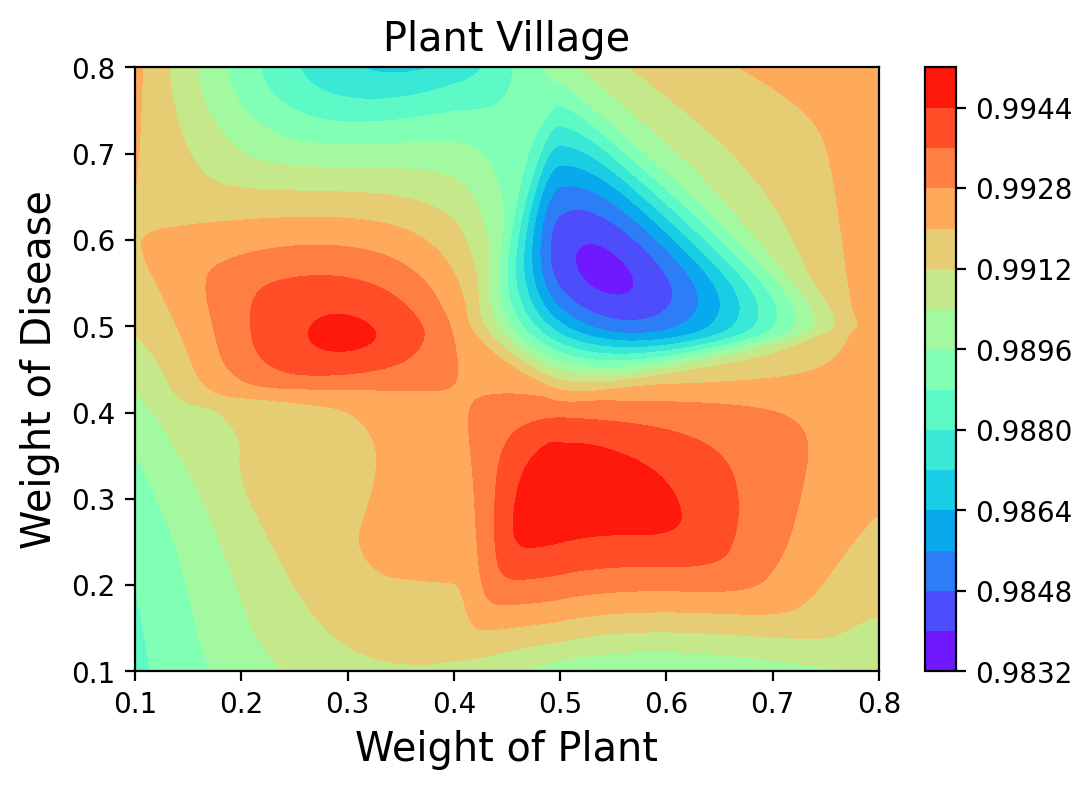}
															\centering
															\caption{}
															\label{fig:heatmap_Plant_Village_wp_wd}
														\end{subfigure}&  
														\begin{subfigure}{0.26\textwidth}
															\centering
															\includegraphics[width=\textwidth]{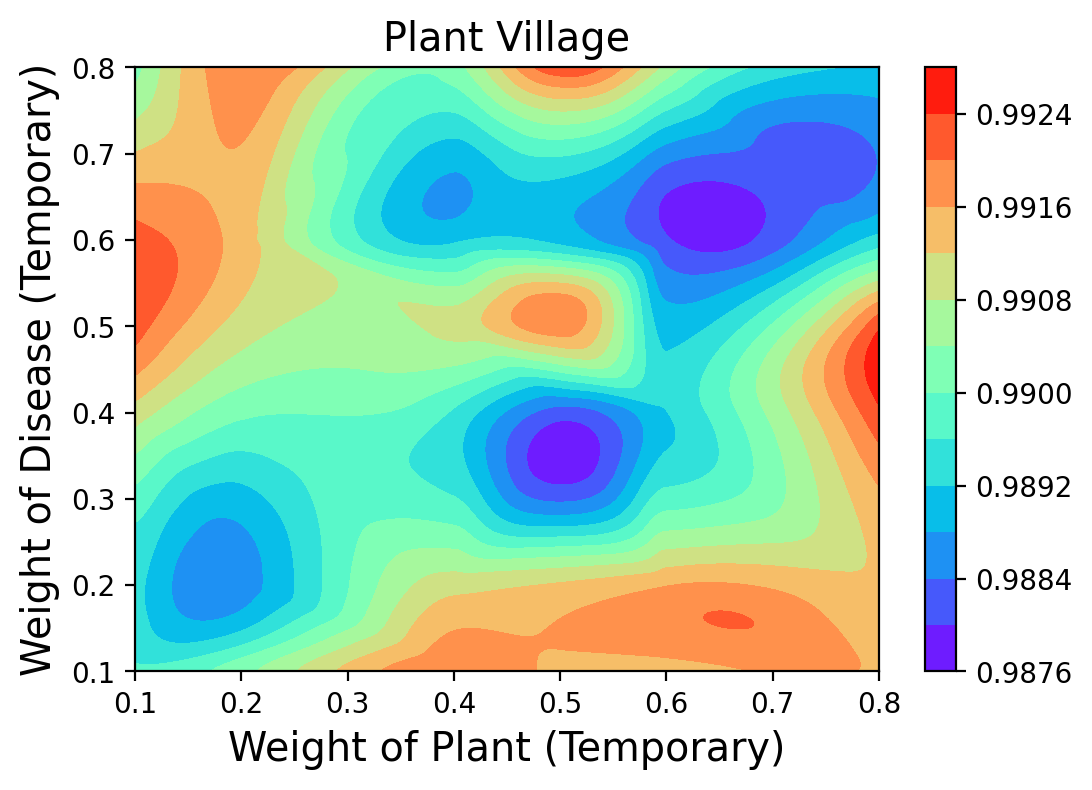}
															\caption{}
															\label{fig:heatmap_Plant_Village_wpt_wdt}
														\end{subfigure}\\
														\begin{subfigure}{0.26\textwidth}
															\centering
															\includegraphics[width=\textwidth]{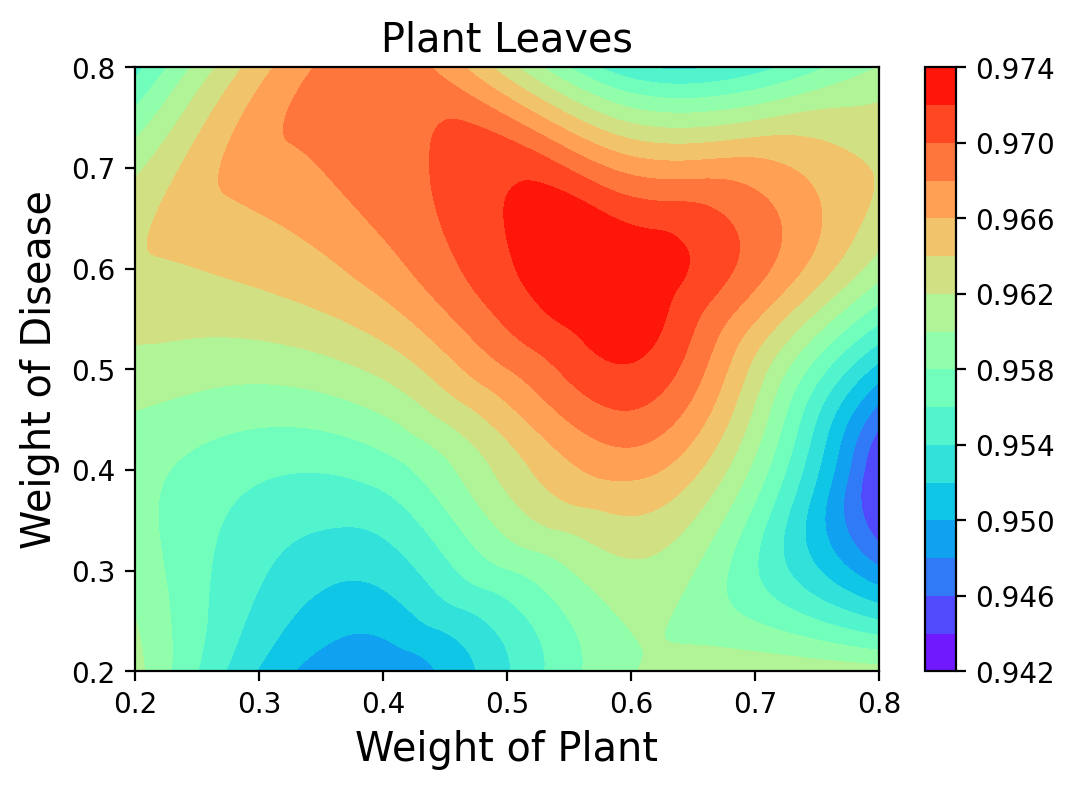}
															\centering
															\caption{}
															\label{fig:heatmap_Plant_Leaves_wp_wd}
														\end{subfigure}&
														\begin{subfigure}{0.26\textwidth}
															\centering
															\includegraphics[width=\textwidth]{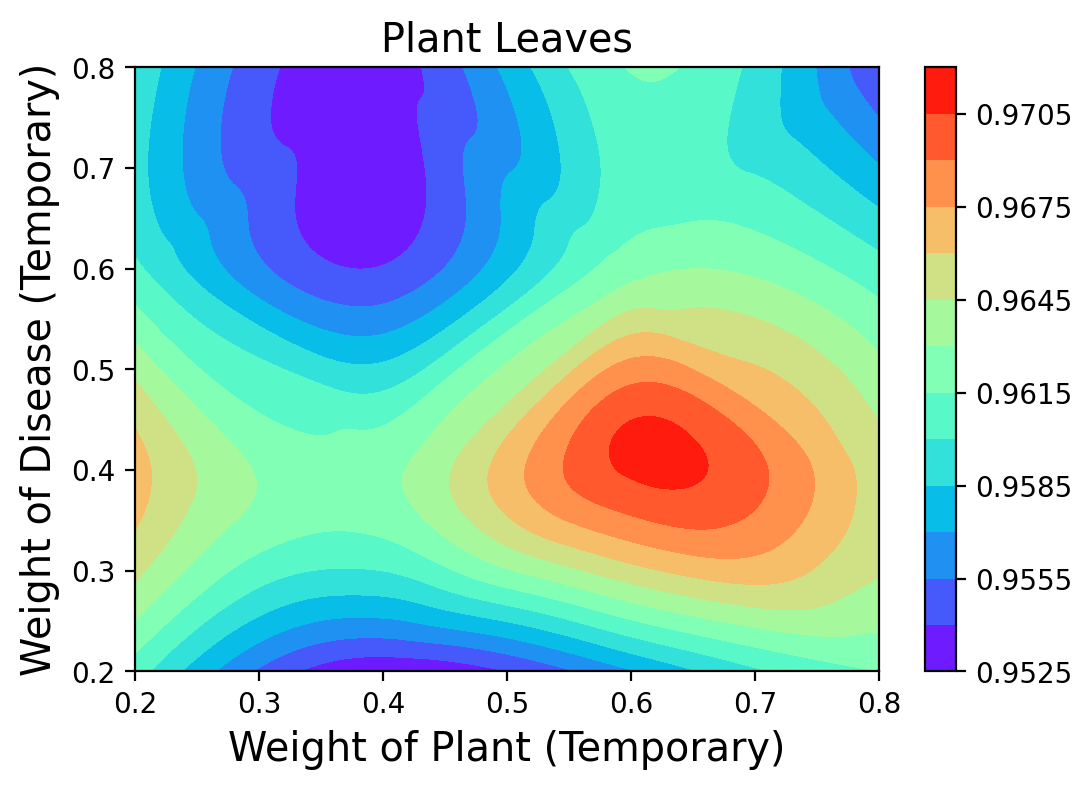}
															\centering		
															\caption{}
															\label{fig:heatmap_Plant_Leaves_wpt_wdt}
														\end{subfigure} \\
														\begin{subfigure}{0.26\textwidth}
															\centering
															\includegraphics[width=\textwidth]{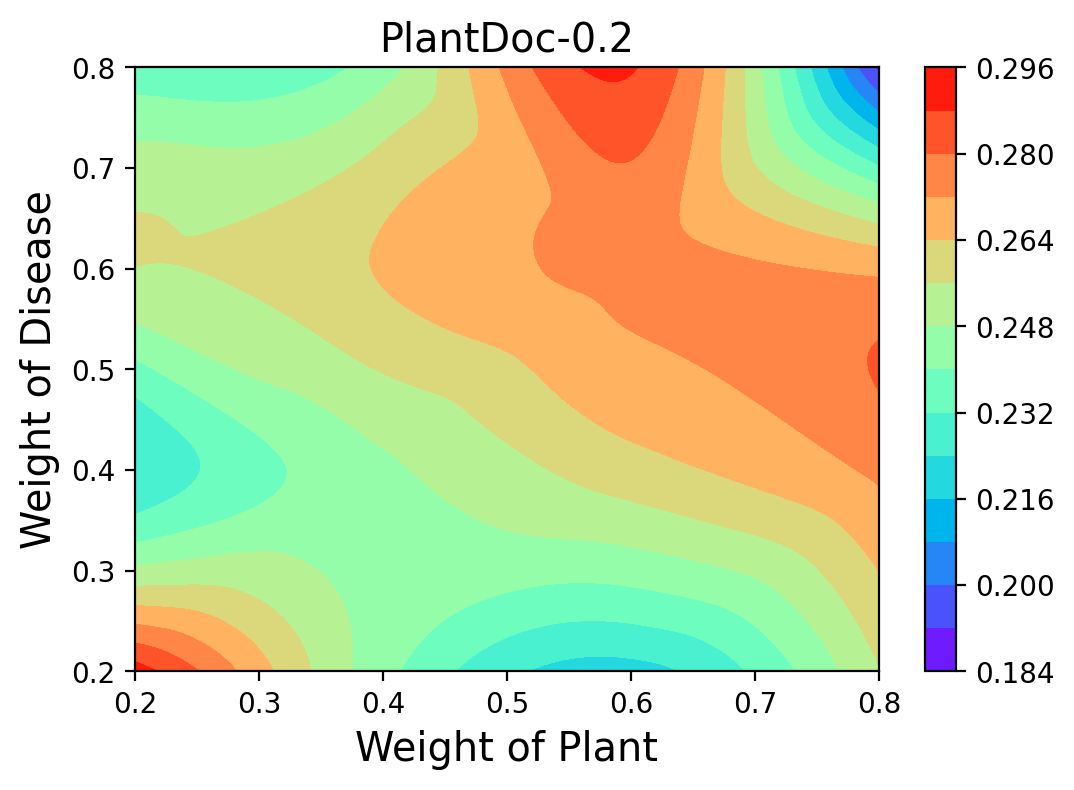}
															\centering
															\caption{}
															\label{fig:heatmap_PlantDoc_20_wp_wd}
														\end{subfigure}
														&
														\begin{subfigure}{0.26\textwidth}
															\centering
															\includegraphics[width=\textwidth]{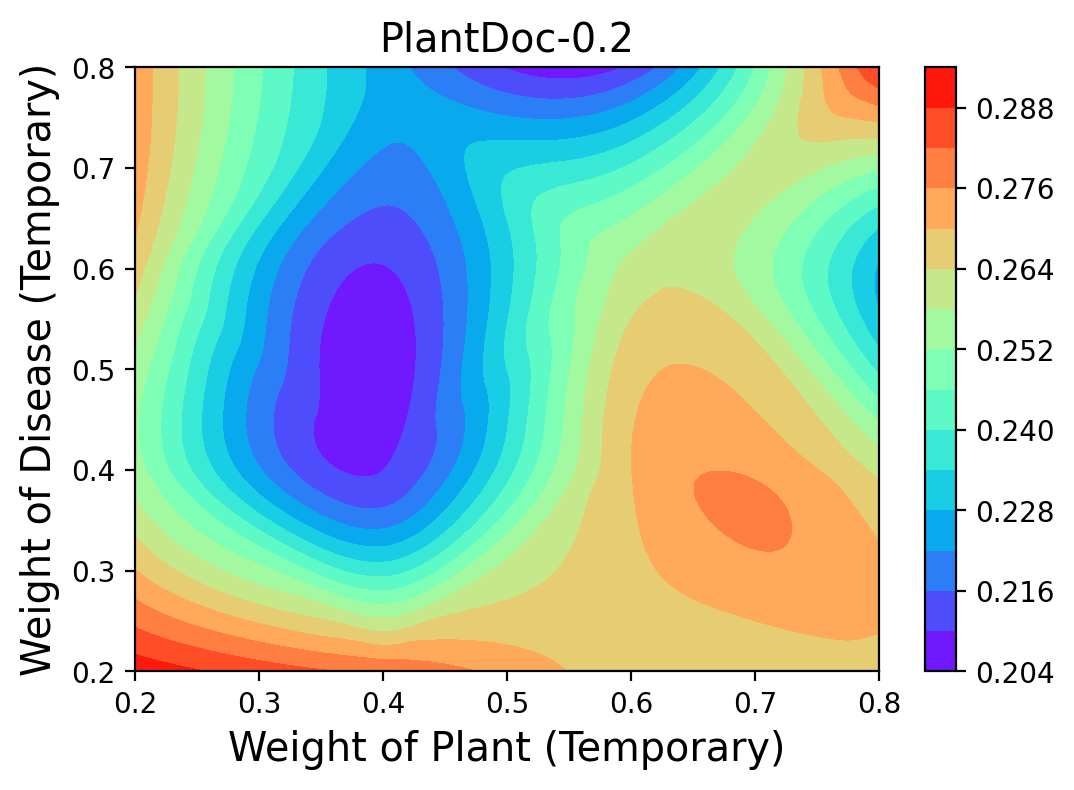}
															% \centering
															\caption{}
															\label{fig:heatmap_PlantDoc_20_wpt_wdt}
														\end{subfigure}	
														\\
														\begin{subfigure}{0.26\textwidth}
															% \centering
															\includegraphics[width=\textwidth]{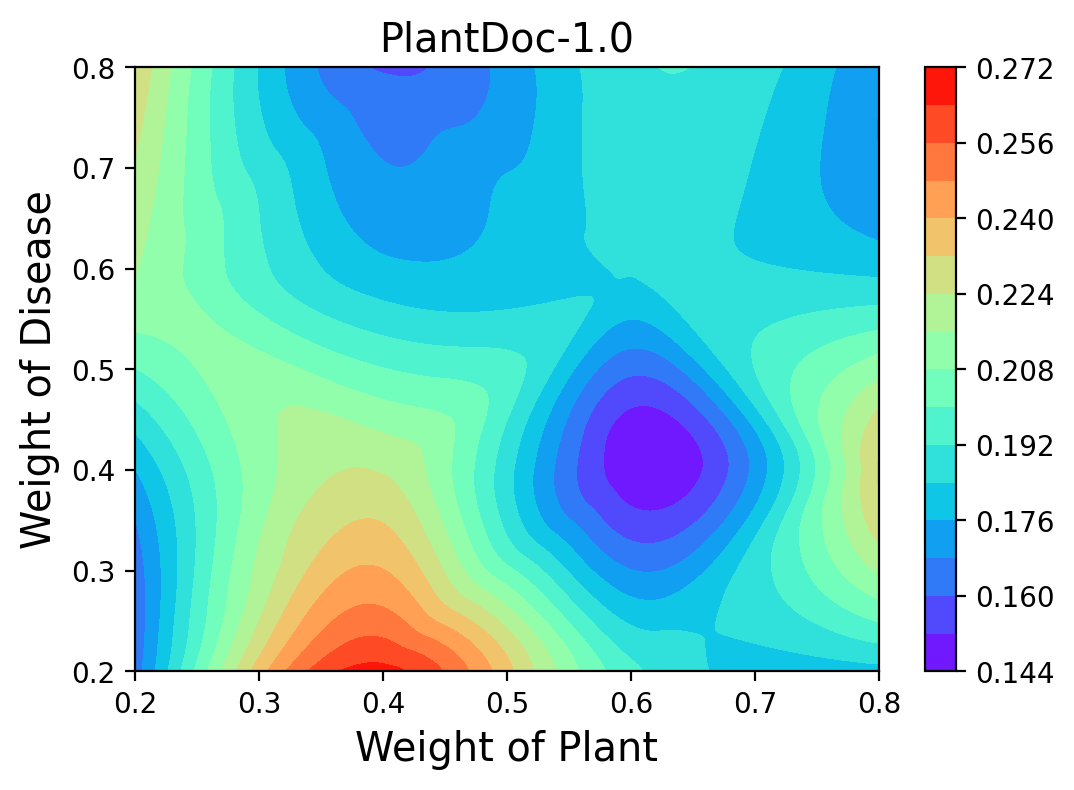}
															\caption{}
															\label{fig:heatmap_PlantDoc_original_wp_wd}
														\end{subfigure}
														&
														\begin{subfigure}{0.26\textwidth}
															% \centering
															\includegraphics[width=\textwidth]{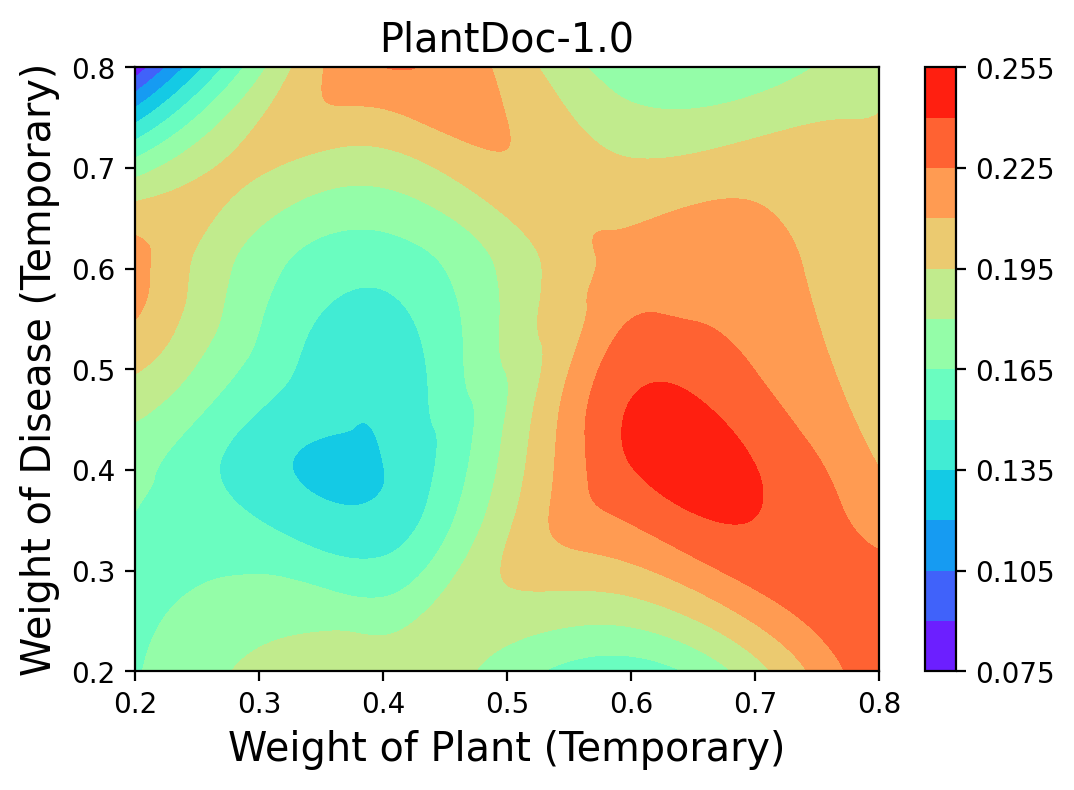}
															% \centering
															\caption{}
															\label{fig:heatmap_PlantDoc_original_wpt_wdt}
														\end{subfigure}
														\\
													\end{tabular}
													\vskip -0.3cm   
													\caption{Combinations of Balance Weights for Plant Prediction and Disease Prediction.}
													\label{fig:Comparison_of_weights}
													\vskip -0.5cm   
												\end{figure}

												As presented in Section \ref{Models}, GSMo-CNN can use balance weights for the loss functions. We show the heat maps for the balance weights of plant and disease on the final prediction layer and the balance weights of plant and disease on the temporary layer in Figure 
												\ref{fig:Comparison_of_weights}. The hotter colour represents the higher F1-score. As we can see, \ST{in general, the balance weights for plant and disease need to be analysed specifically, as different tasks have varying weight values. For example, in Figure \ref{fig:heatmap_Plant_Village_wp_wd}, the highest F1-scores of GSMo-CNN in the Plant Village dataset appear in 0.3 Weights of Plant (WP) \& 0.5 Weights of Disease (WD) and 0.5 - 0.6 WP \& 0.2 - 0.4 WD. But in Figure \ref{fig:heatmap_Plant_Leaves_wp_wd}, the highest F1-scores of GSMo-CNN in the Plant Leaves dataset appear in 0.6 WP \& 0.6 WD, which trend to similar.}

												\subsubsection{Prediction Layer}
												
\begin{figure}[h!]
													\centering
													\begin{subfigure}{0.35\textwidth}
														\centering
														\includegraphics[width=\textwidth]{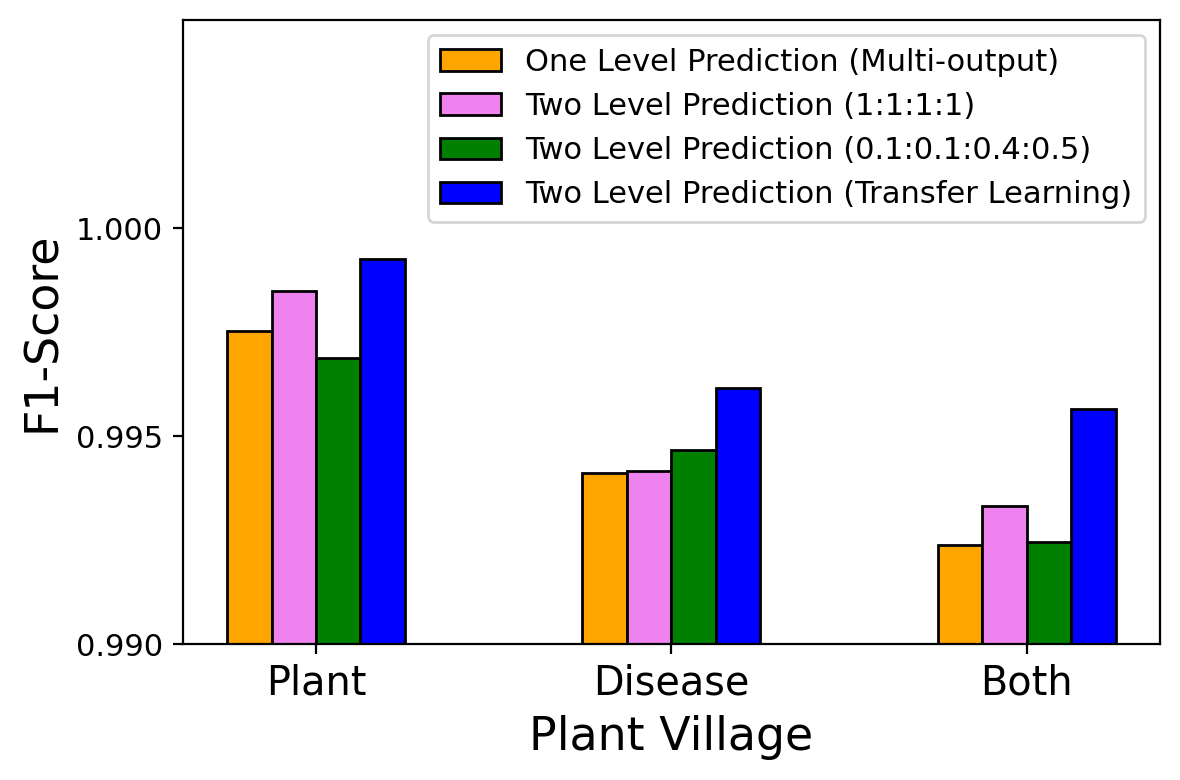}
														%		 		\caption{Plant Village}
														\label{fig:Multi_output_BW_and_IBW_Plant_Village}
													\end{subfigure}
													\begin{subfigure}{0.35\textwidth}
														\centering
														\includegraphics[width=\textwidth]{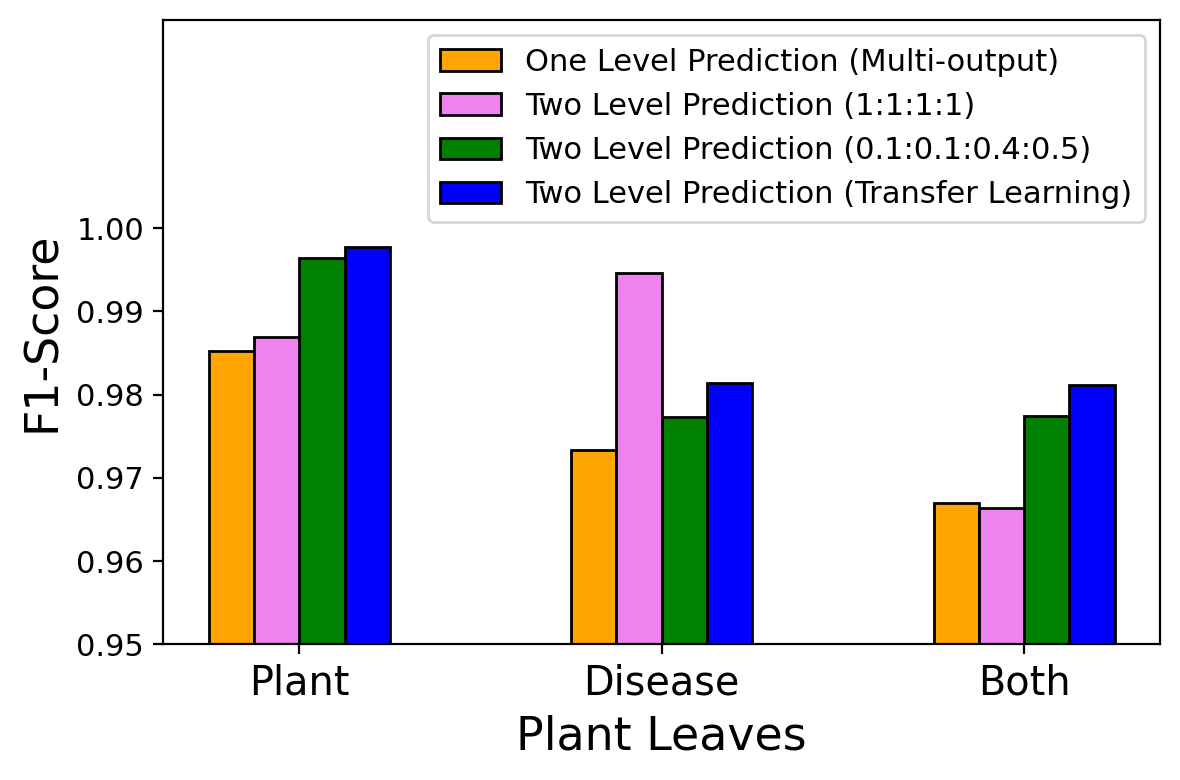}
														\centering
														%								\caption{Plant Leaves}
														\label{fig:Multi_output_BW_and_IBW_Plant_Leaves}
													\end{subfigure}\\
													\begin{subfigure}{0.35\textwidth}
														\centering
														\includegraphics[width=\textwidth]{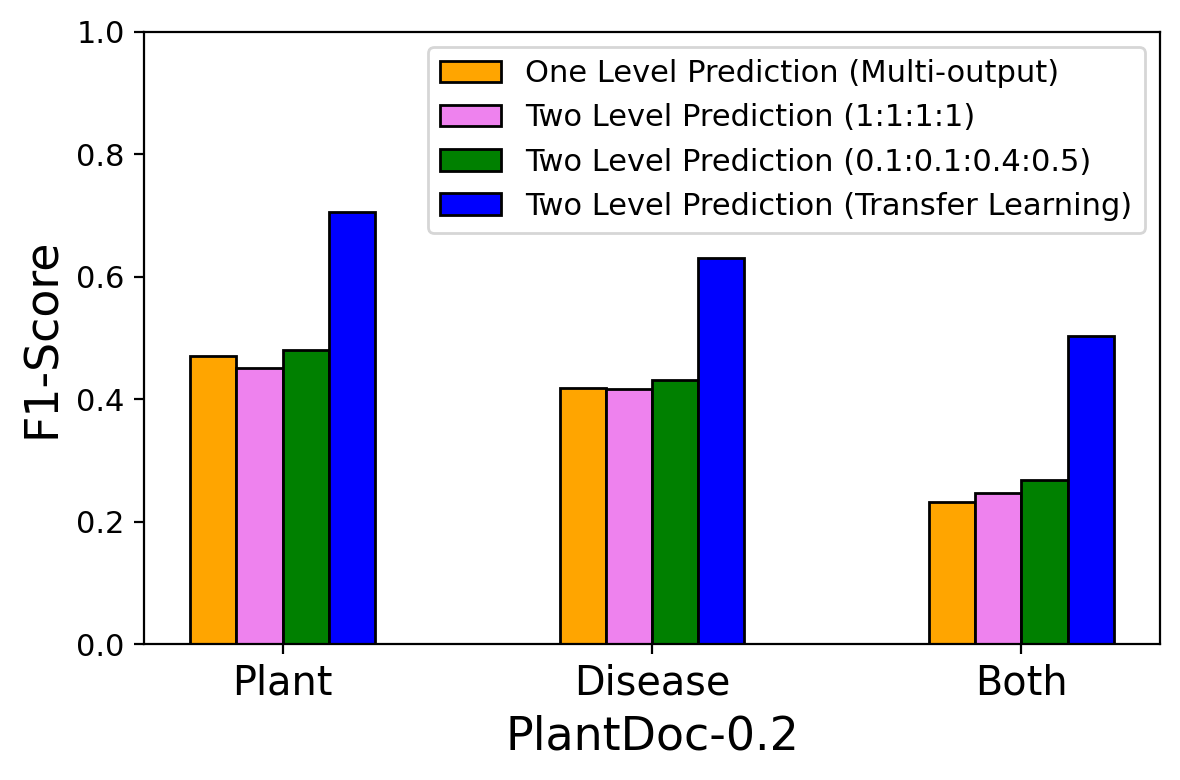}
														%								\caption{PlantDoc-0.2}
														\label{fig:Multi_output_BW_and_IBW_PlantDoc_20}
													\end{subfigure}
													\begin{subfigure}{0.35\textwidth}
														\centering
														\includegraphics[width=\textwidth]{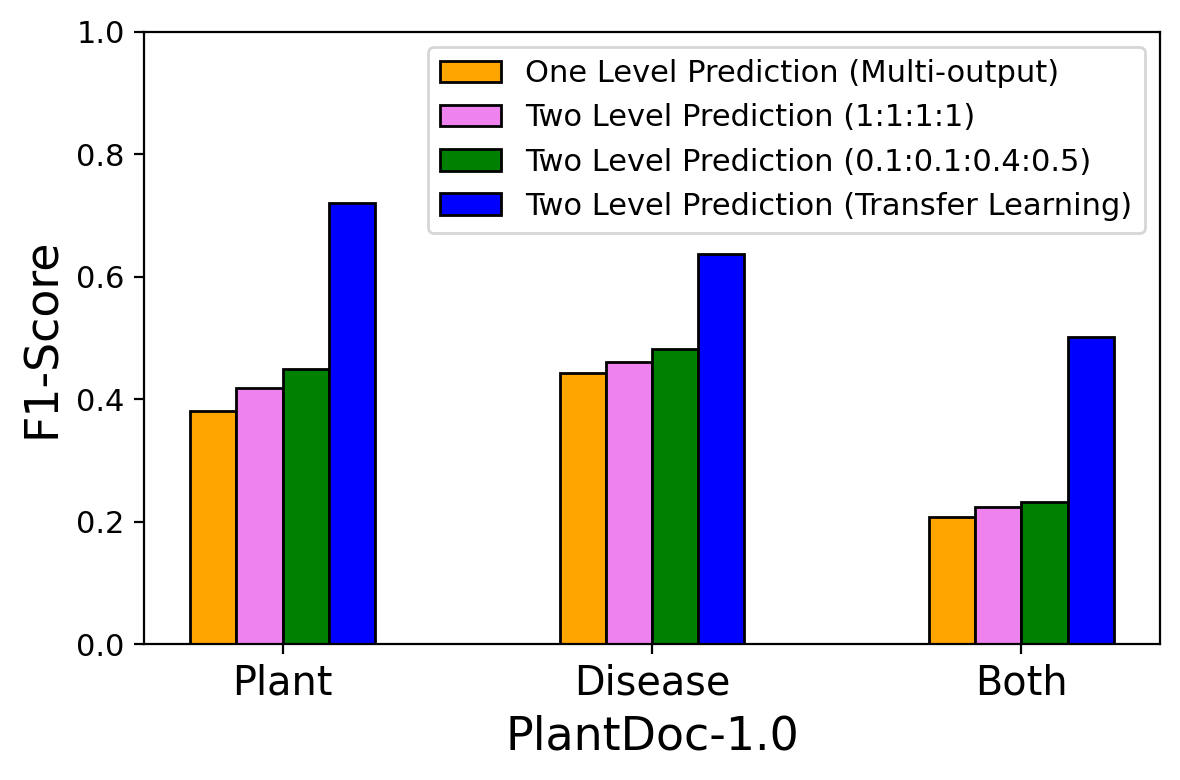}
														\centering
														%								\caption{PlantDoc-1.0}
														\label{fig:Multi_output_BW_and_IBW_PlantDoc_Original}
													\end{subfigure}					
													\vskip -0.5cm   
													\caption{Comparison of 1st \& 2nd Prediction Layer.}
													\label{fig:Comparison_of_Multi_output_BW_and_IBW}
													\vskip -0.6cm   
												\end{figure}

												In Figure \ref{fig:Comparison_of_Multi_output_BW_and_IBW}, we show that by stacking prediction layers we can achieve the performance and the balance weights can help increase F1-score. Finally, we show that transfer learning would be useful in this study.
												
												As we can see, with one prediction level we have the multi-output approach which achieves good results on Plant Village and Plant Leaves, although the performance on PlantDoc-0.2 and PlantDoc-1.0 is not high it is still comparable or better than multi-model, multi-label, multi-task approaches. When adding a stack of another prediction layer on top of the first (temporary) prediction layer with cross connections: disease (temporary) $\rightarrow$ plant; plant (temporary) $\rightarrow$ disease, we can achieve the improvement in all cases, except plant identification and disease classification in PlantDoc-0.2. With balance weights ($\beta_1:\delta_1:\beta_2:\delta_2=0.1:0.1:0.4:0.5$) for the training loss functions, further improvement can be seen in 9 out of 12 cases.  The other 3 cases are plant identification (Plant Village), disease classification (Plant Leaves), and combined prediction (Plant Village) where the balance weights do not show their advantage. Finally, the usefulness of transfer learning is apparent as it only fails to increase the performance of disease classification on Plant Leaves. Especially on PlantDoc-0.2 and PlantDoc-1.0, we can achieve significant improvement with large margins.

												\section{Conclusion and Future Work}
												\label{sec:conclusion}
												We presented a comprehensive survey and empirical study on deep learning for plant identification and disease classification from leaf images. The paper aims to address the gaps in modern plant pathology where cutting-edge technologies such as deep learning have been adopted largely but there is still a lack of benchmarking studies. Besides reviewing currently used methods, we also show the available methods from machine learning literature which haven’t been employed for plant identification and/or disease classification. Furthermore, we investigate the hypothesis that a single model for multi-prediction would be more useful than multiple models, each for a task, in terms of implementation, computation, memory saving, and effectiveness. To this end, we categorise different approaches into multi-model, multi-label, multi-output, and multi-task where different CNN structures can be the backbone, if applicable. For completeness, we also propose a new model, based on the multi-output approach and the idea of classifier chain, by stacking and cross-connecting prediction layers. We run intensive experiments to evaluate and compare these backbone models and approaches in uniform settings. We found that:
												\ST{\begin{itemize}
														\item InceptionV3 is a best choice for a backbone CNN in this study as it performs better than AlexNet, VGG16, ResNet101, MobileNetV2, EfficientNet, ViT, and our custom CNN.
														\item Using a single model for both tasks is more useful than using separate models for each of them. Single models can be more convenient, i.e. easier for model selection, more memory saving, more efficient, and we also achieve better results.
														\item Stacking and cross-connecting prediction layers can improve the accuracy and F1-score for plant identification and disease classification. This approach is also flexible where balance weights can be applied to search for a better combination of the loss functions at the prediction layers.
														\item Transfer learning is promising. We showed that by transferring InceptionV3’s weights trained on ImageNet we can improve the performance of our new model significantly.
												\end{itemize}}
												%For future work, we would like to develop an algorithm to automatically determine the balance weights. With many public datasets, each for different types of plants and diseases, we would like to see the advantage of multi-task learning.  This will be the focus of our next study.
												
												\ST{The future directions of this research will be extensive and have the potential to significantly benefit the agricultural industry.
													First, based on the promising results of GSMo-CNNs in this paper, we can expand the idea of a hierarchical combination of labels for further improvement. One idea is to pioneer the development of an algorithm capable of autonomously determining optimal balance weights. This innovation holds immense potential, particularly in scenarios characterized by a multitude of publicly available datasets, each associated with diverse plant species and diseases. Additionally, we can explore the integration of deep supervision to smooth the gradients during the training at different prediction levels. Here, at each output layer, we compute a prediction loss and add it to the final loss function.}
												
												\ST{Second, we found that each dataset has different sets of plant species and disease types. While we can build a model on each dataset for a narrow task, it is unable to deploy it to another task for new plants and diseases. We intend to delve deeper into the realm of life-long learning, seeking to uncover the full extent of its advantages and applications. A foundation model for plant pathology can be developed so that it can continually acquire and adapt knowledge over time. In particular, the model can learn from newly available datasets without re-training from the previously learned datasets. At the same time, one can distill knowledge from the model for different downstream tasks.}
												
												\ST{We believe that these efforts and endeavours will propel our research further, leading to new breakthroughs that will benefit the field of precision agriculture and smart agriculture.}
												
												\bibliographystyle{ACM-Reference-Format}
												\bibliography{main.bib}
												%\bibliography{mybibfile.bib}
											\end{document}